\documentclass[final,5p,times,twocolumn]{elsarticle}

 \usepackage[numbers]{natbib}

\usepackage{amsthm,amssymb,mathrsfs,setspace,pstricks,booktabs,mathtools,amsmath}
\usepackage{subcaption}
\usepackage{algorithm}
\usepackage{graphicx}
\usepackage{algpseudocode}
\usepackage{url}

\usepackage[utf8]{inputenc}
\UseRawInputEncoding 
\begin{document}
\let\WriteBookmarks\relax
\def\floatpagepagefraction{1}
\def\textpagefraction{.001}

% Short title
%\shorttitle{Leveraging social media news}

% Short author
%\shortauthors{CV Radhakrishnan et~al.}

% Main title of the paper
\title{An evaluation of  Google Translate for Sanskrit to English translation via sentiment and semantic analysis }                 
%\tnotetext[1]{This document is the results of the research
 %  project funded by the National Science Foundation.}

\author[2,fn1]{Akshat Shukla}    
\author[1,fn1]{Chaarvi Bansal}  
\author[4]{Sushrut Badhe }
\author[3]{Mukul Ranjan}
\author[5,cor]{Rohitash Chandra} 

\address[1]{Department of Computer Science and Information Systems, Birla Institute of Technology and Science Pilani, Pilani, Rajasthan, India}
\address[2]{Department of Civil Engineering, Indian Institute of Technology Delhi, Delhi, India}
\address[3]{Department of Electronics and Electrical Engineering,Indian Institute of Technology Guwahati, Guwahati, Assam, India}

\address[4]{Midam Charitable Trust, Puducherry, India}

\address[5]{Transitional Artificial Intelligence Research Group, School of Mathematics and Statistics, University of New South Wales, Sydney, Australia}

% Corresponding author text
\cortext[cor]{Corresponding author} 

% Footnote text
\fntext[fn1]{Equal contributions}
%\fntext[fn2]{Another author footnote, this is a very long footnote and
%  it should be a really long footnote. But this footnote is not yet
  %sufficiently long enough to make two lines of footnote text.}

% For a title note without a number/mark
%\nonumnote{This note has no numbers. In this work we demonstrate $a_b$
  %the formation Y\_1 of a new type of polariton on the interface
 % between a cuprous oxide slab and a polystyrene micro-sphere placed
 % on the slab.
 % }

% Here goes the abstract
\begin{abstract} 
Google Translate has been prominent for language translation; however, limited work  has been done in evaluating   the  quality of translation when compared to  human experts.  Sanskrit one of the oldest written languages in the world. In 2022, the Sanskrit language was added to the Google Translate engine.  Sanskrit is known as the mother of  languages such as Hindi and an ancient source of the Indo-European group of languages.  Sanskrit is the original language for sacred Hindu texts such as the Bhagavad Gita. 
In this study, we present a framework that evaluates the Google Translate for Sanskrit using the Bhagavad Gita. We first publish  a translation of the Bhagavad Gita in Sanskrit  using  Google Translate. Our framework then compares Google Translate version of Bhagavad Gita with  expert translations using sentiment  and semantic analysis via BERT-based language models. Our results indicate that  in terms of sentiment and semantic analysis, there is low level of similarity in selected verses of Google Translate when compared to expert translations. In the qualitative evaluation, we find that Google translate is unsuitable for translation of certain Sanskrit words and phrases due to its poetic nature, contextual significance, metaphor and imagery. The mistranslations  are not surprising since the Bhagavad Gita is known as  a difficult text not only to translate, but also to interpret since it relies on contextual, philosophical  and historical information. Our framework lays the foundation for automatic evaluation of other languages by Google Translate. 
 
\end{abstract}

% Use if graphical abstract is present
% \begin{graphicalabstract}
% \includegraphics{figs/grabs.pdf}
% \end{graphicalabstract}

% Research highlights
%\begin{highlights}
%\item Research highlights item 1
%\item Research highlights item 2
%\item Research highlights item 3
%\end{highlights}

% Keywords
% Each keyword is seperated by \sep

\maketitle

\section{Introduction}  

Deep learning methods have proven to be   powerful in handling   data in different formats such as numerical, textual, video, audio, and image in large volumes \cite{najafabadi2015deep}. Natural Language Processing (NLP) \cite{manning1999foundations} is a field of artificial intelligence that empowers machines to process, interpret and understand text and language just as humans. In the past, NLP found numerous applications in the field of text processing such as sentiment analysis \cite{Chandra2022-mh, dang2020sentiment}, topic modeling \cite{kirill2020propaganda, egger2022topic}, speech translation \cite{bertoldi2007speech,nakamura2006atr}, named entity recognition \cite{mikheev1999named,marrero2013named}, etc.  NLP combines the field of computational linguistics with deep learning, statistics, and machine learning \cite{nadkarni2011natural}. In the last decade, a variety of deep learning models have been applied for NLP that have boosted the field with a number of innovations \cite{socher2012deep}. Semantic and sentiment analysis are two of the most prominent text processing applications given their applications in social media and marketing. It has shown that sentiment analysis can also be used for predictive modelling for election outcomes  via  the US 2020 general elections \cite{chandra2021biden}.
 
Language translation models use computer systems to  translate text in a source language to an equivalent text in the target language \cite{garg2018machine}. An efficient translation model is a key to many trans-lingual applications \cite{mizera2010multilingual}, cross-language information retrieval \cite{oard1998cross}, computer-assisted language learning \cite{beatty2013teaching}, etc. In the past, numerous systems have been proposed that either improve the quality of the generated translations \cite{johnson2007improving} or study the robustness of these systems by evaluating their performance for different target languages \cite{bisang2022evaluate}. Neural machine translation (NMT) \cite{kalchbrenner2013recurrent,zhang2020neural} uses a recurrent neural network (RNN) model to predict the likelihood of a sequence of words. It typically models an entire sentence in a single integrated model \cite{sutskever2014sequence,sennrich2015improving}. On the other hand, the Transformer \cite{vaswani2017attention} is an attention-based model that remains a dominant architecture for several language pairs \cite{barrault2019findings}. The self-attention layers of the Transformer model learn the dependencies between words in a sequence by examining links between all the words in the paired sequences and by directly modeling those relationships \cite{wdowiak2021sicilian}. Language translation is perhaps one of the most difficult modelling  task considering the fluidity of human language \cite{mathur2020tangled}. Nowadays, deep neural network models and encoder-decoder attention-based RNN such as the \textit{bidirectional encoder representations from Transformer} (BERT) \cite{devlin2018bert}, have achieved state-of-the-art results in language modelling tasks with special properties\cite{tenney2019bert, kitaev2020reformer}. 

%NLP used for religion (Bhagavad Gita and Upshands paper) Chandra and Venkatesh, Chandra and Ranjan
The \textit{Bhagavad Gita} (translates as the \textit{song of God}) is  sacred Hindu text \cite{gandhi2010bhagavad,hiltebeitel1976ritual}  that captures the essence of Hindu philosophy  \cite{dasgupta1975history}. The Mahabharata, one of the earliest and largest  epics written Sanskrit language using the style of narrative poetry, features the Bhagavad Gita as a chapter that captures a philosophical conversation between Lord Krishna and Arjuna about duty and ethics (Karma and Dharma) in the context of the Kurushetra war \cite{rajagopalachari1970mahabharata}.  The Bhagavad Gita shares the themes in a style similar to the Upanishads \cite{rao2002mind,gough2013philosophy}, a collection of Hindu philosophy  and sacred texts that predates and also influenced Greek philosophy \cite{lomperis1984hindu,scharfstein1998comparative}.  In the past, NLP has been utilized  to decipher and evaluate translations of major Hindu texts, including the Bhagavad Gita and Upanishads.  Chandra et al. \cite{Chandra2022-xi} used  NLP    to map the themes (topics) between the Bhagavad Gita and the Upanishads.  Since the translation of a poem can break not only the rhythm but also modify the essence of the text,   semantic and sentiment analysis can provide a means to evaluate the quality of translations. Hence, Chandra et al. \cite{Chandra2022-mh} implemented  a semantic and sentiment analysis on different translations of the Bhagavad Gita.

In May 2022, Google added support for the Sanskrit language in its addition of 24 languages \cite{sanskritgoogle} to \textit{Google Translate}, making a total of 133 languages worldwide. The team developed a new monolingual language model learning approach  for zero-resource translation \cite{siddhant2020leveraging}; i.e., translation for languages with no in-language parallel text and no language-specific translation examples \cite{zhang2016exploiting,zhao2015learning}. The model was trained to learn representations of under-resourced languages directly from monolingual text using the \textit{masked sequence-to-sequence} (MASS) task. MASS adopts the encoder-decoder framework for reconstructing a sentence fragment given the remaining part of the sentence. The  encoder takes a sentence with a randomly masked fragment (several consecutive tokens) as input, and its decoder tries to predict this masked fragment \cite{song2019mass}. 

% Analysis of quality 
In the past, some studies have analyzed the translation quality of Google Translate using computational models. Xiaoning et al. \cite{xiaoning2008using} used Google  Translate in cross-lingual information retrieval in order to translate the queries from English to Chinese, where a Kullback–Leibler (KL) divergence model was used for information retrieval. The authors indicate that Google Translate was chosen because of superior performance for named entity translation. Li et al. \cite{li2014comparison}  compared the   Google Translate with human (expert) translation, focusing on  Chinese to English  translation. The study reported that translation by Google Translate was highly correlated with the original text and the human expert. Rahimi et al. \cite{zand2017comparative}  studied the English-Persian translation of Google Translate.  Kalchbrenner et al. \cite{kalchbrenner2013recurrent} compared the accuracy of machine translation  and reported that  NMT improved the semantic aspects of the translation, despite some limitations. Abdur et al. \cite{md2019exploring} compared the English translations of Baidu and Google Translate and reported that there is a scope for improvement for both search engines, and one is not necessarily superior to the other. Patil et al. \cite{patil2014use} evaluated the accuracy of Google Translate in medical communication and found that Google Translate was not accurate when it comes to medical phrases and hence should not be blindly trusted. The authors also found  that European languages performed better than other languages and thus confirmed the presence of a translation bias. It is important to note that not many studies have evaluated the quality of translations of Google Translate for low-resource languages \cite{ranathunga2021neural}, i.e. languages with data scarcity such as Sanskrit. 
 
In this paper, we present a framework that evaluates the quality of Google Translate by focusing on the Sanskrit language. In this study, we first publish  a Sanskrit to English translation of the Bhagavad Gita using  Google Translate. Our proposed framework extends the methodology  Chandra et al. \cite{Chandra2022-mh} that compared translations of three different translations of the Bhagavad Gita  using  semantic and sentiment analysis. This study performs sentiment analysis via BERT and semantic analysis via a sentence embedding model to compare the Bhagavad Gita translation by Google Translate with translation by known experts. It further extracts keywords using KeyBERT to analyze the central themes in all three translations. Although the study's main aim is to evaluate the quality of Sanskrit translations by Google Translate, our framework is designed to be easily extended to other languages to evaluate Google Translate. Finally, we qualitatively evaluate selected Google Translate verses of the Bhagavad Gita with help of a Sanskrit translator. 

The rest of the paper is organized as follows. Section 2 provides an overview of the framework used for analysis. Section 3 presents the analysis of the results. Section 4 gives a detailed discussion, and Section 5 concludes the study. 

\section{Methodology}

\subsection{Data extraction and processing}

The Bhagavad Gita is divided into 18 chapters, each containing a sequence of questions and answers between Lord Krishna and Arjuna on various subjects, including the Karma philosophy. This organization is symbolic because the Mahabharata War lasted 18 days \cite{rajagopalachari1970mahabharata}. In this study, we use three different Bhagavad Gita translations (Mahatma Gandhi \cite{desai1946gospel}, Eknath Easwaran \cite{easwaran1985trans}, Sri Purohit Swami \cite{swami1937bhagavad}) to compare with the translation by  Google Translate. We selected the significant and prominent translations from different historical periods. In order to prevent any translation biases, we picked the translations where the translators were from a Hindu background. We processed the raw data from the three sets of translations using the  methodology  described by Chandra et al.. \cite{Chandra2022-mh} where semantic and sentiment analysis was implemented for comparing selected translations of the Bhagavad Gita. 

%We need a Table here that shows how the three translatoons, names and certain phrases such as Karma and Dharma were pre-processed along with different names of Krishna and Arjuna. This is given in previous BG paper - you can use same Table and provide more information as needed. 

\subsection{Google Translate}

Google Translate is a free-to-use web-based  translation tool  developed by Google in April 2006 \cite{googletranslate}. Google Translate is a multilingual NTM that translates texts, websites, and documents from a given language to a target language as specified by the user \cite{googletranslate3}. Even though Google Translate \cite{googletranslate,googletranslate2} has made significant advances in recent years (as of December 2022), it only  covers 133 written languages all over the world \cite{googletranslate4}, out of thousands of written and spoken languages. Note that Google Translate does not cater to automatic speech recognition i.e spoken languages, it is a text-based translation tool. There are challenges faced by Google Translate to data scarcity, the absence of digitized data for languages (low-resource languages), and the absence of translated texts. Hence, a roadblock exists in the development of functional translation models for low-resource languages such as   Sanskrit \cite{googletranslate5}. Note that Sanskrit is an ancient language used in Hindu texts; however,   only about 24,821  \cite{mccartney2022sanskrit}  Sanskrit speakers (based on 2011 census)   who are mostly in remote and rural communities of India. The lack of data is a problem for language identification models since it forces them to learn to translate from a limited monolingual text. To overcome these challenges, Google  made several modifications to the basic architecture of Google Translate which included \textit{back translation} overcome the lack of parallel (translated) data \cite{googletranslate5}. Back translation is  a localization quality control method where content is translated back to its original language and then compared to the source \cite{edunov2018understanding}.

 %The languages such as Malayalam had more than 34 million speakers in 2011 \cite{chandramouli2011census} as its the state language of Kerala, India.

\subsection{Google Translate - Bhagavad Gita}

\begin{figure*}[htbp]
\centering
    \includegraphics[scale=0.8]{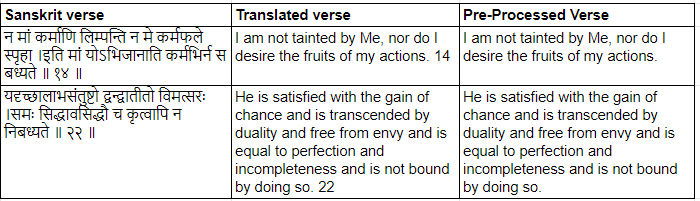}
    \caption{Original  Sanskrit script (Devanagari)  of the Bhagavad Gita with translation and further processing. }
    \label{fig:pre-pro}
\end{figure*}

We need to  translate all the 18 chapters of the Bhagavad Gita from  Sanskrit to  English using the Google Translate's application programmer interface (API). We extracted all the verses from the Sanskrit Bhagavad Gita   \footnote{\url{https://vedabase.io/en/library/bg/1/1/}} available on the \textit{Bhaktivedanta Vedabase} from  Swami Prabhupada who translated the Bhagavad Gita originally in 1968 \cite{prabhupada1972bhagavad}.  Note that the Sanskrit language is written using the Devanagari script \cite{bright1996devanagari} which can be directly used as an input to Google Translate API. We pre-processed the data with the following steps:

\begin{enumerate} 
\item Arranging the verses chapter-wise in different files
\item Removed verse numbering in Bhagavad Gita 
\item Converted verses  to a single line
\item Added the original Sanskrit version  to the file
\end{enumerate}
Figure \ref{fig:pre-pro} shows an example of the above pre-processing process.

Finally, we published the translation by Google Translate online via Github \footnote{\url{https://github.com/sydney-machine-learning/Google-Sanskrit-translate-evaluation/tree/main/BG-Google-Translated}}.

Figure \ref{fig:pre-pro} shows th Sanskrit script (Devanagari)  of the Bhagavad Gita with processed translation. 

\subsection{Sentiment and Semantic Analysis}

A word embedding is  used for the representation of words  from the text  in the form of a real-valued vector so that it can be used for processing by statistical and deep learning models \cite{li2018word}. The  real-valued vectors used for word embedding are selected to preserve the semantic and syntactic qualities of the word appearing in a text corpus \cite{ghannay2016word}.  A number of word embedding models exist that have certain strengths and weaknesses \cite{wang2019evaluating}.  Mikolov et al. \cite{mikolov} introduced \textit{Word2Vec} model in 2013 which has been widely used word embedding learned using a shallow neural network model. Thus, a simple cosine function can be used to test the level of similarity between two words. Cosine similarity is a metric to measure the text-similarity between two documents irrespective of their size. A word is represented into a vector form and the text documents are represented in n-dimensional vector space. The cosine similarity metric measures the cosine of the angle between two n-dimensional vectors projected in a multi-dimensional space.

 BERT is a transformer-based model that was introduced by Devlin et al. \cite{bert} in 2018 which comprises numerous bidirectional transformers that empower it to capture contextual information before and after a word. Note that BERT is a pre-trained model that has been trained   from unlabeled data extracted from the \textit{BooksCorpus} featuring 800 million words and the \textit{English Wikipedia} featuring  2,500 million  words. Since BERT gives context-enriched embedding, it outperformed traditional NLP models such as  Word2Vec on text processing tasks such as semantic and sentiment analysis \cite{9647258}. The word embeddings generated by Word2Vec are context-independent and cannot address the problem of polysemous words \cite{9647258}.  The embedding generated by BERT, on the other hand, is context-dependent, i.e., the same word can have multiple vector representations depending upon the context in which it is being used \cite{9647258}.

Sentiment analysis, also referred to as opinion mining and emotion analysis, identifies the emotional tone behind a body of text \cite{medhat2014sentiment}. Recent innovations involve machine learning and deep learning to mine text for sentiment, and subjective information \cite{zhang2018deep}. Sentiment analysis systems help in gathering insights from unorganized and unstructured text. It can be applied to varying scopes such as document, paragraph, sentence, and sub-sentence levels \cite{medhat2014sentiment}. There are primarily three different systems currently in use for performing sentiment analysis. Rule-based systems perform sentiment analysis based on predefined lexicon-based rules \cite{asghar2017lexicon}, whereas automatic systems learn from data with machine learning techniques \cite{mohammad2016sentiment}. A hybrid sentiment analysis, on the other hand, combines both approaches \cite{appel2016hybrid}. In addition to identifying sentiment, it can also extract the polarity (or the amount of positivity and negativity), subject and opinion holder within the text \cite{feldman2013techniques}. 

% paragraph about semantic analysis 
Semantic analysis, on the other hand, is the process of drawing meaning from text. Semantic analysis is key to contextualization that helps disambiguate language data so that text-based NLP applications can be more accurate \cite{goddard2011semantic}. It allows computers to understand and interpret sentences, paragraphs, or whole documents, by analyzing their grammatical structure and identifying relationships between individual words in a particular context \cite{nasukawa2003sentiment}. It’s the driving force behind machine learning tools such as  chatbots, search engines, and text analysis applications \cite{medhat2014sentiment}. By feeding semantically enhanced  algorithms with samples of text, NLP methods can make accurate predictions based on past observations \cite{maulud2021state}.

\subsection{Framework} 

\begin{figure*}[htbp!]
\centering
    \includegraphics[scale=0.1]{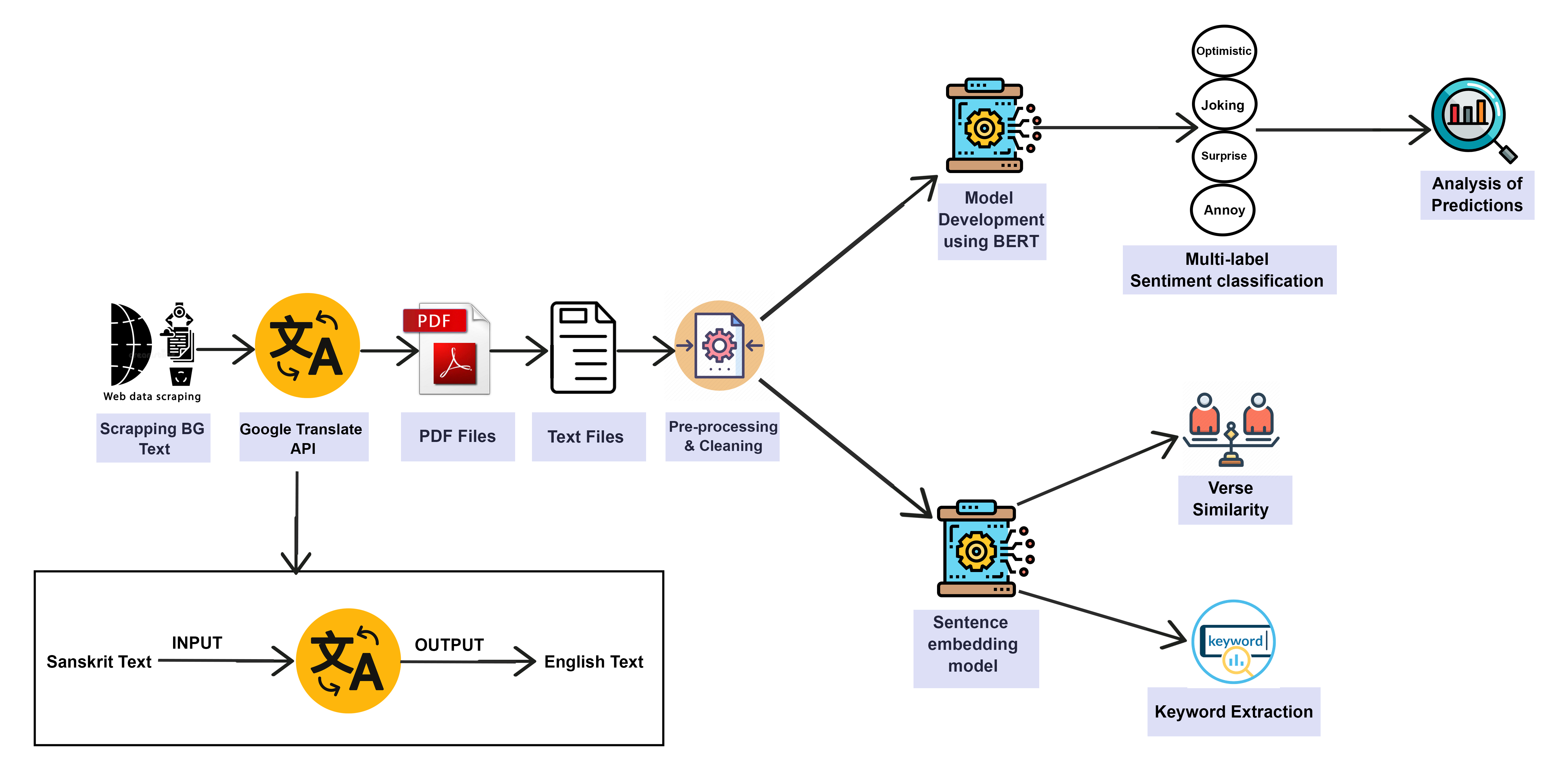}
    \caption{Framework showing major components that include using Google Translate for translating the original Sanskrit version of the Bhagavad Gita to English. We use semantic and sentiment analysis to compare the Google Translate version to expert translations from the literature that includes translations by Mahatma Gandhi and Eknath Easwaren. }
    \label{fig:frame}
\end{figure*}

We present a framework that  compares translations and implements sentiment and semantic analysis, adopted from Chandra and Kulkarni \cite{Chandra2022-mh} (Figure \ref{fig:frame}). We utilize this framework by comparing the Bhagavad Gita by Google Translate  with three expert-based translations. Our framework provides further insights into the various themes discussed by these different translations.
We extracted the Bhagavad Gita Sanskrit slokas (verses)   from Bhaktivedanta Vedabase \footnote{\url{https://vedabase.io/en/library/bg/1/1/}} using web data scrapping process. We provide this text as input to the Google Translate  API, which gives the corresponding English translated text as an output. We then store the output in printable document format (PDF) format. Afterwards, we convert the the PDF files  to text files for  pre-processing and cleaning of text, where we remove  verse numbers, symbols, etc. Our framework implements  the BERT-base model for sentiment analysis by predicting the sentiments of different verses of the four translations. We use multi-label sentiment classification in our framework  where  a verse can be both empathetic and optimistic, simultaneously. We then train our sentiment analysis  component in the framework using an expert-labeled SenWave dataset  \cite{yang2020senwave} which features 10 different sentiments labeled by a group of 50 experts for 10,000 tweets worldwide during the COVID-19 pandemic. We fine-tuned (trained) the BERT-base sentiment analysis model using the SenWave dataset so that it can recognize the respective sentiments in a multi-label setting, originally used for COVID-19 sentiment analysis \cite{chandra2021covid} and  for Bhagavad Gita sentiment analsyus \cite{Chandra2022-mh}. The conventional sentiment polarity score has ambiguity due to varied  expressions that feature metaphor, humor, and   expressions hard for machines to understand. Hence,  multi-label sentiment classification provides further insights. We compare verse-by-verse and chapter-by-chapter sentiments of the chosen translations as shown in  Figure \ref{fig:frame}.

Furthermore, we perform semantic analysis to reveal the variations in the   translations so that we get an indication of how similar or different the expert-based translations are when compared to the Google Translate version of the Bhagavad Gita. We perform semantic analysis through a sentence embedding model  (MPNet \cite{mpnet}) which is based on the BERT model as shown in the framework (Figure \ref{fig:frame}). MPNet sentence embedding model generates high-quality embedding for our encoded verses o the Bhagavad Gita. We use the uniform manifold approximation and projection (UMAP) \cite{UMAP} dimensionality reduction technique to visualize the high-dimensional vectors. We investigate the nature in terms of the similarity of the chapters based on data  visualization through the plot of the first two dimensions obtained from UMAP. 

Furthermore, we extract keywords from the text to examine the major topics using KeyBERT which provides the keyboard that describes significant themes (Figure \ref{fig:frame}). We note that various other techniques can be used, such as \textit{rapid automatic keyword extraction} (RAKE)\cite{RAKE}, \textit{yet another keyword extractor} (YAKE)\cite{Yake}, and term frequency-inverse document frequency (TF-IDF) \cite{TF-IDF}. However, these are based on statistical characteristics, unlike KeyBERT which is based on the semantic similarity of the text. Hence, we use KeyBERT as it considers the text's semantic aspects. 

%xxx

\subsection{Experimental setup}

We train the BERT (base) model on the SenWave dataset by pre-processing the tweets as done by Chandra and Kulkarni \cite{Chandra2022-mh}.
We utilised  the trained models from previous study about sentiment analysis of the Bhagavad Gita \cite{Chandra2022-mh} via  the GitHub repository \footnote{https://github.com/sydney-machine-learning/sentimentanalysis\_bhagavadgita}. The SenWave dataset consists of  10,000 Tweets that  were labeled according to 10 different sentiments by human experts. There is an additional  label related to the "official report" related to COVID-19 which we deleted in data processing.

% The tweets express emotions such as  "optimistic," "pessimistic," "anxious," or "thankful" and a tweet may be categorized as both "optimistic" and "anxious" simultaneously;  hence, this is a multi-label sentiment classification problem. 

%need more technical information, such as hyperparameters, learning rate - mention Adam optimer if used, any archtectural properties of BERT such as numver of neurals or parameters, training time in terms of epochs etc. Mention how much time it takes to train the model - in minutes and also to run the model on one translation.  

\section{Results}

\subsection{Data Analysis}
 
\begin{figure*}
     \centering
     \begin{subfigure}[b]{.45\linewidth}
         \centering
         \includegraphics[width=\linewidth]{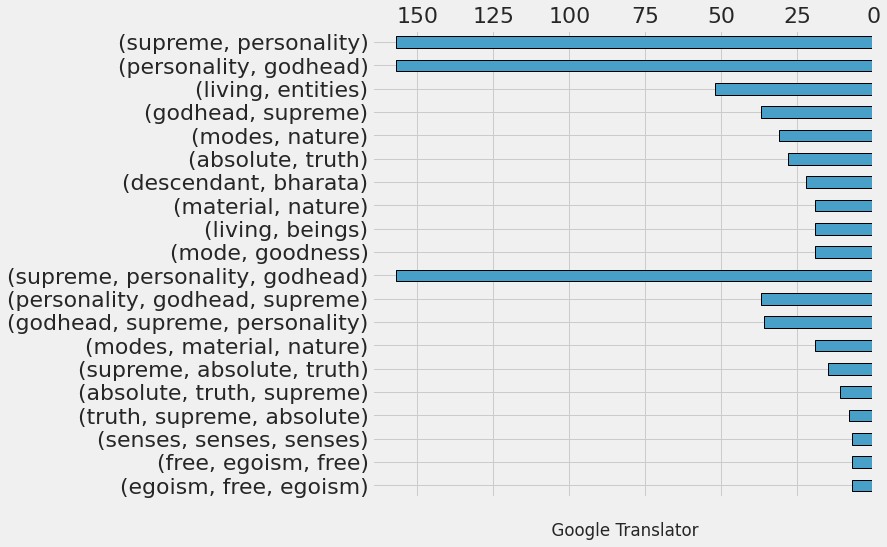}
          \caption{The Bhagavad Gita translations by Google Translate.}
          \label{fig:google_bi_tri}
     \end{subfigure}
     \hfill
     \begin{subfigure}[b]{.45\linewidth}
         \centering
         \includegraphics[width=\linewidth]{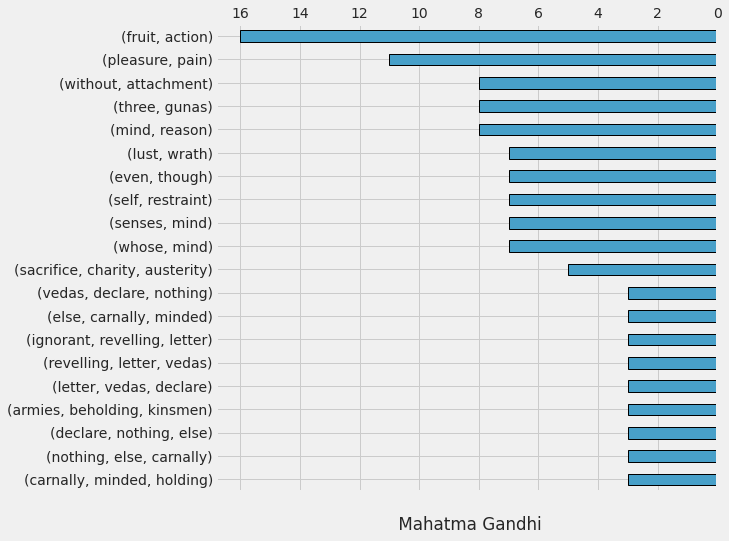}
          \caption{The Bhagavad Gita translation by Mahatma Gandhi.}
          \label{fig:gandhi_bi_tri}
     \end{subfigure}
     \hfill
     \begin{subfigure}[b]{.45\linewidth}
         \centering
         \includegraphics[width=\linewidth]{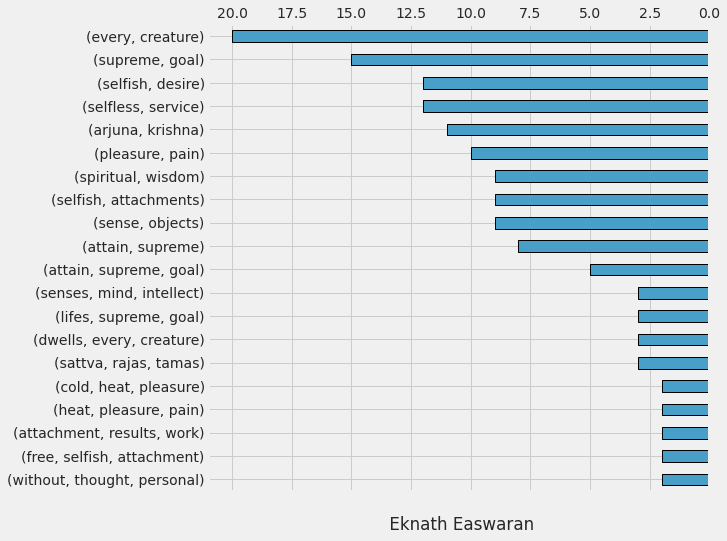}
          \caption{The Bhagavad Gita translation by Eknath Easwaran.}
          \label{fig:Eknath_bi_tri}
     \end{subfigure}
     \hfill
    \begin{subfigure}[b]{.45\linewidth}
         \centering
        \includegraphics[width=\linewidth]{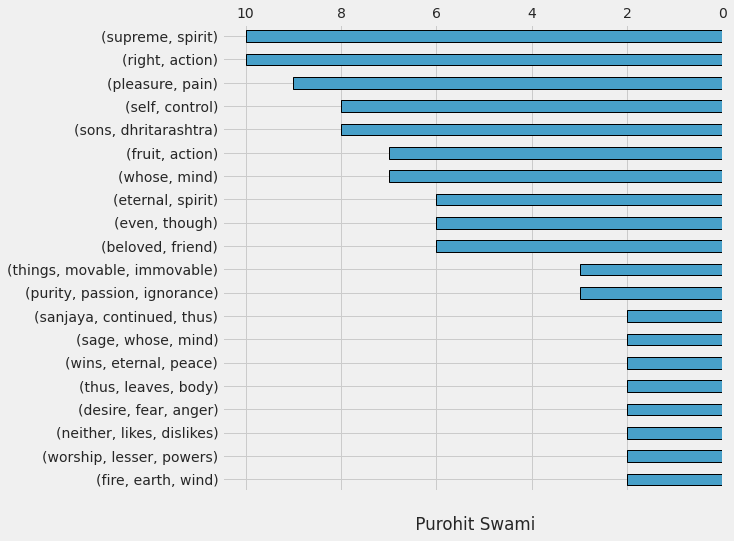}
         \caption{The Bhagavad Gita translation by Sri Purohit Swami.}
        \label{fig:Purohit_bi_tri}
     \end{subfigure}
        \caption{Visualisations of top 10 bi-grams and tri-grams for different Bhagavad Gita translations.}
        \label{fig:bitrigram}
\end{figure*}

The n-gram \cite{robertson1998applications} in NLP provides a statistical overview of a text  through a continuous sequence of words and elements. We first present the top-ten bi-gram and tri-grams along with top-twenty optimistic and pessimistic sentiments bi-grams and tri-grams in the text for the different translations as shown in Figure \ref{fig:bitrigram}.

We now analyze the bi-grams and tri-grams of Google Translate   version and Eknath Easwaran's  version. We only compare with Eknath Easwaren for simplicity, the comparison with Mahatma Gandhi and Sri Purohit Swami's translations can also be done in a similar manner. We observe that the concept of a "supreme spirit," or the "Atman" is mentioned in both translations, but the path taken to achieve this realization varies between them. The Google Translate version in Figure \ref{fig:google_bi_tri} features the tri-gram [absolute, truth, supreme], thus reflecting a path of absolute truth. Eknath Easwaran's translation \ref{fig:Eknath_bi_tri} features bi-grams [supreme, goal], [selfless, service], and [selfish, attachment], thus stressing the importance of selfless service devoid of selfish attachments and desires.  It is interesting to note that Chapter 3 is titled 'Selfless Service' by Eknath Easwaran's translation.

Furthermore, we observe from Figure \ref{fig:bitrigram} that the top three bi-grams and ti-grams are different for the two translations. Both translations have  used different words to describe similar themes. Google Translate  features [supreme, personality], [personality, godhead], and [living, entities] as the top 3 bi-grams and different permutations of [supreme, personality, godhead] as the top 3 tri-grams. Mahatma Gandhi's translation features [fruit, action], [pleasure, pain], and [without, attachment] as the top three bi-grams and [sacrifice, charity, austerity], [vedas, declare, nothing] and [else, carnality, minded] as the top three tri-grams.
Eknath Easwaran's translation features [every, creation], [supreme, goal], and [selfish, desire] as the top three bi-gram and [attain, supreme, goal], [senses, mind, intellect] and [dwells, every, creation] as the top three tri-grams. Shri Purohit Swami's translation features [supreme, spirit], [right, action], and [pleasure, pain] as the top three bi-grams and [thing, movable, immovable], [purity, passion, ignorance] and [sanjaya, continued, thus] as the top three tri-grams. 
Hence, a mere word-to-word comparison through bi-grams and tri-grams reflects differences in the translations.

\subsection{Sentiment Analysis}

Next, we use the BERT model for verse-by-verse sentiment analysis of the respective Bhagavad Gita translations.

\begin{figure}[htbp]
\centering
    \includegraphics[scale=0.27]{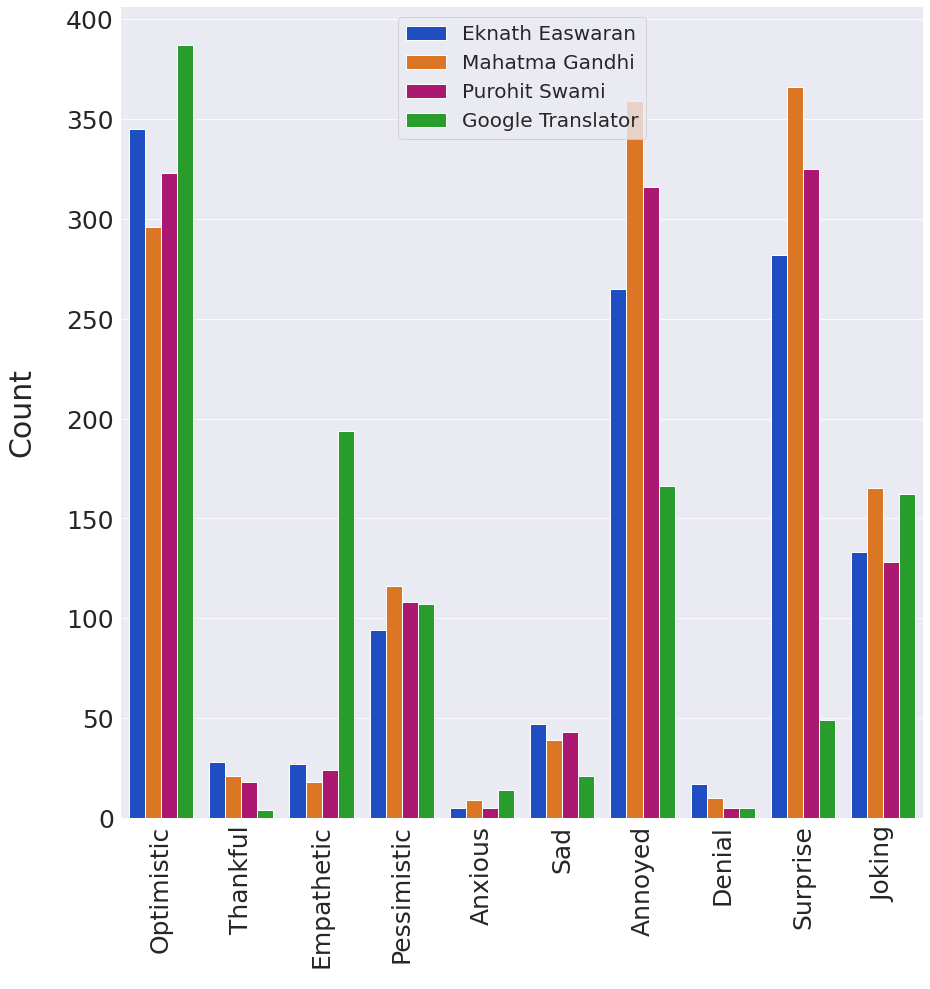}
    \caption{Cumulative Sentiments of the chapters}
    \label{fig:cumul_sa}
\end{figure}

\begin{figure*}
     \centering
     \begin{subfigure}[b]{.30\linewidth}
         \centering
         \includegraphics[width=\linewidth]{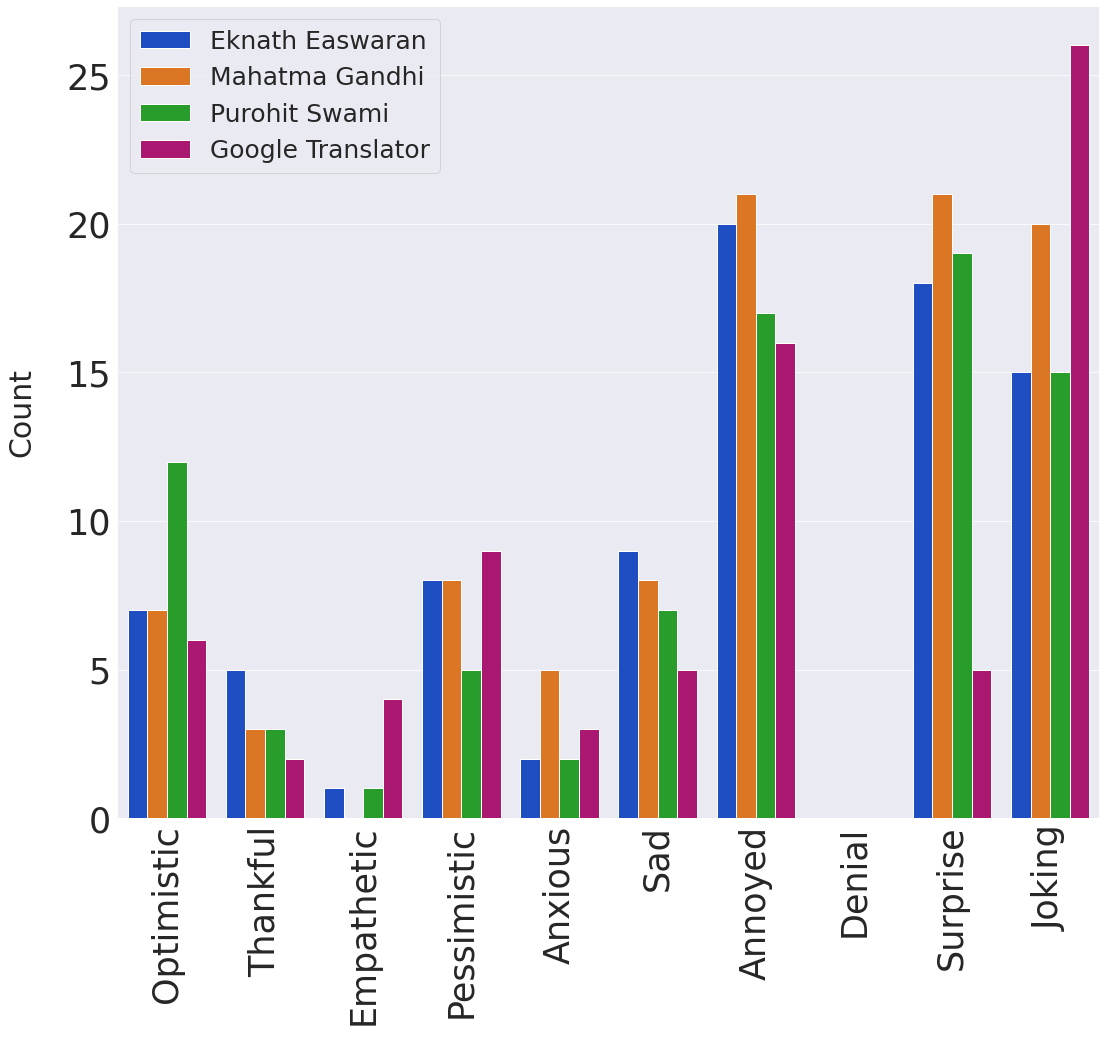}
        \caption{Chapter-1}
        \label{fig:chap1_sa}
     \end{subfigure}
     \hfill
     \begin{subfigure}[b]{.30\linewidth}
         \centering
         \includegraphics[width=\linewidth]{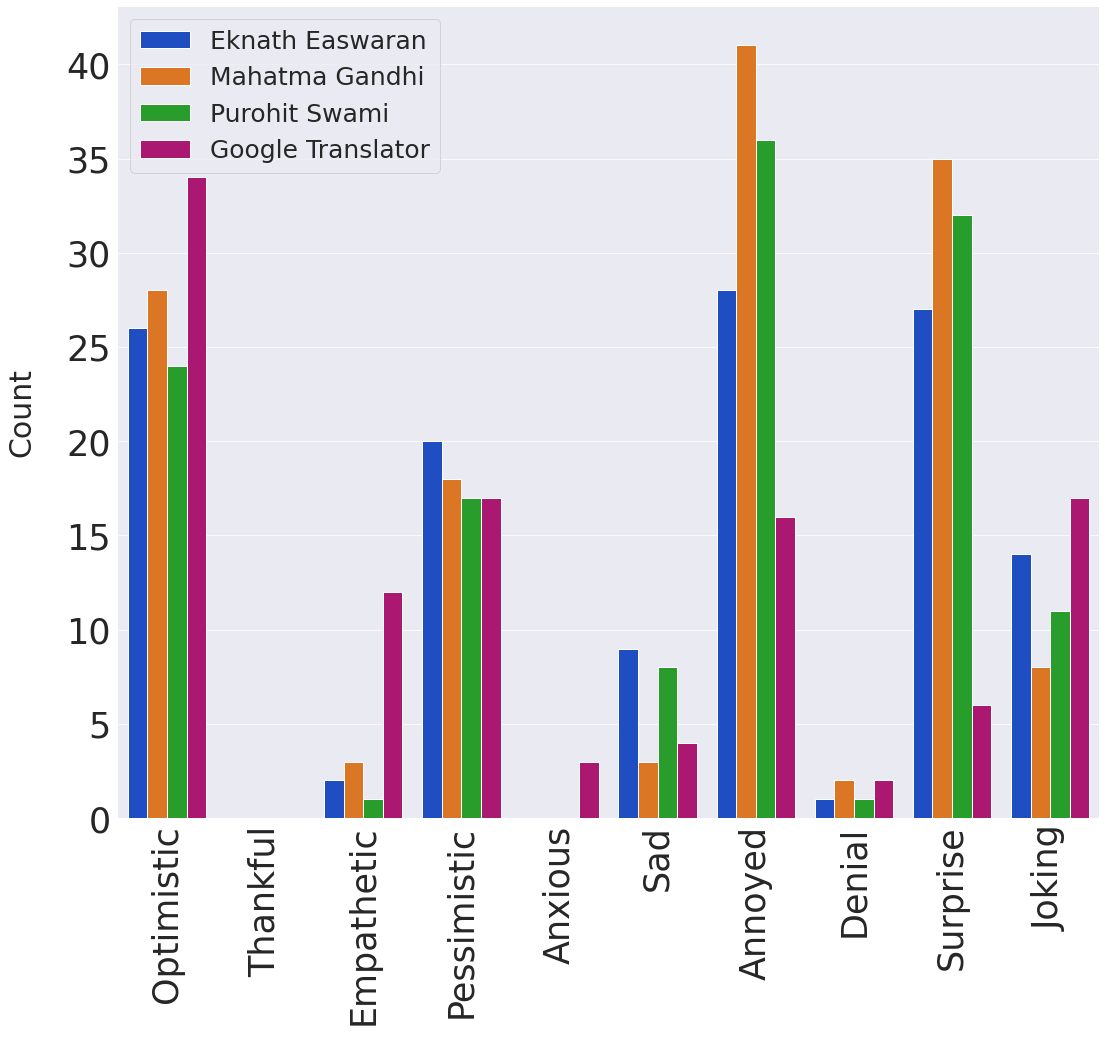}
        \caption{Chapter-2}
        \label{fig:chap2_sa}
     \end{subfigure}
     \hfill
     \begin{subfigure}[b]{.30\linewidth}
         \centering
         \includegraphics[width=\linewidth]{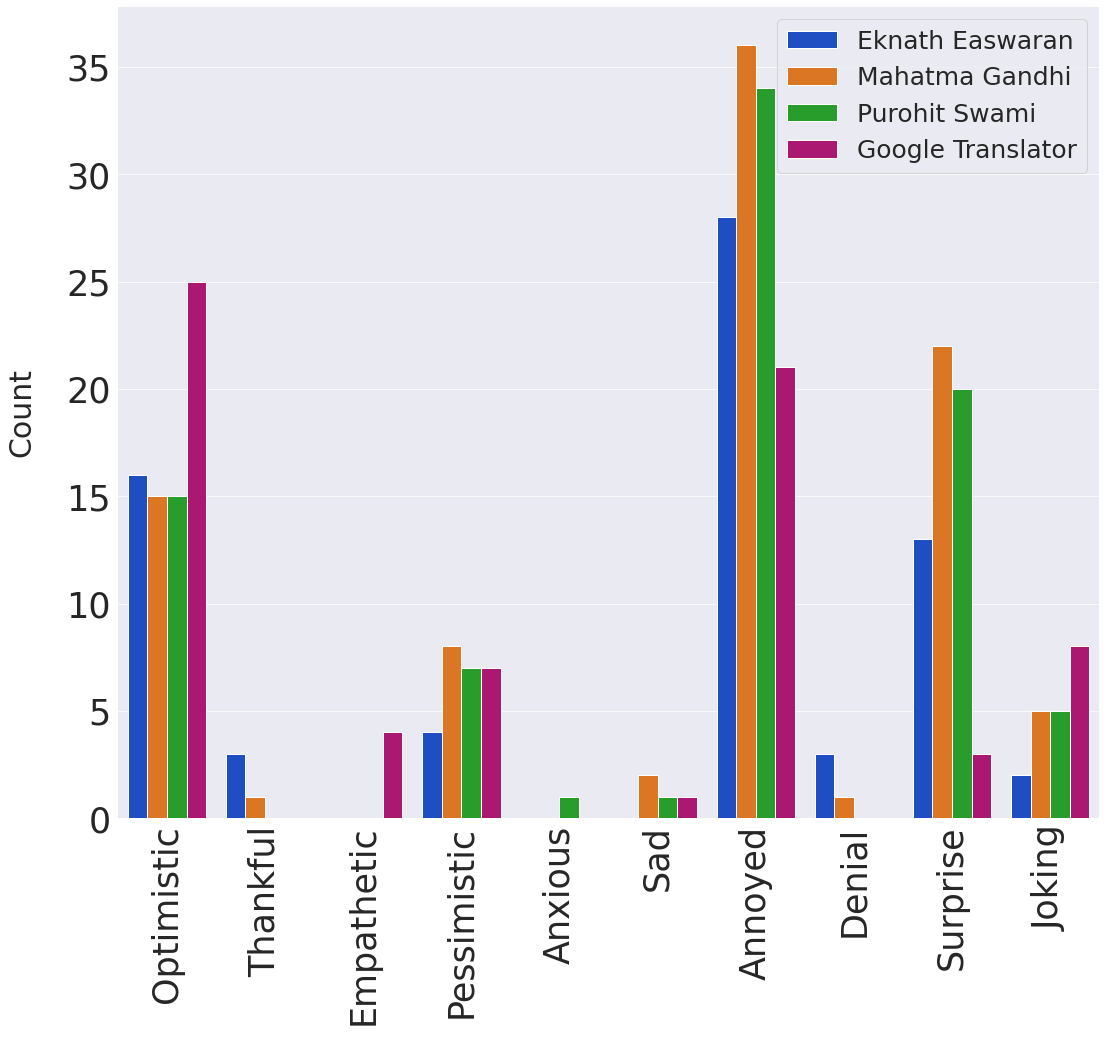}
        \caption{Chapter-3}
        \label{fig:chap3_sa}
     \end{subfigure}
     \hfill
     \begin{subfigure}[b]{.30\linewidth}
         \centering
         \includegraphics[width=\linewidth]{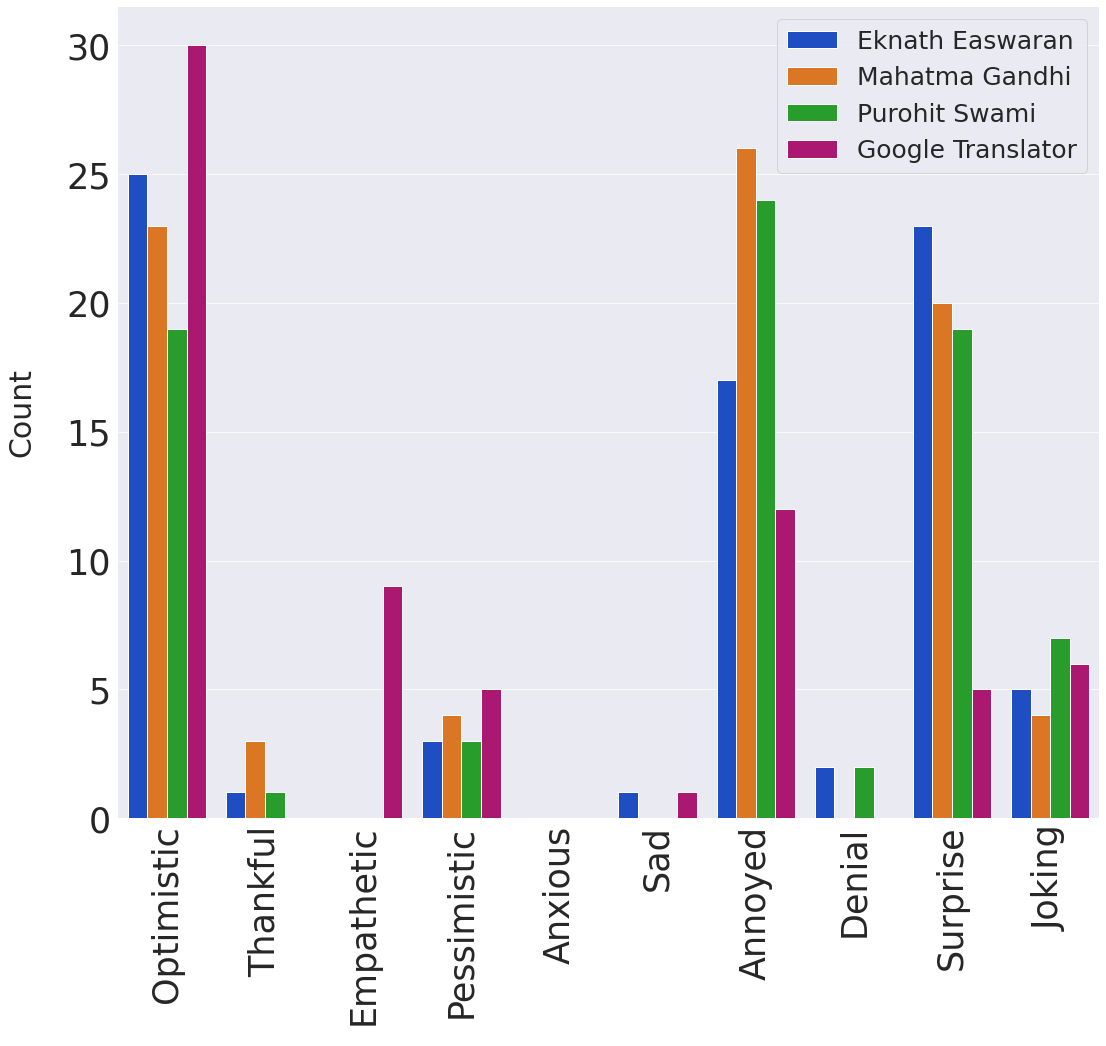}
        \caption{Chapter-4}
        \label{fig:chap4_sa}
     \end{subfigure}
     \hfill
     \begin{subfigure}[b]{.30\linewidth}
         \centering
         \includegraphics[width=\linewidth]{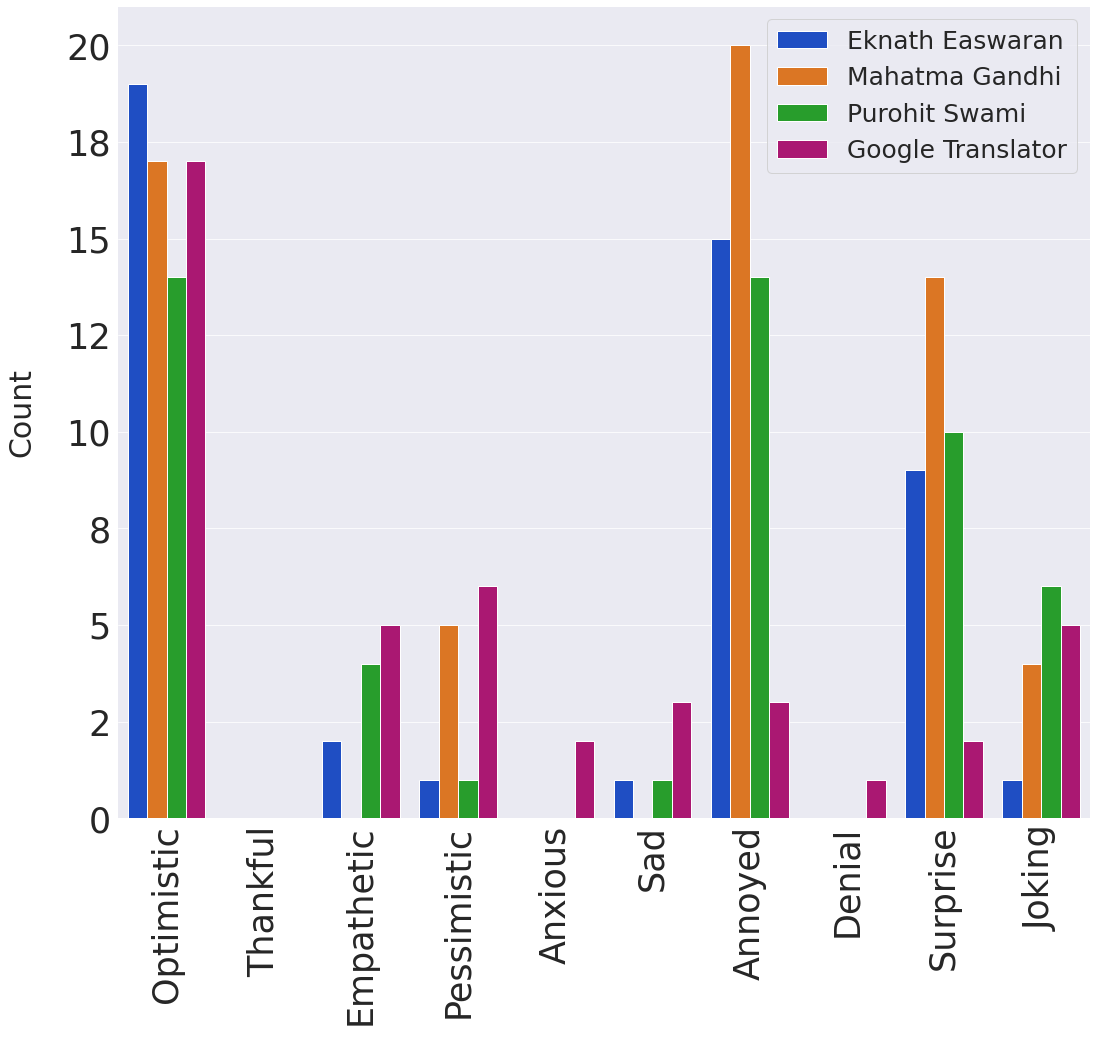}
        \caption{Chapter-5}
        \label{fig:chap5_sa}
     \end{subfigure}
     \hfill
     \begin{subfigure}[b]{.30\linewidth}
         \centering
         \includegraphics[width=\linewidth]{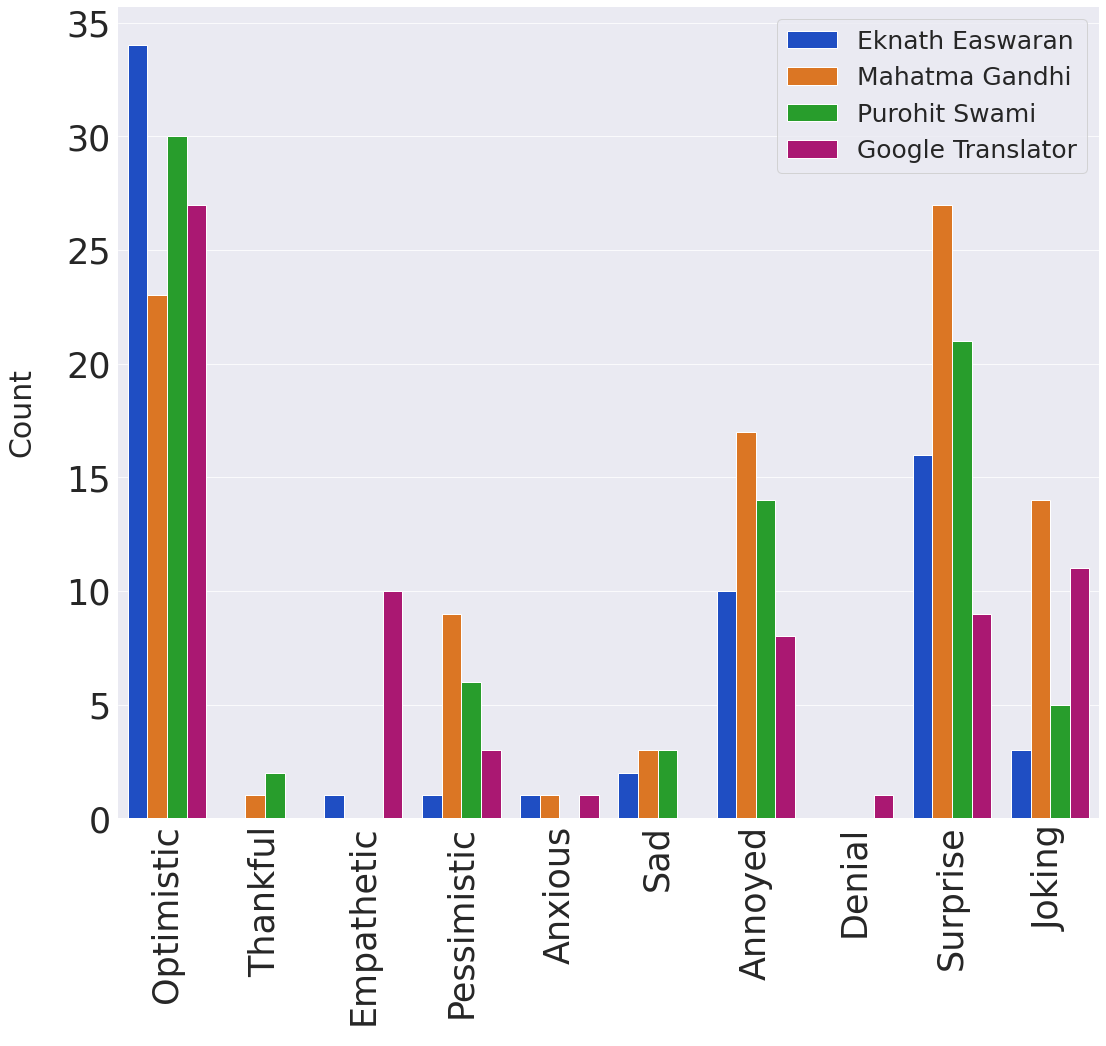}
    \caption{Chapter-6}
    \label{fig:chap6_sa}
     \end{subfigure}
     \hfill
     \begin{subfigure}[b]{.30\linewidth}
         \centering
         \includegraphics[width=\linewidth]{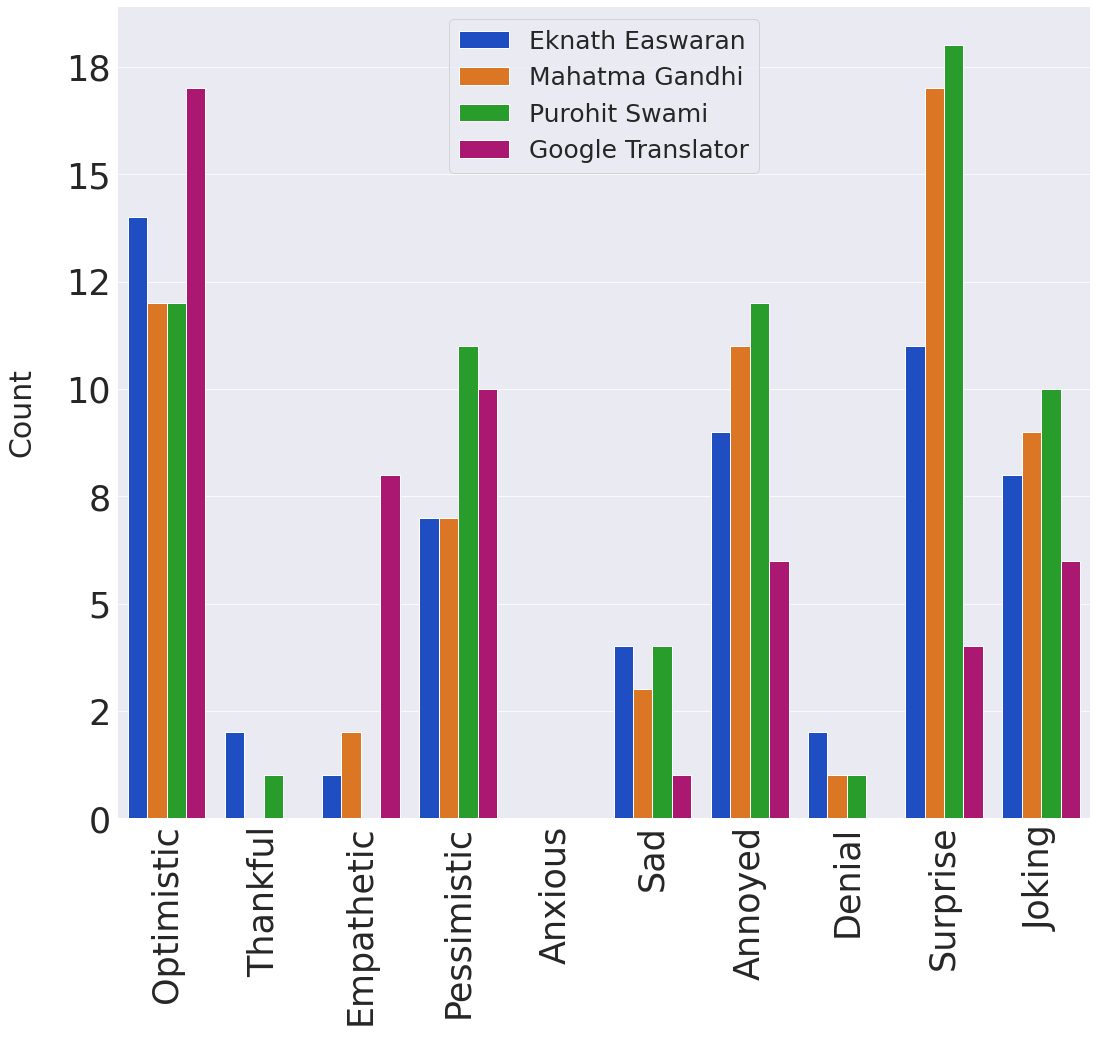}
    \caption{Chapter-7}
    \label{fig:chap7_sa}
     \end{subfigure}
     \hfill
     \begin{subfigure}[b]{.30\linewidth}
         \centering
         \includegraphics[width=\linewidth]{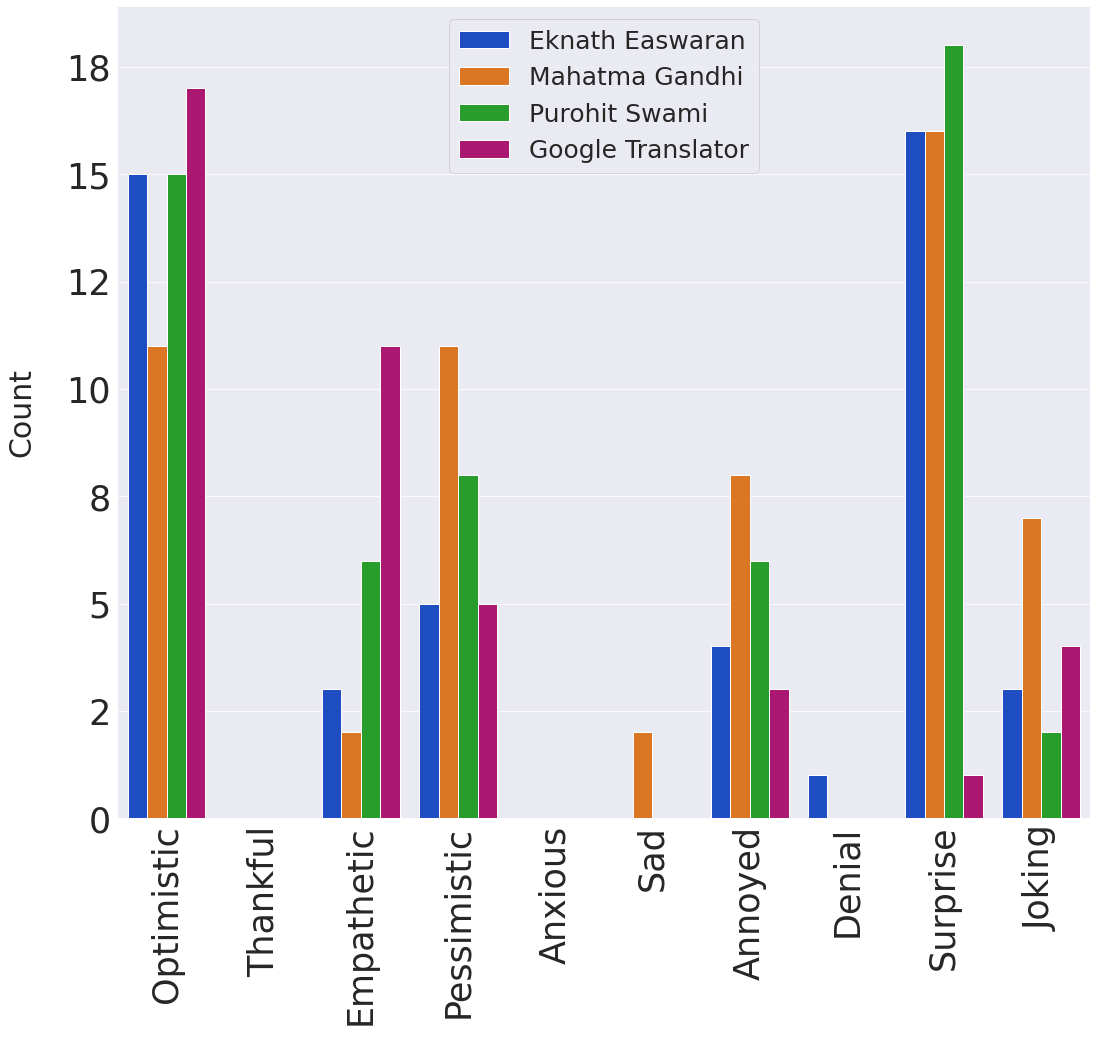}
    \caption{Chapter-8}
    \label{fig:chap8_sa}
     \end{subfigure}
     \hfill
     \begin{subfigure}[b]{.30\linewidth}
         \centering
         \includegraphics[width=\linewidth]{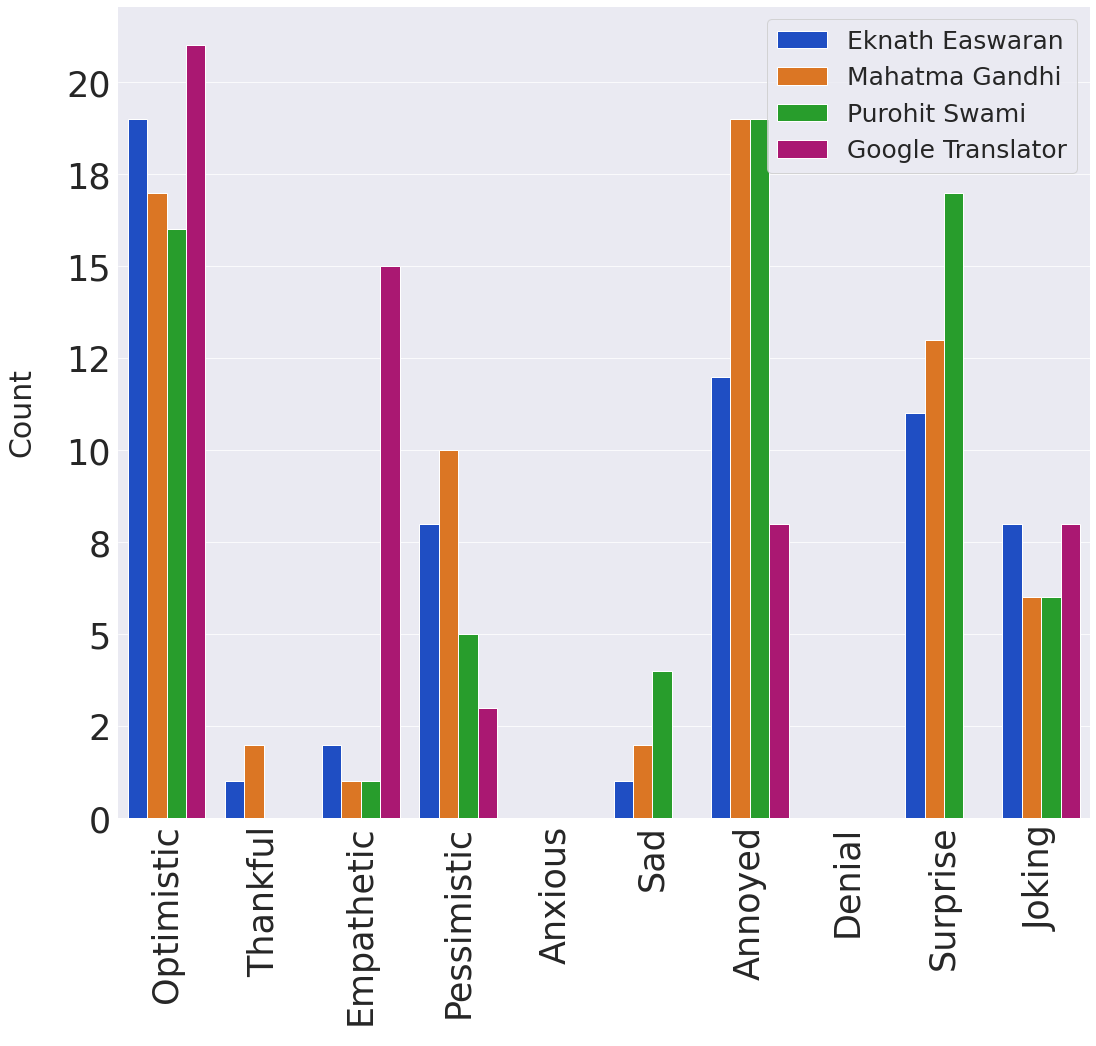}
    \caption{Chapter-9}
    \label{fig:chap9_sa}
     \end{subfigure}
        \caption{Chapter-wise Sentiment Analysis of Chapter 1 - Chapter 9.}
        \label{fig:chap_1_9}
\end{figure*}

\begin{figure*}
     \centering
     \begin{subfigure}[b]{.30\linewidth}
         \centering
         \includegraphics[width=\linewidth]{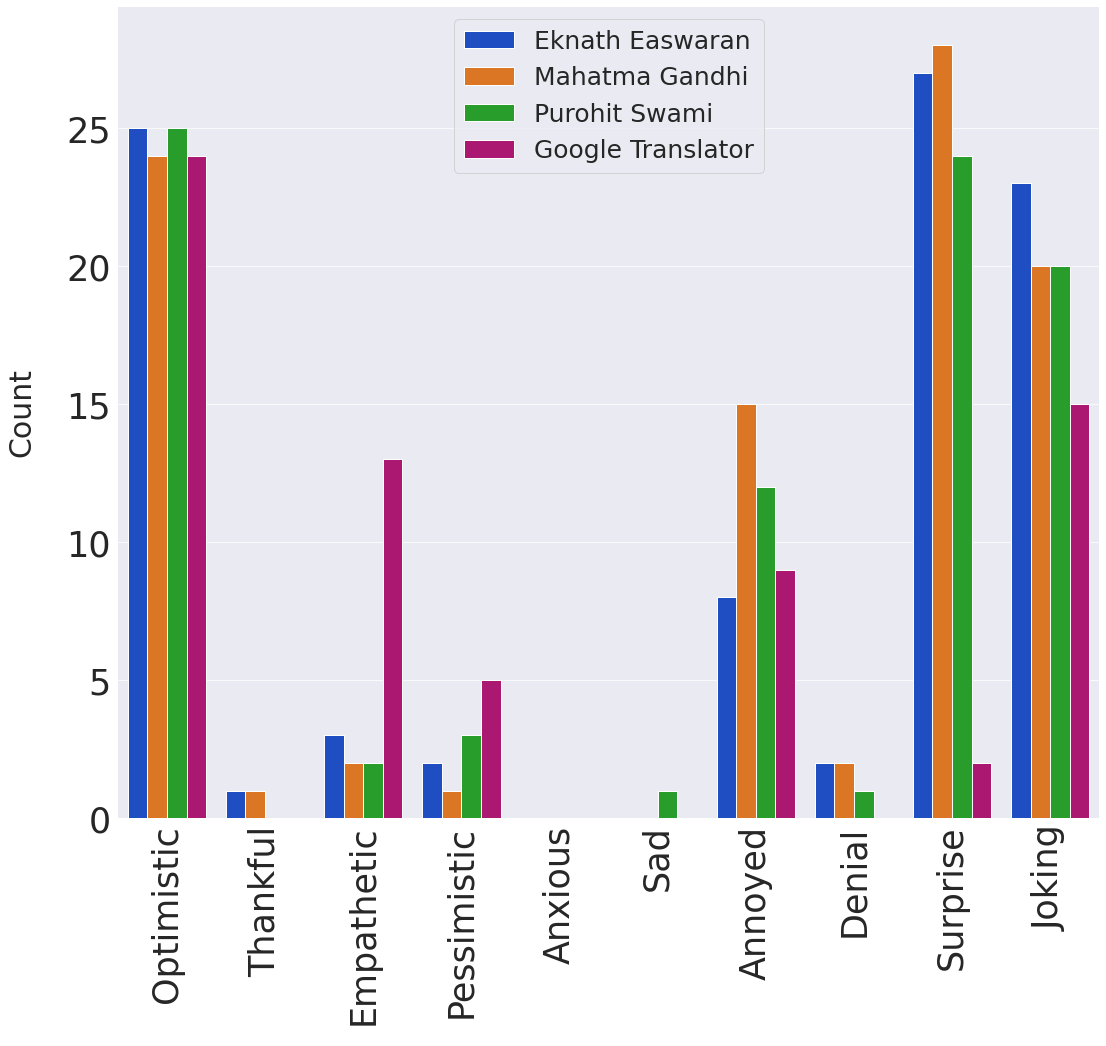}
    \caption{Chapter-10}
    \label{fig:chap10_sa}
     \end{subfigure}
     \hfill
     \begin{subfigure}[b]{.30\linewidth}
         \centering
         \includegraphics[width=\linewidth]{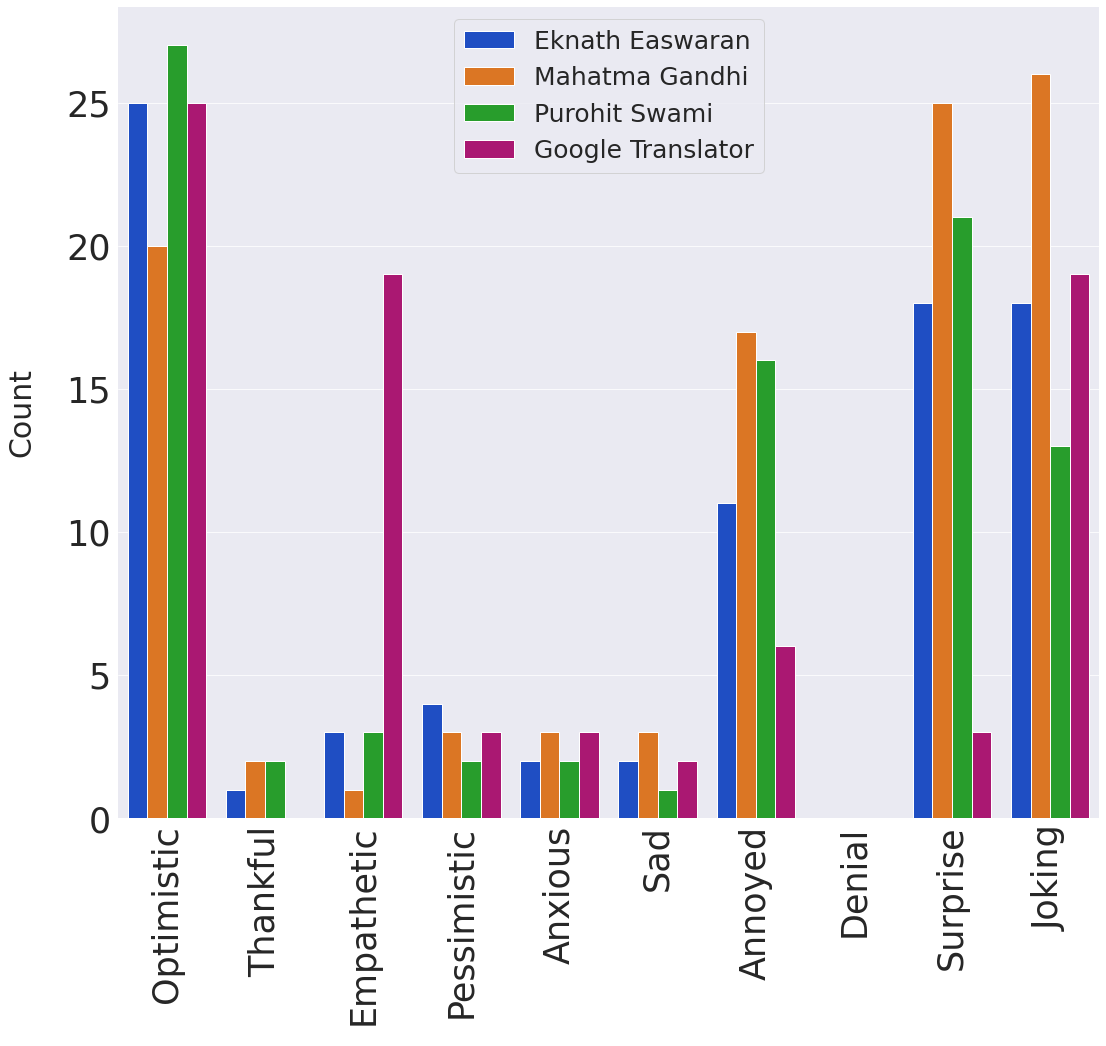}
    \caption{Chapter-11}
    \label{fig:chap11_sa}
     \end{subfigure}
     \hfill
     \begin{subfigure}[b]{.30\linewidth}
         \centering
         \includegraphics[width=\linewidth]{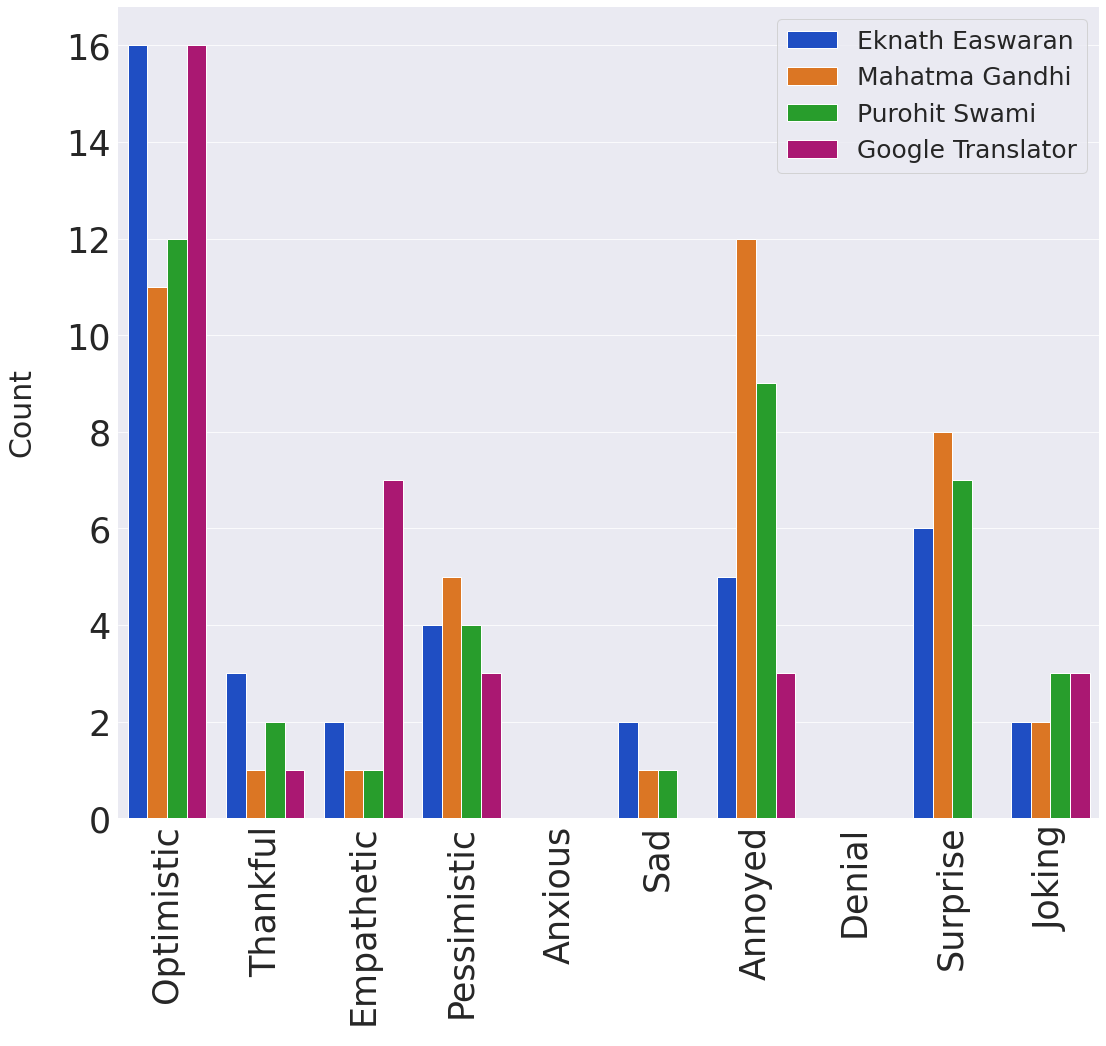}
    \caption{Chapter-12}
    \label{fig:chap11_sa}
     \end{subfigure}
     \hfill
     \begin{subfigure}[b]{.30\linewidth}
         \centering
         \includegraphics[width=\linewidth]{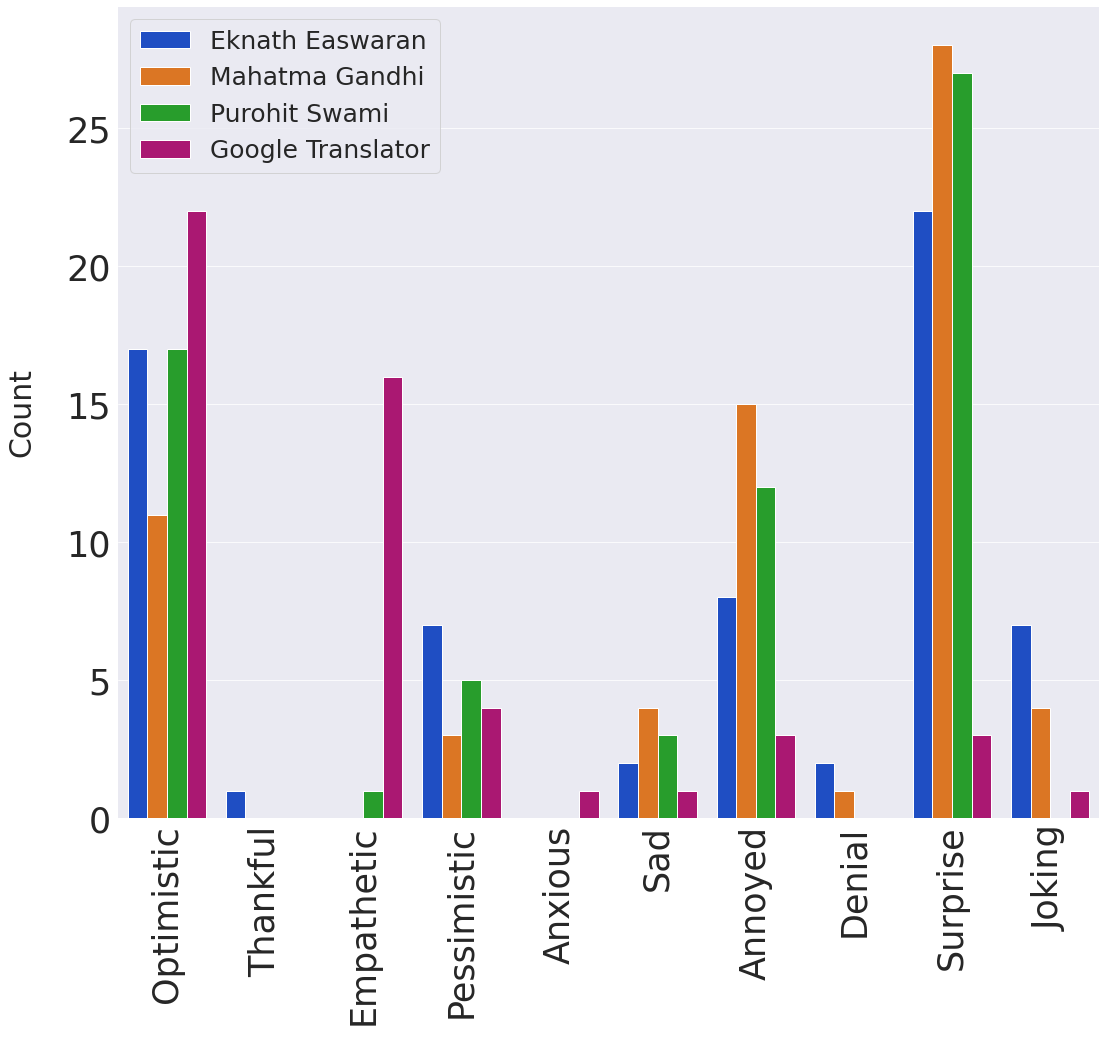}
    \caption{Chapter-13}
    \label{fig:chap13_sa}
     \end{subfigure}
     \hfill
     \begin{subfigure}[b]{.30\linewidth}
         \centering
         \includegraphics[width=\linewidth]{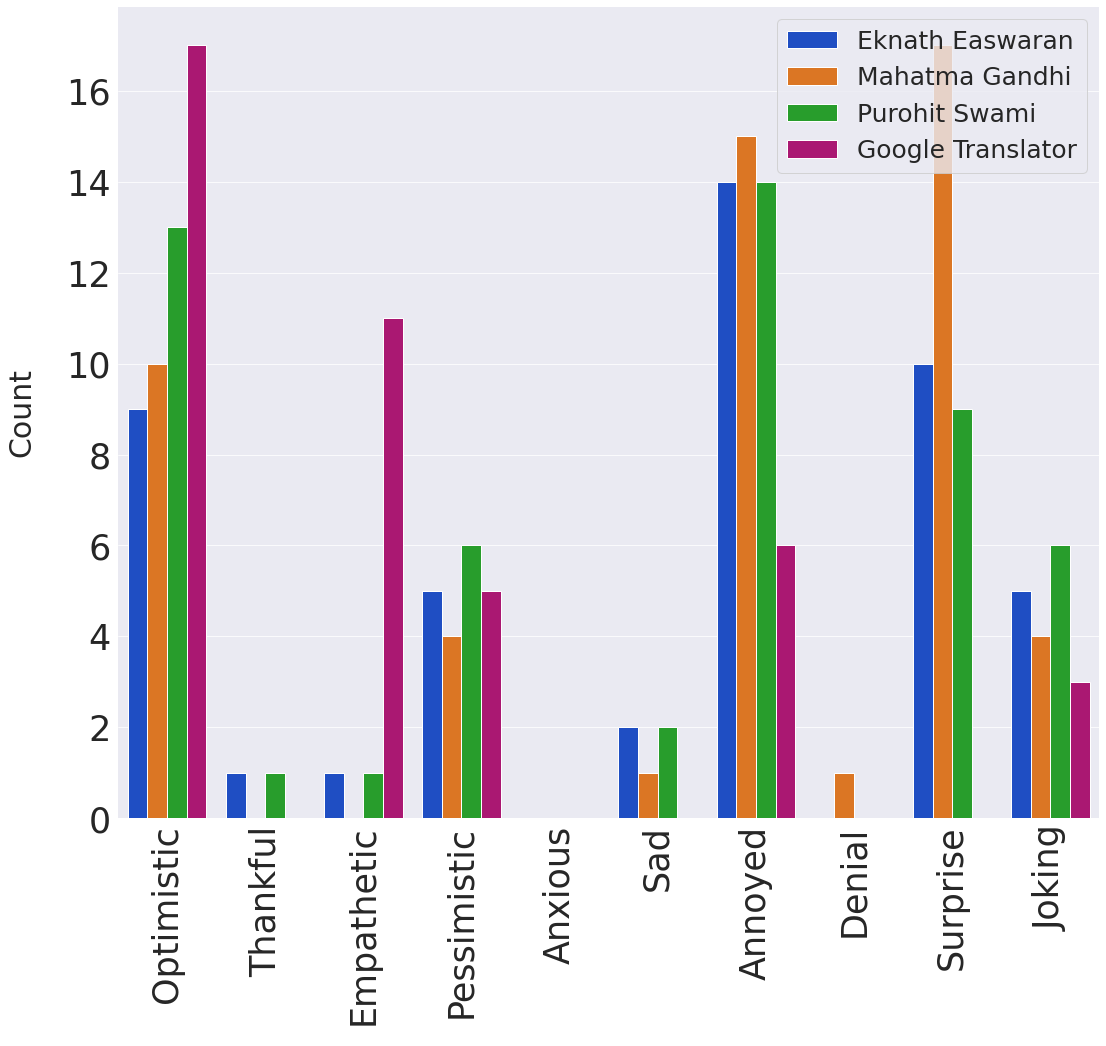}
    \caption{Chapter-14}
    \label{fig:chap14_sa}
     \end{subfigure}
     \hfill
     \begin{subfigure}[b]{.30\linewidth}
         \centering
         \includegraphics[width=\linewidth]{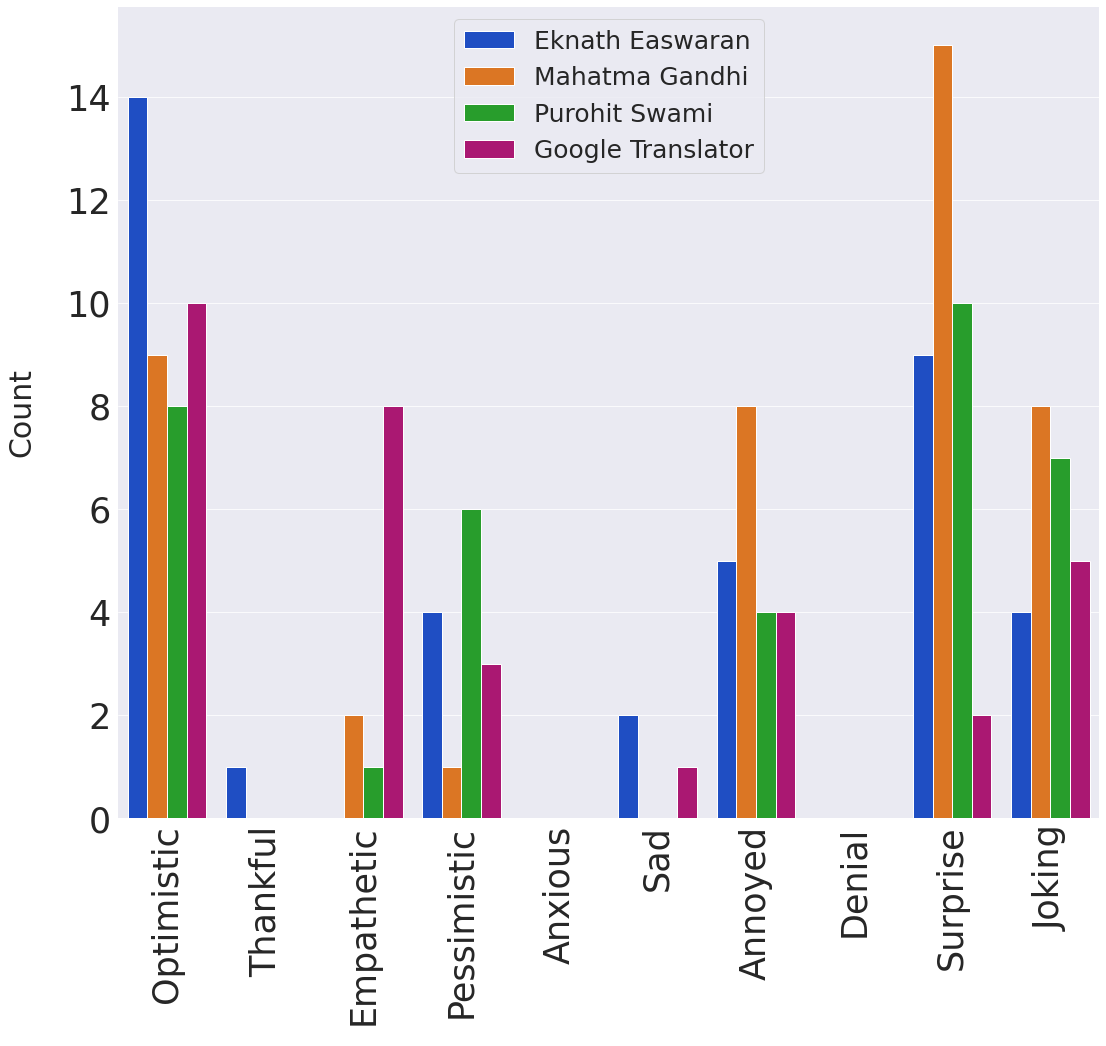}
    \caption{Chapter-15}
    \label{fig:chap15_sa}
     \end{subfigure}
     \hfill
     \begin{subfigure}[b]{.30\linewidth}
         \centering
         \includegraphics[width=\linewidth]{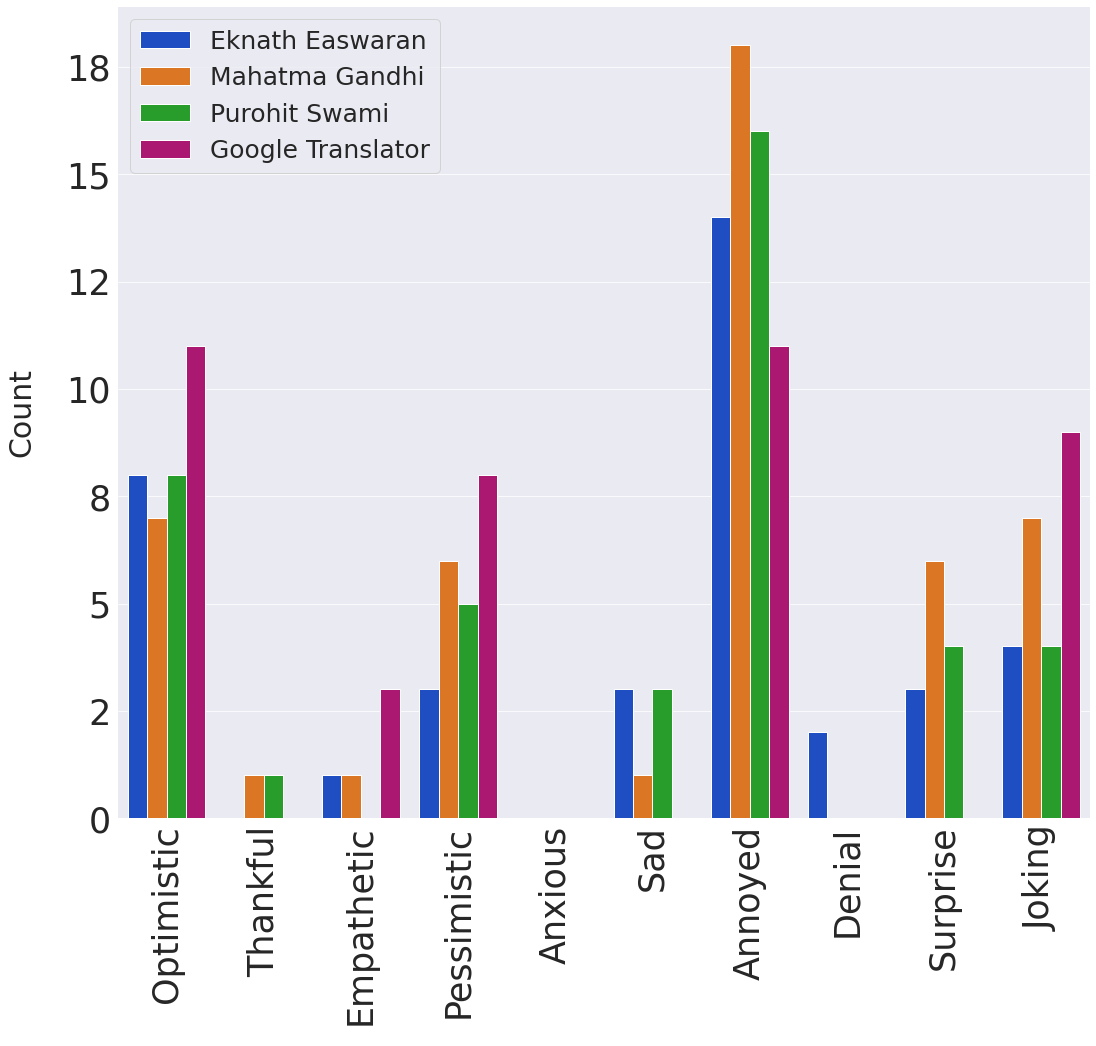}
    \caption{Chapter-16}
    \label{fig:chap16_sa}
     \end{subfigure}
     \hfill
     \begin{subfigure}[b]{.30\linewidth}
         \centering
         \includegraphics[width=\linewidth]{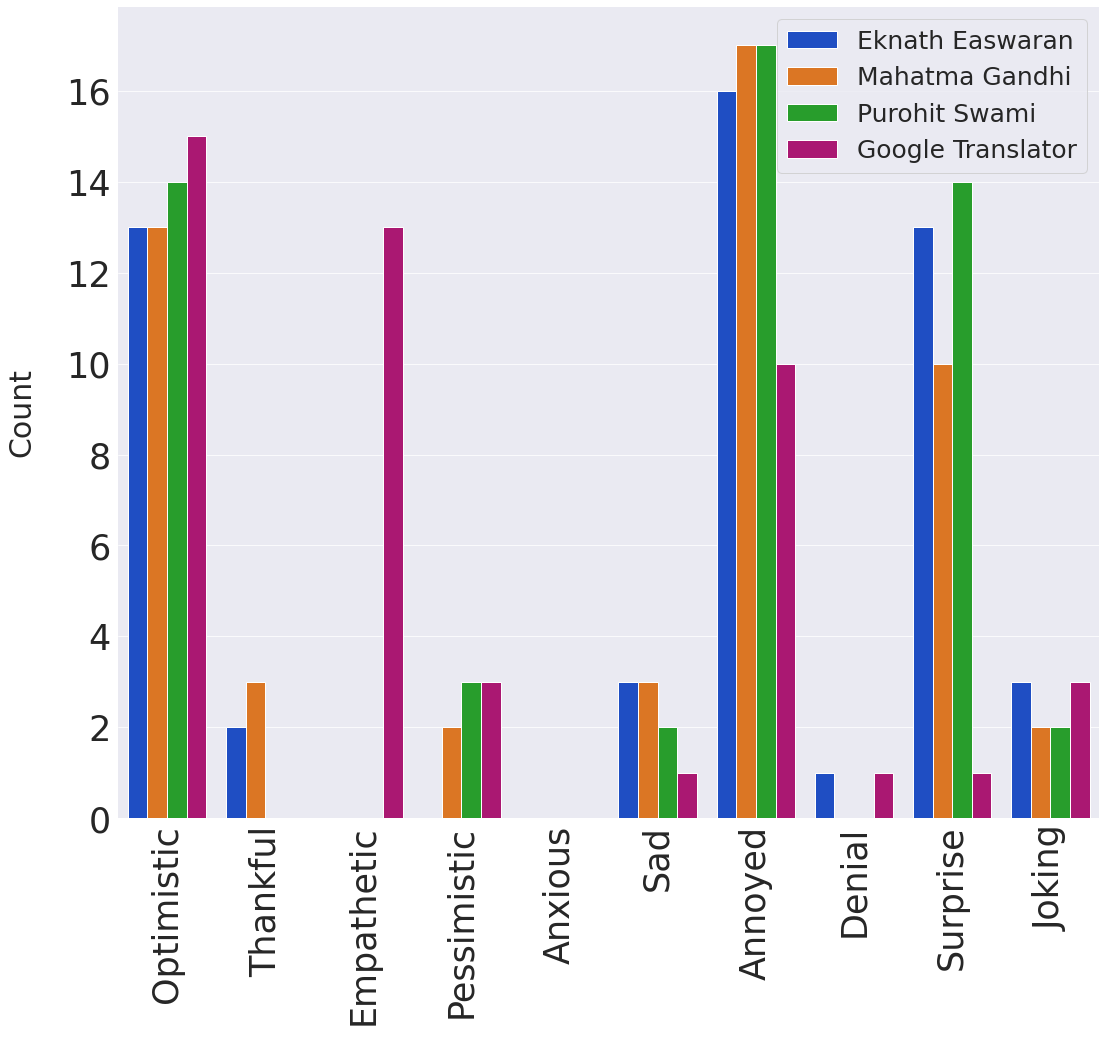}
    \caption{Chapter-17}
    \label{fig:chap17_sa}
     \end{subfigure}
     \hfill
     \begin{subfigure}[b]{.30\linewidth}
         \centering
         \includegraphics[width=\linewidth]{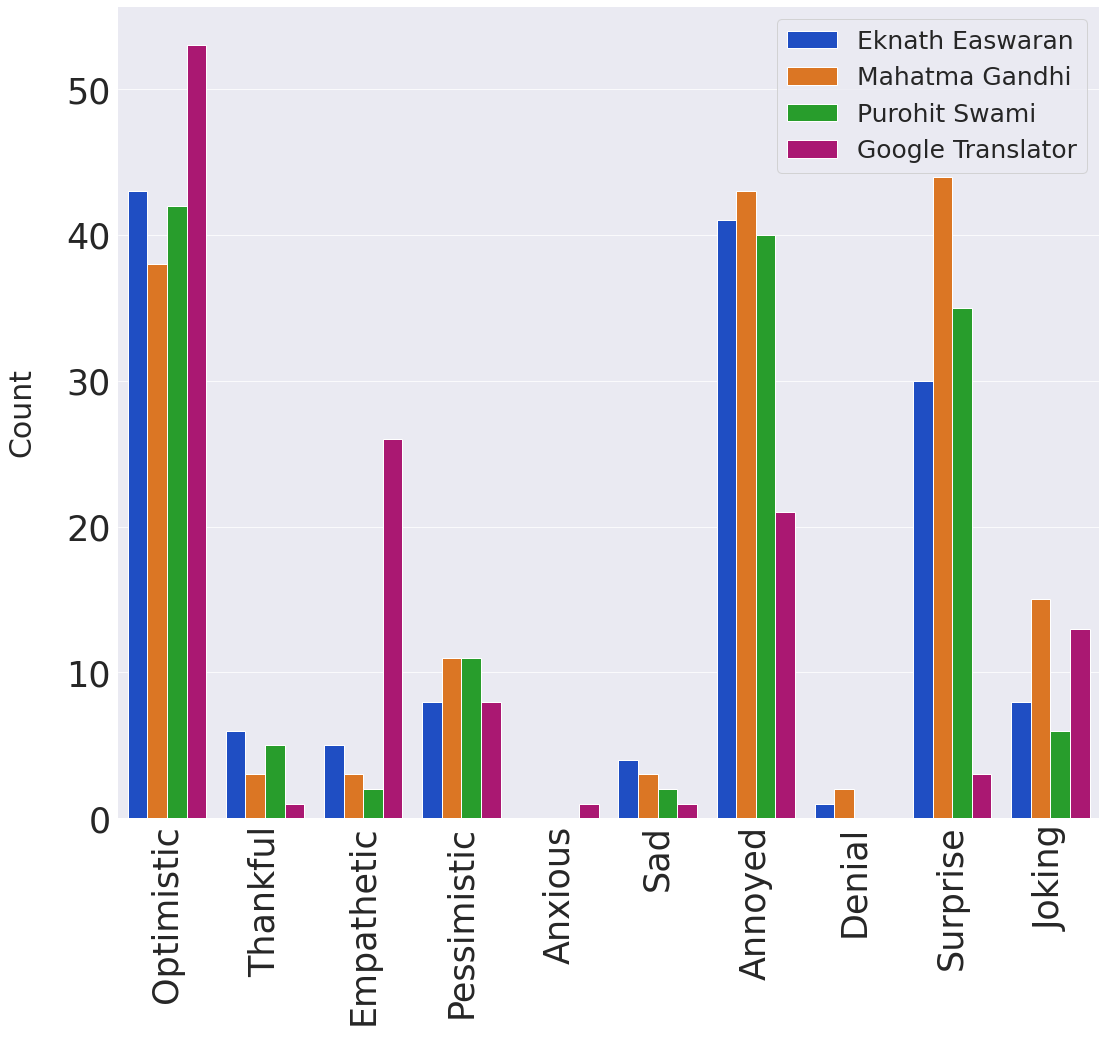}
    \caption{Chapter-18}
    \label{fig:chap18_sa}
     \end{subfigure}
        \caption{Chapter-wise Sentiment Analysis of Chapter 10 - Chapter 18.}
        \label{fig:chap_10_18}
\end{figure*}

We visualize chapter-wise sentiment analysis for all four translations as depicted by Figure \ref{fig:chap_1_9} and Figure \ref{fig:chap_10_18} along with cumulative sentiment analysis for all chapters as depicted by Figure \ref{fig:cumul_sa}. In cumulative sentiment analysis (Figure \ref{fig:cumul_sa}), we observe that \textit{thankful}, \textit{anxious}, \textit{sad}, and \textit{denial} are the least expressed sentiments across all four translations whereas \textit{optimistic} is the most expressed. We also observe that sentiments \textit{surprise} and \textit{annoyed} are under-expressed. In contrast, sentiment \textit{empathetic} is over-expressed by Google Translate when  compared to the other three translations. The sentiments \textit{optimistic}, \textit{pessimistic}, \textit{joking}, and \textit{anxious} are equally expressed in all four translations. We further note that \textit{optimistic}, and \textit{empathetic} are the leading sentiments for Google Translate  while \textit{annoyed}, \textit{pessimistic} and \textit{surprise} are leading sentiments for Mahatma Gandhi's version. Thus indicating that Google Translate leads in optimistic sentiments and Mahatma Gandhi's version leads in pessimistic sentiments.

\begin{figure*}[ht]
     \centering
     \begin{subfigure}[b]{.45\linewidth}
         \centering
         \includegraphics[width=\linewidth]{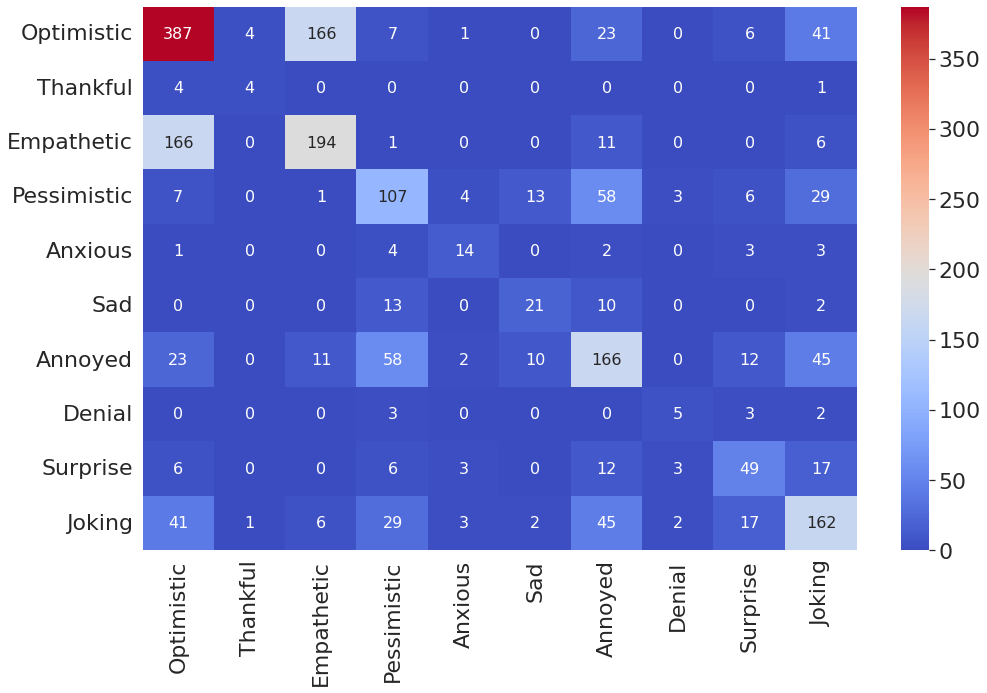}
         \caption{Google Translator}
        \label{fig:chap5_gt}
     \end{subfigure}
     \hfill
     \begin{subfigure}[b]{.45\linewidth}
         \centering
         \includegraphics[width=\linewidth]{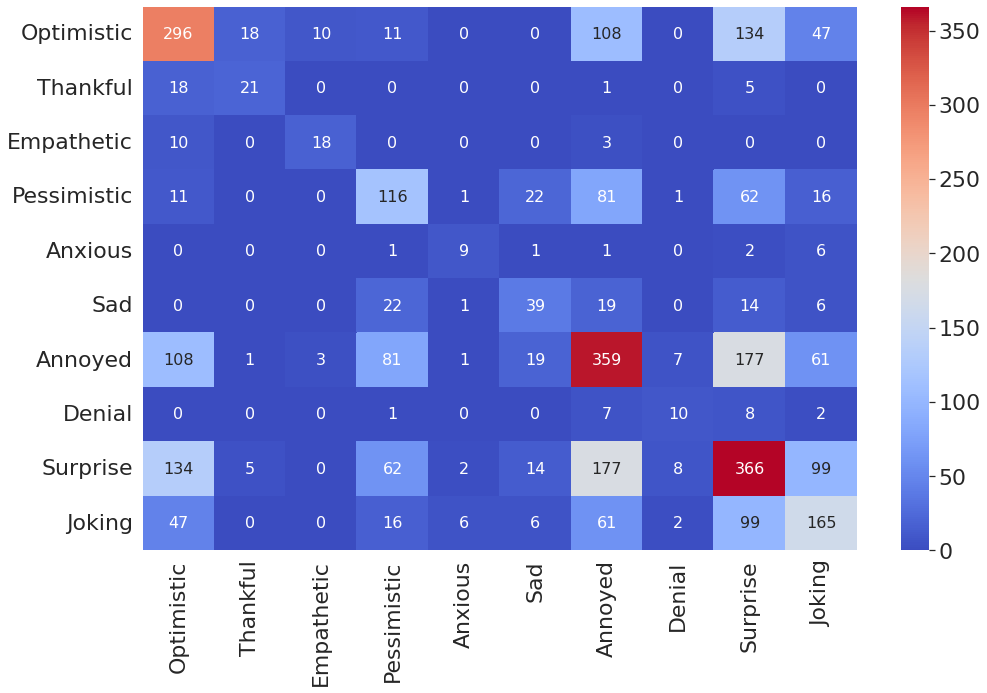}
        \caption{Mahatma Gandhi}
        \label{fig:chap5_mg}
     \end{subfigure}
     \hfill
     \begin{subfigure}[b]{.45\linewidth}
         \centering
         \includegraphics[width=\linewidth]{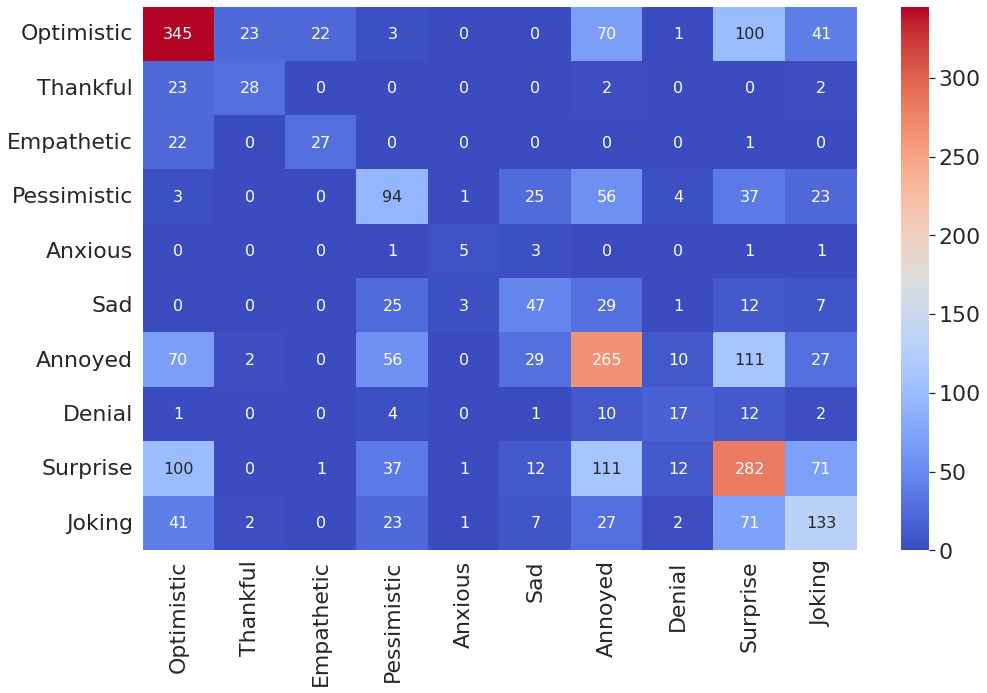}
          \caption{Eknath Easwaran}
          \label{fig:chap5_ee}
     \end{subfigure}
     \hfill
     \begin{subfigure}[b]{.45\linewidth}
         \centering
         \includegraphics[width=\linewidth]{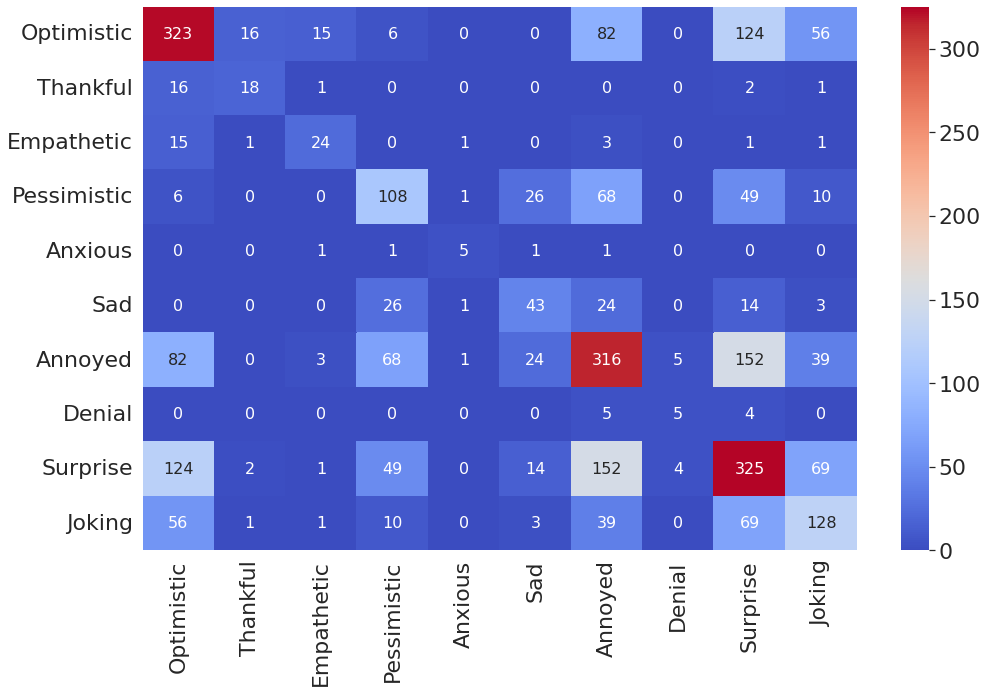}
        \caption{Purohit Swami}
        \label{fig:chap5_ps}
     \end{subfigure}
        \caption{Heat Map of different Bhagavad Gita translations.}
        \label{fig:heatmap}
\end{figure*}

Figure \ref{fig:heatmap} displays a heat map showing the frequency of a specific sentiment in each translation of all the verses compared to the other sentiments. We observe that in the case of  Google Translate in  Figure \ref{fig:chap5_gt},  \textit{empathetic} is the key sentiment in addition to the sentiments \textit{optimistic}, \textit{annoyed} and \textit{joking}, which are key sentiments for the rest of the three translations as shown by Figure \ref{fig:chap5_mg}, Figure \ref{fig:chap5_ee} and Figure \ref{fig:chap5_ps}. We further observe that the sentiment combination [\textit{optimistic}, \textit{empathetic}] are the leading combinations of sentiments of  Google Translate. In the other three versions, the leading combinations of sentiments are [\textit{annoyed}, \textit{surprise}] followed by [\textit{surprise}, \textit{optimistic}] and [\textit{annoyed},\textit{optimistic}]. It is also important to note that for Google Translate, the sentiments such as \textit{thankful} and \textit{denial} are the least expressed sentiments. In contrast, the sentiments such as  \textit{denial} and \textit{anxious} are the least expressed sentiments in the other three versions. 

Finally, we measure the diversity and similarity of sentiments expressed with verse-by-verse comparison for all four translations. Table \ref{table:jaccard} shows the Jaccard similarity score computed on the predicted sentiments for three pairs of texts for the selected chapters. The score is highest for Eknath Easwaran's version and Google Translate (GT-Easwaren), indicating they had the highest overlap of the predicted sentiments. The comparison of Gandhi-Easwaren shows the baseline from previous study \cite{Chandra2022-mh} where we find that GT-Easwaren has a much lower score, hence a much lower similarity. This indicates that Google Translate has not been as effective as human experts in translating the Bhagavad Gita. 

\begin{table}
\small
\centering
\begin{tabular}{|p{1.45cm}|p{1.4cm}|p{1.4cm}|p{1.4cm}|p{1.4cm}|} 
 \hline
 \textbf{Chapters} & \textbf{GT-Gandhi} & \textbf{GT-Purohit} & \textbf{GT-Easwaren} & \textbf{Gandhi-Easwaren}\\ [0.5ex]
 \hline
 Chapter 3 & 0.42 & 0.388 & 0.412 & 0.604\\ 
 \hline
 Chapter 5 & 0.374 & 0.373 & 0.401 & 0.568\\
 \hline
 Chapter 7 & 0.353 & 0.363 & 0.393 & 0.559\\
 \hline
 Chapter 8 & 0.341 & 0.362 & 0.377 & 0.547\\
 \hline
 Chapter 9 & 0.331 & 0.353 & 0.348 & 0.501\\
 \hline
 Chapter 10 & 0.324 & 0.351 & 0.357 & 0.523\\
 \hline
 Chapter 11 & 0.309 & 0.324 & 0.350 & 0.507\\
 \hline
 Chapter 12 & 0.315 & 0.323 & 0.357 & 0.500\\
 \hline
 Chapter 15 & 0.309 & 0.319 & 0.354 & 0.494\\
 \hline
 Chapter 16 & 0.316 & 0.328 & 0.359 & 0.500\\
 \hline
 Chapter 17 & 0.323 & 0.332 & 0.355 & 0.510\\
 \hline
 \textbf{Average} & \textbf{0.338} & \textbf{0.347} & \textbf{0.369} & \textbf{0.526}\\
 \hline
\end{tabular}
\caption{Sentiment analysis of selected pairs of translations by Google Translate (GT) with Jaccard
similarity score of the predicted sentiments for selected Chapters. We provide the mean of the scores at the bottom and lower score indicates lower similarity. The comparison of Gandhi-Easwaren shows the baseline from previous study \cite{Chandra2022-mh}.  }
\label{table:jaccard}
\end{table}

\subsection{Semantic Analysis}

Next, we provide the semantic analysis of the texts and compare the four translations. Using the MPNet-base model, we encode all the verses and present the verse-by-verse cosine similarity, grouped by chapter, for the three translations with Google Translate. We report both the mean and standard deviation of the score. In Table \ref{table:cosine-similarity}, we observe that Chapter 3 is semantically most similar, whereas Chapter 17 is semantically least similar. Further, in the pair-wise comparison, Google Translate and Shri Purohit Swami’s translations are most similar. These two translations also have the highest Jaccard similarity score for the predicted sentiments (Table \ref{table:jaccard}). We finally compare Gandhi-Easwaren to show a baseline   from previous study \cite{Chandra2022-mh} where we find that GT-Easwaren has a   much lower similarity that shows  that Google Translate has not been as effective when compared to human experts. 

\begin{table}
\small
\centering
\begin{tabular}{|p{1.5cm}|p{1.6cm}|p{1.6cm}|p{1.6cm}|p{1.60cm}|} 
 \hline
 \textbf{Chapters} & \textbf{GT-Gandhi} & \textbf{GT-Purohit} & \textbf{GT-Easwaren} & \textbf{Gandhi-Easwaren} \\ [0.5ex]
 \hline
 Chapter 3 & 0.52(0.156) & 0.58(0.148) & 0.59(0.120) & 0.63(0.133)\\ 
 \hline
 Chapter 5 & 0.34(0.082) & 0.61(0.133) & 0.51(0.187) & 0.63(0.129)\\
 \hline
 Chapter 7 & 0.35(0.194) & 0.56(0.232) & 0.35(0.100) & 0.70(0.144)\\
 \hline
 Chapter 8 & 0.36(0.086) & 0.34(0.104) & 0.38(0.098) & 0.66(0.123)\\
 \hline
 Chapter 9 & 0.33(0.108) & 0.36(0.113) & 0.35(0.103) & 0.68(0.126)\\
 \hline
 Chapter 10 & 0.33(0.121) & 0.37(0.118) & 0.38(0.093) & 0.76(0.096)\\
 \hline
 Chapter 11 & 0.36(0.118) & 0.38(0.108) & 0.38(0.105) & 0.71(0.109)\\
 \hline
 Chapter 12 & 0.35(0.122) & 0.40(0.159) & 0.35(0.118) & 0.61(0.120)\\
 \hline
 Chapter 15 & 0.40(0.135) & 0.39(0.129) & 0.37(0.142) & 0.69(0.116)\\
 \hline
 Chapter 16 & 0.38(0.126) & 0.37(0.128) & 0.41(0.089) & 0.66(0.096)\\
 \hline
 Chapter 17 & 0.30(0.077) & 0.35(0.128) & 0.33(0.115) & 0.65(0.111)\\
 \hline
 \textbf{Average} & \textbf{0.34(0.111)} & \textbf{0.43(0.142)} & \textbf{0.40(0.110)} & \textbf{0.67(0.119)} \\
 \hline
\end{tabular}
\caption{Semantic Analysis using cosine similarity score for comparing selected chapter pairs  of the translations. The mean score is given with standard deviation (in brackets) for all the verses in the respective chapters at the bottom (*). The lower score indicates  less similarity.  The comparison of Gandhi-Easwaren shows the benchmark from previous study \cite{Chandra2022-mh}.}
\label{table:cosine-similarity}
\end{table}

Next, we present some of the semantically most similar verses in Table \ref{table:most-similar-verse}. In Chapter 3 - Verse 13, we observe that all translations have conveyed a similar meaning; however, choice of words is different for all four. Our framework   assigns a high similarity score to all three pairs. In Chapter 11 - Verse 21 and Chapter 12 - Verse 19, we observe that Google Translate  and Eknath Easwaran have used somewhat similar words and thus have obtained a higher similarity score (Score 2). We present some of the semantically least similar verses in Table \ref{table:least-similar-verse}. We observe that for Chapter 12 - Verse 19, Google Translate  and Eknath Easwaran convey very different themes and thus have been given a very low similarity score.

\begin{table*}[htbp!]
\small
\centering
\begin{tabular}{|p{1cm}|p{0.75cm}|p{2.5cm}|p{2.5cm}|p{2.5cm}|p{2.5cm}|p{1cm}|p{1.25cm}|p{1cm}|} 
 \hline
 \textbf{Chapter} & \textbf{Verse}  & \textbf{GT} & \textbf{Gandhi} & \textbf{Easwaran} & \textbf{Swami} & \textbf{GT-Gandhi} & \textbf{GT-Easwaran} & \textbf{GT-Purohit} \\ [0.5ex]
 \hline
 3 & 13 & Those who eat the remains of the sacrifice are freed from all sins and enjoy the sins of the sinners who cook for their own sake & The righteous men who eat the residue of the sacrifice are freed from all sin, but the wicked who cook for themselves eat sin. & The spiritually minded, who eat in the spirit of service, are freed from all their sins; but the selfish, who prepare food for their own satisfaction, eat sin. & The sages who enjoy the food that remains after the sacrifice is made are freed from all sin: but the selfish who spread their feast only for themselves feed on sin only. & 0.919 & 0.705 & 0.836\\ 
 \hline
 7 & 9 & I am pious and fragrance on the earth and I am the effulgence of fire and I am the life of all living beings and I am the austerities of all living beings. & I am the sweet fragrance in earth; the brilliance in fire; the life in all beings; and the austerity in ascetics. & I am the sweet fragrance in the earth and the radiance of fire; I am the life in every creature and the striving of the spiritual aspirant. & I am the Fragrance of earth, the Brilliance of fire. I am the Life Force in all beings, and I am the Austerity of the ascetics. & 0.862 & 0.873 & 0.855\\ 
 \hline
 12 & 12 & Knowledge is the best way to practice knowledge and meditation is superior to meditation. From meditation, renunciation of the fruits of action is attained by renunciation. & Better is knowledge than practice, better than knowledge is concentration,better than concentration is renunciation of the fruit of all action, from which directly issues peace. &Better indeed is knowledge than mechanical practice. Better than knowledge is meditation. But better still is surrender of attachment to results, because there follows immediate peace. &Knowledge is superior to blind action, meditation to mere knowledge,renunciation of the fruit of action to meditation, and where there is renunciation peace will follow. & 0.681 & 0.739 & 0.813 \\ 
 \hline
 17 & 16 & The mind, grace, silence, self-control and self-control is called the mental state of self-realization. & Serenity, benignity, silence, self-restraint, and purity of the spirit—these constitute austerity of the mind. & Calmness, gentleness, silence, self-restraint, and purity: these are the disciplines of the mind. & Serenity, kindness, silence, self-control and purity – this is austerity of mind. & 0.569 & 0.669 & 0.554 \\ 
 \hline
\end{tabular}
\caption{Semantically most similar verses using the cosine similarity (score) using selected translations for comparison (Gandhi, Easwaren, Swami) vs Google Translate (GT).}
\label{table:most-similar-verse}
\end{table*}

\begin{table*}[htbp!]
\small
\centering
\begin{tabular}{|p{1cm}|p{0.75cm}|p{2.5cm}|p{2.5cm}|p{2.5cm}|p{2.5cm}|p{1cm}|p{1.25cm}|p{1cm}|} 
 \hline
 \textbf{Chapter} & \textbf{Verse}  & \textbf{GT} & \textbf{Gandhi} & \textbf{Easwaran} & \textbf{Swami} & \textbf{GT-Gandhi} & \textbf{GT-Easwaran} & \textbf{GT-Swami} \\ [0.5ex]
 \hline
 11 & 41 & O Krishna, I thought that I was a friend, O Krishna, O friend of the demigods. & If ever in carelessness, thinking of You as comrade, I addressed You saying, ‘O Krishna!', ‘O Yadava!' not knowing Your greatness, in negligence or in affection, & Sometimes, because we were friends, I rashly said, Oh, Krishna! Say, friend! casual, careless remarks. Whatever I may have said lightly, whether we were playing or resting, alone or in company, sitting together or eating, & Whatever I have said unto You in rashness, taking You only for a friend and addressing You as `O Krishna! O Yadava! O Friend!’ in thoughtless familiarity, no understanding Your greatness; & 0.36 & 0.38 & 0.48 \\ 
 \hline
 17 & 26 & This is used in the same way as the truth, O son of Pṛthā, and in the praiseworthy action, which is used in the same way as the words of the Lord. & SAT is employed in the sense of ‘real' and ‘good'; O Arjuna, SAT is also applied to beautiful deeds. & Sat means that which is; it also indicates goodness. Therefore it is used to describe a worthy deed.& `Sat’ means Reality or the highest Good, and also, O Arjuna, it is used to mean an action of exceptional merit.& 0.37 & 0.39 & 0.36 \\ 
 \hline
\end{tabular}
\caption{Semantically least similar verses using the cosine similarity (score) using selected comparisons (Gandhi, Easwaren, Swamni) vs Google Translate (GT).}
\label{table:least-similar-verse}
\end{table*}

In addition, we examine the semantic score by showing actual verses from translated versions of a chosen chapter. We select Chapter 12 because it includes the least verses, making it easier to include it in the paper. Table \ref{table:ch12} presents arbitrarily selected verses from Chapter 12 with the cosine similarity score. We also present the mean and standard deviation of the scores to give a sense of the general semantic similarity of the verses in the chapter for the comparison of chosen translations.

\begin{table*}
\small
\centering
\begin{tabular}{|p{1cm}|p{0.75cm}|p{2.5cm}|p{2.5cm}|p{2.5cm}|p{2.5cm}|p{1cm}|p{1.25cm}|p{1cm}|} 
 \hline
 \textbf{Chapter} & \textbf{Verse}  & \textbf{GT} & \textbf{Gandhi} & \textbf{Easwaran} & \textbf{Swami} & \textbf{GT-Gandhi} & \textbf{GT-Easwaran} & \textbf{GT-Swami} \\ [0.5ex]
\hline
12 & 1 & Arjuna said: Those devotees who are thus constantly engaged in worshiping You, who are also the most unmanifest of the unmanifest, who are the best in yoga? & Of the devotees who thus worship You, incessantly attached, and those who worship the Imperishable Unmanifest, which are the better yogins? The Lord Said: & ARJUNA Of those steadfast devotees who love you and those who seek you as the eternal formless Reality, who are the more established in yoga? & “Arjuna asked: My Lord! Which are the better devotees who worship You, those who try to know You as a Personal God, or those who worship You as Impersonal and Indestructible? & 0.52 & 0.68 & 0.70 \\ 
\hline
12 & 8 & Concentrate on Me in Me in Me, fix your mind on Me. You will live in Me. In Me alone, there is no doubt that there is no doubt about it. &  On Me set your mind, on Me rest your conviction; thus without doubt shall you remain only in Me hereafter. & Still your mind in me, still your intellect in me, and without doubt you will be united with me forever. & Then let your mind cling only to Me, let your intellect abide in Me; and without doubt you shall live hereafter in Me alone. & 0.61 & 0.67 & 0.60 \\ 
\hline
12 & 13 & He is not hated by all living beings, friendly and compassionate. & Who has ill-will towards none, who is friendly and compassionate, who has shed all thought of ‘mine' or ‘I', who regards pain and pleasure alike, who is long-suffering; & That one I love who is incapable of ill will, who is friendly and compassionate. Living beyond the reach of I and mine and of pleasure and pain, & He who is incapable of hatred towards any being, who is kind and compassionate, free from selfishness, without pride, equable in pleasure and in pain, and forgiving, & 0.34 & 0.21 & 0.39\\ 
\hline
12 & 15 & He who is freed from all joy, anger, fear and anxiety, who is not afraid of the world and who is not afraid of this world. & Who gives no trouble to the world, to whom the world causes no trouble, who is free from exultation, resentment, fear and vexation,—that man is dear to Me. & Not agitating the world or by it agitated, they stand above the sway of elation, competition, and fear: that one is my beloved. & He who does not harm the world, and whom the world cannot harm, who is not carried away by any impulse of joy, anger or fear, such a one is My beloved. & 0.70 & 0.50 & 0.70\\ 
\hline
12 & 20 & Those who worship this nectar of religious principles as described above are very dear to Me and are very dear to Me. & They who follow this essence of dharma, as I have told it, with faith, keeping Me as their goal,—those devotees are exceeding dear to Me. & Those who meditate upon this immortal dharma as I have declared it, full of faith and seeking me as lifes supreme goal, are truly my devotees, and my love for them is very great. & Verily those who love the spiritual wisdom as I have taught, whose faith never fails, and who concentrate their whole nature on Me, they indeed are My most beloved.” & 0.69 & 0.61 & 0.66\\ 
 
\hline  
\hline
\end{tabular}
\caption{Semantic similarity of verses selected from Chapter 12  with cosine similarity (score) using selected translations (Gandhi, Easwaren, Swamni) to compare with  Google Translate (GT). We also provide the score mean and standard deviation (in brackets) of the scores at the bottom (*).}
\label{table:ch12}
\end{table*}

\section{Evaluation by Sanskrit Expert }

We further evaluate selected verses from Google Translate in comparison with expert translations, with help of a Sanskrit researcher, Sushrut Badhe \footnote{\url{https://en.wikipedia.org/wiki/Sushrut_Badhe}} who has published a translation of the Bhagavad Gita in 2015 \cite{badhe2015}. The unique part of this translation was  that the rhythm and rhyme was maintained in the English translation   following the original Sanskrit version. We note that the rhythm and rhyme are the key attributes of the Bhagavad Gita in Sanskrit since it was written to be sung and remembered through oral traditions for thousands of years. In consultation with Sushrut Badhe, we provide the following analyses about selected chapters and verses included in the paper. 

%हिन्दी शब्द का सम्बन्ध संस्कृत शब्द 'सिन्धु' से माना जाता है।
\subsection{Semantically most similar verses}

\begin{figure*}[htbp]
\centering
    \includegraphics[scale=0.23]{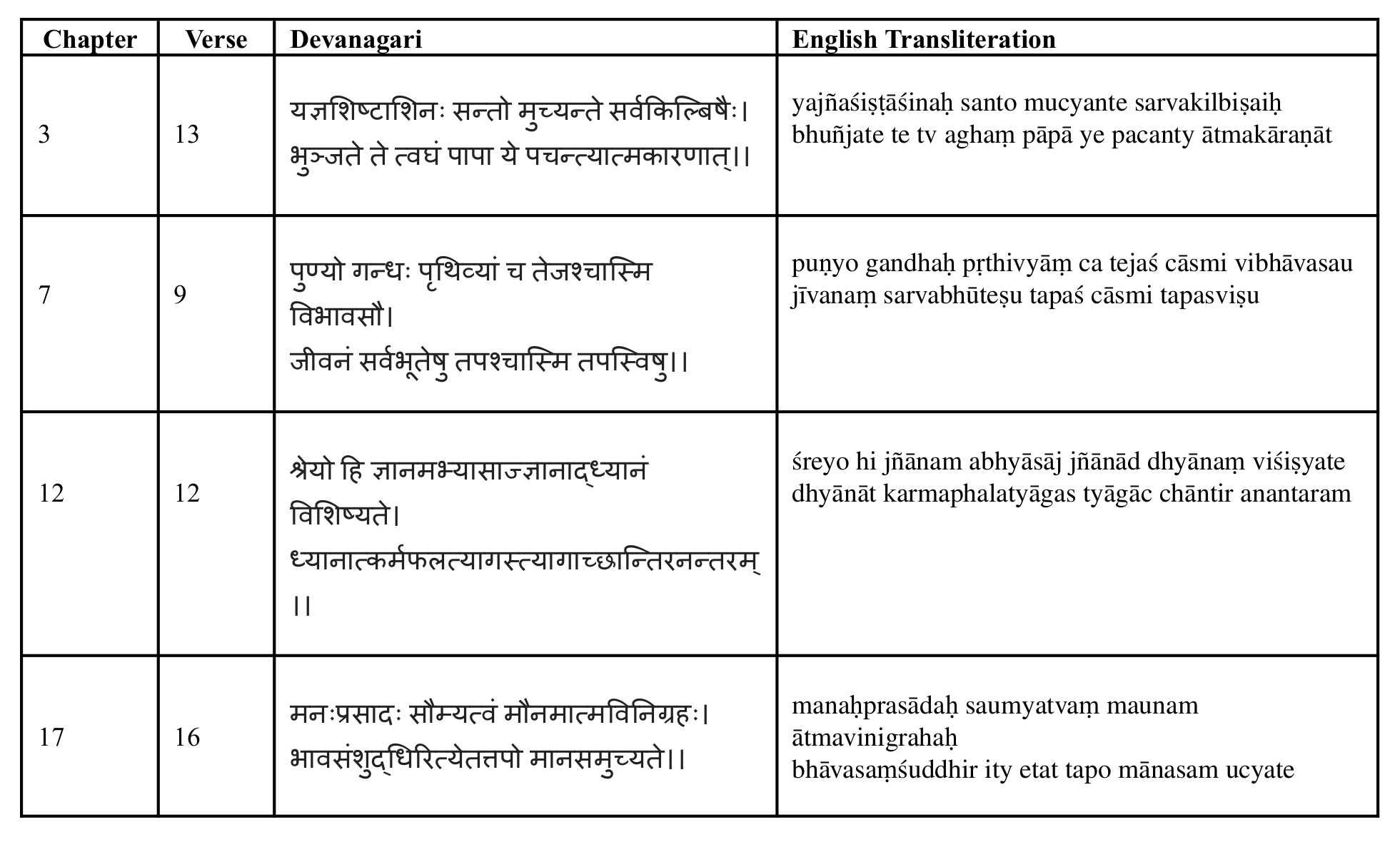}
    \caption{ An extension of semantically most similar verses across translations in Table   \ref{table:most-similar-verse} showing original Sanskrit verses from the Bhagavad Gita \cite{prabhupada1972bhagavad} in Devanagri and English Transliteration.  }
    \label{fig:fig3}
\end{figure*}

Table \ref{table:most-similar-verse}, we show selected semantically most similar verses using three expert translations (Gandhi, Easwaren, and Swami) and Google Translate (GT), with accomianying original Sanskrit verses in Figure \ref{fig:fig3}. In Chapter 3: Verse 13, we find that both GT-Gandhi  (i.e GT vs Gandhi) and GT-Swami are more semantically similar than GT-Easwaran. However, the GT version merges both the lines of the verse and gives a confusing answer and loses contextual significance entirely. The original Sanskrit Verse (Chapter 3: Verse 13) of the Bhagavad Gita implies that those who consume the food that is a remainder after performing sacrifice are freed of all their sins whereas those who cook and consume only for themselves end up consuming only sin. Google translate version conveys a wrong meaning. Also the word  \textit{santo}, which is significant and refers to the saints and spiritual minded people; has been omitted arbitrarily in the translation. The translations of Easwaran, Gandhi, and Swami, though semantically dissimilar do not lose contextual significance.

In Chapter 8: Verse 21 of Table \ref{table:most-similar-verse}, the values of cosine similarity for all three combinations are nearly equal with GT-Easwaran  showing the maximum semantic similarity.  In this case, all  the four translations are contextually significant. Google translate version has accurately translated the word \textit{punyo} (Figure \ref{fig:fig3}) as ‘pious’. Easwaren and Gandhi have translated it as ‘sweet’ whereas Swami has omitted its translation. In this verse, Google translate appears to be the more accurate version. In Chapter 11: Verse 21, both GT-Swami and GT-Easwaran-  are more semantically similar than GT-Gandhi.The translations of Eashwaren, Gandhi and Swami  are contextually significant. The GT version is  incorrect and bereft of logical sense or contextual significance.

In Chapter 12: Verse 19  of Table \ref{table:most-similar-verse}, GT-Swami and GT-Easwaran are more semantically similar than GT-Gandhi. The translations of Gandhi, Easwaren, and Swami are contextually significant.  
GT version, \textit{“Knowledge is best way to practice knowledge and meditation is superior to meditation. From meditation, renunciation of fruits of actions is attained by renunciation.”}, is bereft of logic and contextual significance.

In Chapter 17: Verse 16  of Table \ref{table:most-similar-verse}, GT-Easwaran  is most semantically similar. The translations of Easwaren, Gandhi and  Swami  are contextually significant. GT version has only literal word to word translation which does not convey a clear meaning and lacks contextual significance.  The word \textit{manaḥprasādaḥ} (Figure \ref{fig:fig3}) is wrongly translated as ‘the mind, grace’ and this affects the logical meaning of the translation.

\subsection{Semantically less similar verses}

\begin{figure*}[htbp]
\centering
    \includegraphics[scale=0.23]{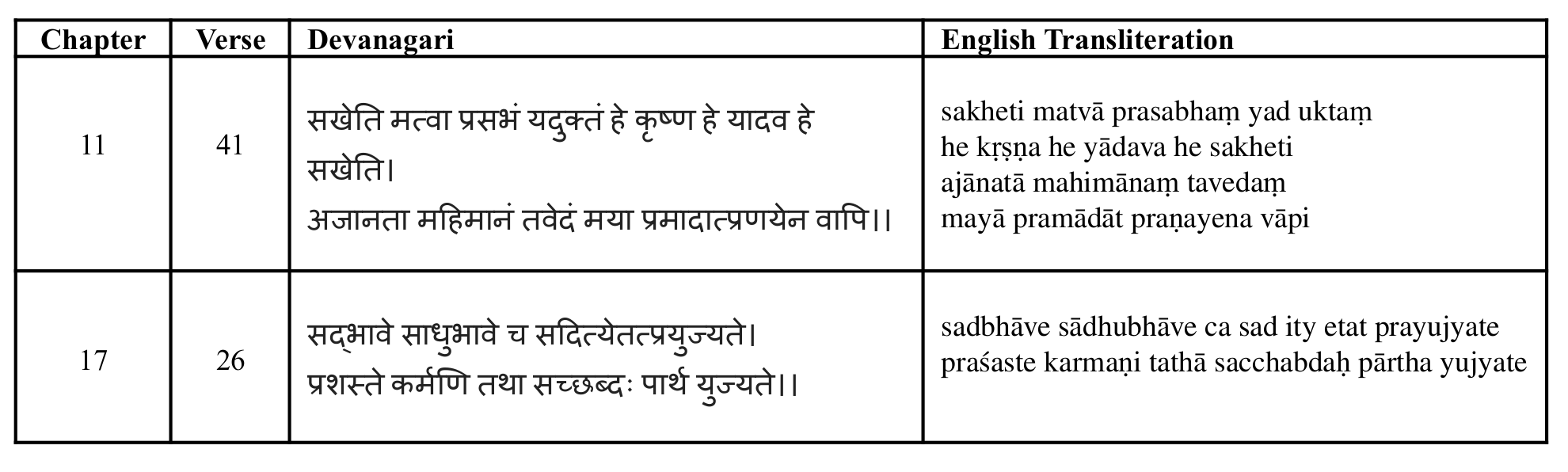}
    \caption{ Semantically least similar verses across translations in Table \ref{table:least-similar-verse}  showing original Sanskrit verses from the Bhagavad Gita  \cite{prabhupada1972bhagavad}  in Devanagri and English Transliteration.}
    \label{fig:fig4}
\end{figure*}

Table \ref{table:least-similar-verse} presents selected less similar verses, having low cosine scores of semantic similarity with original shown in Figure \ref{fig:fig4}. In both verses, we find that GT only gave a literal translation that was of no contextual significance or meaning.

In Chapter 11: Verse 41 of Table \ref{table:least-similar-verse}, GT-Swami is most similar in terms of its cosine value of semantic similarity. The GT version, \textit{“O Krishna, I thought that I was a friend, O Krishna, O friend of the demigods”}, does not convey a logical sense and is incorrect.

In Chapter 17: Verse 26, GT-Easwaran is most similar in terms of its cosine value of semantic similarity.  
The google translator version, \textit{“This is used in the same way as the truth, O son of Pritha, and in the praiseworthy action, which is used in the same way as the words of the Lord.”}, lacks both logic and contextual significance.

\subsection{Chapter 12: Arbitrarily selected  verses}

% where we find that in all the selected  verses, Eashwaren, Gandhi  and Swami retained contextual significance whereas GT version doesn’t retain any contextual significance and not sound. For instance,  the GT version of  Verse 12.8, “Concentrate on Me in Me in ME, fix your mind in Me. You will live in Me. In Me alone, there is doubt that there is no doubt about it.” is completely bereft of logic and significance.

 Chapter 12 of the Gita is considered to be one of the important chapters as it contains the verses that are relevant to the crux of the teaching of the Gita – the way of Bhakti (devotion). We select five arbitrarily verses from Chapter 12 (Table \ref{table:ch12}). In general, we find that the translations of Easwaren, Gandhi and Swami  did not lose contextual significance. However, on the contrary, we find that GT conveyed no contextual meaning in all five verses.

Chapter 12: Verse 1, GT-Swami is most similar in terms of its cosine value of semantic similarity. The translations of Easwaren, Gandhi and Swami are contextually significant. The GT version, \textit{“Those devotees who are constantly engaged in worshipping You, who are also the most unmanifest of the unmanifest, who are the best in yoga?”}  is bereft of logical or contextual significance.

Chapter 12: Verse 8, GT-Easwaran is most similar in terms of its cosine value of semantic similarity. %The translations of EE, MG & SSP are contextually significant. 
The GT version, \textit{“Concentrate on Me in Me in Me, fix your mind in Me. You will live in Me. In Me alone, there is doubt that there is no doubt about it”} does not make any sense  and also incorrect.

Chapter 12: Verse 13, GT-Swami is most similar in terms of its cosine value of semantic similarity. The translations of Easwaren, Gandhi and Swami are contextually significant  are contextually significant.
This verse originally indicates the temperament of a devotee who harbours no hate or ill will for any human being.  The GT version, \textit{“He is not hated by all living beings, friendly and compassionate”} sounds logical but does not hold contextual significance.

Chapter 12: Verse 15, GT-Swami is most similar in terms of its cosine value of semantic similarity.  
The GT version, \textit{“He who is freed from all joy, anger, fear and anxiety, who is not afraid of the world and who is not afraid of this world”} is improper as it misses the meaning of the original verse which implies that the altruistic soul who is free from all the bonds of pleasure, fear, anger and anxiety neither disturbs the world or is not disturbed by it.

Chapter 12: Verse 20, GT-Gandhi is most similar in terms of its cosine value of semantic similarity. 
The GT version, \textit{“Those who worship this nectar of religious principles as described above are very dear to Me and are very dear to Me”} though sounding logical, loses contextual significance in the last part, and features repetition.

\begin{figure*}[htbp]
\centering
    \includegraphics[scale=0.23]{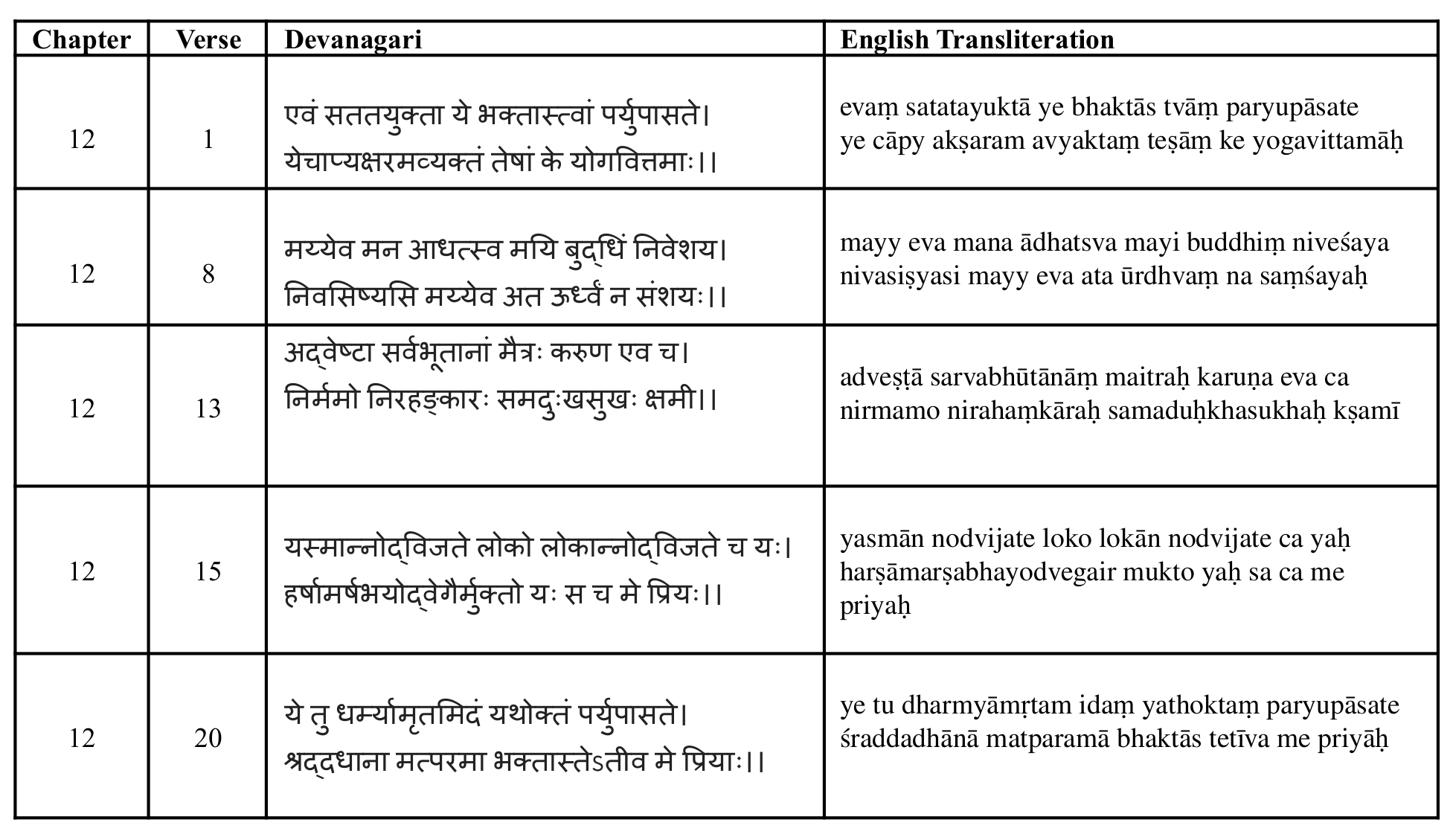}
    \caption{   Semantic similarity of verses selected from Chapter 12  across translations given in Table \ref{table:ch12}  showing original Sanskrit verses from the Bhagavad Gita  \cite{prabhupada1972bhagavad}  in Devanagri and English Transliteration.}
    \label{fig:fig5}
\end{figure*}

\section{Discussion}

%The verses of the Bhagavad Gita are poetic and google translator is unable to understand the contextual significances of the original Sanskrit words. Google translate is unsuitable for poetic Sanskrit works

Among the verses selected for qualitative assessment with assistance of a Sanskrit researcher, we found that only one verse (Table \ref{table:most-similar-verse}, Chapter 7 - Verse 9) was translated correctly, capturing the context and the foundations of Hindu philosophy.  In the rest of the  verses which had contextual references or poetic elements,  these were mistranslated. If we were to look closely at the singular verse translated accurately by google translate, we can see that the original Sanskrit Chapter 7: Verse 9 (Table \ref{table:most-similar-verse},  contains 9 distinct words that are bereft of any wordplay or poetic inferences. However, when we see the Sanskrit Chapter 12: Verse 8  (Table \ref{table:ch12}) due to the presence of words having the same roots, Google Translate is unable to identify the significance, where \textit{“mayy eva mana ādhatsva mayi buddhiṃ niveśaya”} is wrongly translated as \textit{“Concentrate on Me in Me in Me, fix your mind in me”}.

  The discrepancies in translation can be thus, attributed to the inability of Google Translate to understand context of the root words. The same word of Sanskrit language can have multiple meanings which have to be understood depending on the context of the statement. Most of the ancient Sanskrit epics such as Ramayana and Mahabharata, are written in the form of \textit{shloka} (stanza) and they are embedded with references and allegories. Also, the verses from various chapters are inextricably linked. This proves to be a major challenge in translation and can lead to erroneous results if the verses are translated independently without understanding the references. For instance, in the Bhagavad Gita, in the verses (Chapter 9: Verse 34 and Chapter 18: Verse 65), the original Sanskrit words are exactly same in the first three parts of the shloka, only the fourth part is different as shown below in bold:
\begin{itemize}
    \item Sanskrit Transliteration  - Chapter 9: Verse 34: "man-manā bhava mad-bhakto mad-yājī māṁ namaskuru
mām evaiṣhyasi yuktvaivam ātmānaṁ mat-parāyaṇaḥ"
\item  Google Translate -  Chapter 9: Verse 34: "Be mindful of Me, be devoted to Me, live in Me, and bow down to Me
You will come to Me alone, thus uniting yourself and being devoted to Me"

\end{itemize}

\begin{itemize}
\item Sanskrit Transliteration  - Chapter 18: Verse 65:  "man-manā bhava mad-bhakto mad-yājī māṁ namaskuru
mām evaiṣhyasi satyaṁ te pratijāne priyo ‘si me"
\item  Google Translate -  Chapter 18: Verse 65:   "Be mindful of Me, be devoted to Me, live in Me, and bow down to Me
You will come to me I promise you truly you are dear to me"

\end{itemize}

If we were to analyse the Google Translate versions of both verses, they sound fairly similar and do not convey much about the contextual significance of the verses. In Chapter 9, Arjuna continues to be in a state of confusion as he listens intently to Krishna whereas in the Chapter 18, Arjuna’s doubts are completely resolved and these words hold a complete difference in both spiritual and psychological terms. The text of the Gita has been understood to hold a significant psychotherapeutic potential and it has been recommended that its pragmatic use can improve both trust and communication \cite{bhatia2013bhagavad}. %(Indian J Psychiatry. 2013 Jan; 55(Suppl 2): S315–S321. doi: 10.4103/0019-5545.105557)

In Chapter 17: Verse 26 (Table \ref{table:least-similar-verse}) Lord Krishna explains to Arjuna the meaning of \textit{sat} (literally translated as truth) which is part of a triple formula – \textit{‘Om Tat Sat’} introduced in a previous verse of the same chapter (Chapter 17 Verse 25). This reference is lost in the google translation altogether. This is a significant concept from the Bhagavad Gita which had a number of interpretation by prominent scholars since ancient times and been prominent in Vedanta Hindu school of philosophy \cite{sharma1960history,anderson2012investigation}.

In Hindu philosophy, Om is the most sacred term - it has its own alphabet symbol in the Sanskrit and Hindi script known as Devanagari \cite{bright1996devanagari} as shown e.g. in (Figures \ref{fig:fig3}). Om is not really part of the Devanagari script, i.e it is left alone and not used to form other words. Hence, the Devanagari script views Om as scared since it is beyond philosophy and descriptions in Hinduism.   Om is a symbolic representation of the impersonal aspect of God, the Supreme one, an idea so pure and . Om represents all that was there before the birth of the universe; more precisely, before the birth of  the multiverse, since Hinduism introduced the idea of multiverse through its philosophy and mythology \cite{capra2010tao}.  Om refers to the formless Brahman   and is the primordial sound that pervades creation \cite{phillips1986aurobindinos}. Note that Brahman is defined as ultimate reality in the universe (multiverse) \cite{capra2010tao} and is also one of the terms that cant be translated to English easily as it changes meaning in different contexts \cite{chaudhuri1954concept}, similar to Dharma and Karma. Brahman is the pervasive, eternal truth, and consciousness  which does  not change; however, it is the cause of all changes. Hence, Brahman can be seen as a philosophical paradox \cite{krishnaancient}. It can be argued that Brahman is the closest word to the concept of God in Abrahamic religions; however, it is also different since God is known to be creator, protector and observe; whereas Brahman has all these properties, but also remains part of the universe. In the Isha Upanishad \cite{easwaran2007upanishads,greeff1998mysticism}, this verse further defines the property of Brahman:

Om \\
Purnamadah Purnamidam \\
Purnat Purnamudachyate \\
Purnasya Purnamadaya \\
Purnameva Vashishyate \\
Om shanti, shanti, shanti \\

\noindent which has been  translated by Eknath Easwaren \cite{easwaran2007upanishads} as:

Om \\
All this is full. All that is full \\
From fullness, fullness comes \\
When fullness is taken from fullness, \\
Fullness still remains. \\
Om Shanti, Shanti, Shanti \\

Note that full has been translated as infinite,  wholeness, complete, absolute, perfect, and reality by different translators of the Upanishads \cite{easwaran2007upanishads,whitney1890bohtlingk,mehta1970call}. Hence, the translation of the Upanishads poses similar challenges as the Bhagavad Gita. Om is the term that cannot be translated and remains as it is in most translations of Hindu texts. 

%Tat means that everything belongs to God, and by keeping that in mind, the people seeking liberation selflessly do the yajna, tapas  dana without desire of any fruits of the actions.[3]

%Sat  represents all the actions duties which are performed with truthfulness and related to the eternal truth. It is also said the everything present in yajna, tapas  dana are considered as "Sat", as well as actions meant solely for the satisfaction of the Supreme.[4]

%Limitations
%We note that prominently, Google translate has done a relatively good job in translations when compared to existing expert translations (Gandhi, Swami, Easwaren).

A major limitation is that the text that we have given is a philosophical song summarising major schools of Hindu philosophy, which had a number of interpretations, and hence distinct schools were formed. For instance, \textit{Advaita Vedanta} (non-dualism) \cite{anderson2012investigation,nelson1998dualism}  and \textit{Dvaita Vedanta} (dualism) \cite{sharma1960history,widgery1942dvaita} Vedanta schools developed out of philosophical differences in interpretations of the Bhagavad Gita. We note that Advaita  Vedanta became prominent from  Adi Shankara's interpretation of Bhagavad Gita  \cite{varma2018adi,namboodiripad1989adi} in the 8th century, known as the Sankara Bhashya \cite{gambhirananda1984bhagavad}. These schools formed when Sanskrit was a prominent language in studying Hindu philosophy, and the schools were formed not due to mistranslation but due to interpretation. Due to different schools of philosophy, there can be translation bias; i.e. a translator with Advaita Vedanta will translate with biases towards this school of philosophy and Dvaita Vedanta will also do the same. In terms of Google Translate, we note that such bias is not there, but then there are limitations that also create a bias.  Advaita Vedanta has been the most prominent school of Hindu philosophy in last thousand years with various texts of interpretations of the Bhagavad Gita and Vedas through scholars; hence, if these are used in the model training data, then model will philosophical biases.
%Future Research 

\section{Conclusion}

We presented a framework for evaluation of Google Translate using Sanskrit  as an example language. In our framework, we used a combination of semantic and sentiment analysis for comparing  expert translations of the Bhagavad Gita with Google Translate. %The Bhagavad Gita  is a sacred and philosophical Hindu text written in verse form.

In terms of sentiment analysis, a major observation was that the sentiments \textit{optimistic}, \textit{pessimistic}, \textit{joking}, and \textit{anxious} were equally expressed in all four translations. We found that Google Translate  lead in terms of optimistic sentiments and Mahatma Gandhi  lead in pessimistic sentiments. In semantic analysis, we found that Chapter 3 is semantically most similar; whereas Chapter 17 is semantically least similar when comparing the translations with Google Translate.  Generally, we found that Google Translate provided low level of semantic and sentiment similarity when compared to translations by human experts. This indicates that a lot has to be done to improve Google Translate in this domain since we are dealing with philosophical and metaphorical concepts in the Bhagavad Gita and a low resource language (Sanskrit) having a small number of native speakers. Furthermore, although Sanskrit is a low resource language, we note that it is an official language in India. Sanskrit is the main language for various ancient Hindu texts, and hence there has been a lot of focus on Sanskrit in academia. Therefore, the current study has a wide range of implications. Automatic translation of ancient texts could further help ease the burden of translating a text from scratch. 

We further compared selected translations using a qualitative approach  with help of a Sanskrit translator. In the qualitative evaluation, we find that Google translator is unsuitable for translation of poetic Sanskrit words and phrases due to its inability to recognize contextual significance and imagery.  The mistranslations  are not surprising as the Bhagavad Gita is known as  a difficult text to translate  and interpret since it relies on contextual, philosophical  and historical information.  

%  In the qualitative evaluation Google Translate with assistance of a Sanskrit translator, we find that Google translate is unsuitable in capturing certain Sanskrit words and phrases due to metaphorical and poetic style that lays the  foundation of the respective schools of  Hindu philosophy. 

There is a good scope for using our proposed framework for evaluation of Google Translate for other languages. As noted earlier, our current study used Sanskrit which is not much used as a conversational language and we evaluated Google Translate using the Bhagavad Gita which is a poem. Hence, in future work we can evaluate other languages from India, particularly Hindi which has third highest speakers in works as first and second language, after English and Mandarin. Apart from Hindi, our framework is essentially useful for any language which has already been translated by experts, which can be used for comparison with Google Translate version.

\section*{Code and Data }
Github repository:
\footnote{\url{https://github.com/sydney-machine-learning/Google-Sanskrit-translate-evaluation}}

%\bibliographystyle{model1-num-names}
%\bibliography{references}

\begin{thebibliography}{102}
\expandafter\ifx\csname natexlab\endcsname\relax\def\natexlab#1{#1}\fi
\providecommand{\url}[1]{\texttt{#1}}
\providecommand{\href}[2]{#2}
\providecommand{\path}[1]{#1}
\providecommand{\DOIprefix}{doi:}
\providecommand{\ArXivprefix}{arXiv:}
\providecommand{\URLprefix}{URL: }
\providecommand{\Pubmedprefix}{pmid:}
\providecommand{\doi}[1]{\href{http://dx.doi.org/#1}{\path{#1}}}
\providecommand{\Pubmed}[1]{\href{pmid:#1}{\path{#1}}}
\providecommand{\bibinfo}[2]{#2}
\ifx\xfnm\relax \def\xfnm[#1]{\unskip,\space#1}\fi
%Type = Article
\bibitem[{Najafabadi et~al.(2015)Najafabadi, Villanustre, Khoshgoftaar, Seliya,
  Wald, and Muharemagic}]{najafabadi2015deep}
\bibinfo{author}{M.~M. Najafabadi}, \bibinfo{author}{F.~Villanustre},
  \bibinfo{author}{T.~M. Khoshgoftaar}, \bibinfo{author}{N.~Seliya},
  \bibinfo{author}{R.~Wald}, \bibinfo{author}{E.~Muharemagic},
\newblock \bibinfo{title}{Deep learning applications and challenges in big data
  analytics},
\newblock \bibinfo{journal}{Journal of big data} \bibinfo{volume}{2}
  (\bibinfo{year}{2015}) \bibinfo{pages}{1--21}.
%Type = Book
\bibitem[{Manning and Schutze(1999)}]{manning1999foundations}
\bibinfo{author}{C.~Manning}, \bibinfo{author}{H.~Schutze},
  \bibinfo{title}{Foundations of statistical natural language processing},
  \bibinfo{publisher}{MIT press}, \bibinfo{year}{1999}.
%Type = Article
\bibitem[{Chandra and Kulkarni(2022)}]{Chandra2022-mh}
\bibinfo{author}{R.~Chandra}, \bibinfo{author}{V.~Kulkarni},
\newblock \bibinfo{title}{Semantic and sentiment analysis of selected {Bhagavad
  Gita translations using BERT-based} language framework},
\newblock \bibinfo{journal}{IEEE Access} \bibinfo{volume}{10}
  (\bibinfo{year}{2022}) \bibinfo{pages}{21291--21315}.
%Type = Article
\bibitem[{Dang et~al.(2020)Dang, Moreno-Garc{\'\i}a, and De~la
  Prieta}]{dang2020sentiment}
\bibinfo{author}{N.~C. Dang}, \bibinfo{author}{M.~N. Moreno-Garc{\'\i}a},
  \bibinfo{author}{F.~De~la Prieta},
\newblock \bibinfo{title}{Sentiment analysis based on deep learning: A
  comparative study},
\newblock \bibinfo{journal}{Electronics} \bibinfo{volume}{9}
  (\bibinfo{year}{2020}) \bibinfo{pages}{483}.
%Type = Article
\bibitem[{Kirill et~al.(2020)Kirill, Mihail, Sanzhar, Rustam, Olga, and
  Ravil}]{kirill2020propaganda}
\bibinfo{author}{Y.~Kirill}, \bibinfo{author}{I.~G. Mihail},
  \bibinfo{author}{M.~Sanzhar}, \bibinfo{author}{M.~Rustam},
  \bibinfo{author}{F.~Olga}, \bibinfo{author}{M.~Ravil},
\newblock \bibinfo{title}{Propaganda identification using topic modelling},
\newblock \bibinfo{journal}{Procedia Computer Science} \bibinfo{volume}{178}
  (\bibinfo{year}{2020}) \bibinfo{pages}{205--212}.
%Type = Incollection
\bibitem[{Egger(2022)}]{egger2022topic}
\bibinfo{author}{R.~Egger},
\newblock \bibinfo{title}{Topic modelling},
\newblock in: \bibinfo{booktitle}{Applied Data Science in Tourism},
  \bibinfo{publisher}{Springer}, \bibinfo{year}{2022}, pp.
  \bibinfo{pages}{375--403}.
%Type = Inproceedings
\bibitem[{Bertoldi et~al.(2007)Bertoldi, Zens, and
  Federico}]{bertoldi2007speech}
\bibinfo{author}{N.~Bertoldi}, \bibinfo{author}{R.~Zens},
  \bibinfo{author}{M.~Federico},
\newblock \bibinfo{title}{Speech translation by confusion network decoding},
\newblock in: \bibinfo{booktitle}{2007 IEEE International Conference on
  Acoustics, Speech and Signal Processing-ICASSP'07},
  volume~\bibinfo{volume}{4}, \bibinfo{organization}{IEEE},
  \bibinfo{year}{2007}, pp. \bibinfo{pages}{IV--1297}.
%Type = Article
\bibitem[{Nakamura et~al.(2006)Nakamura, Markov, Nakaiwa, Kikui, Kawai,
  Jitsuhiro, Zhang, Yamamoto, Sumita, and Yamamoto}]{nakamura2006atr}
\bibinfo{author}{S.~Nakamura}, \bibinfo{author}{K.~Markov},
  \bibinfo{author}{H.~Nakaiwa}, \bibinfo{author}{G.-i. Kikui},
  \bibinfo{author}{H.~Kawai}, \bibinfo{author}{T.~Jitsuhiro},
  \bibinfo{author}{J.-S. Zhang}, \bibinfo{author}{H.~Yamamoto},
  \bibinfo{author}{E.~Sumita}, \bibinfo{author}{S.~Yamamoto},
\newblock \bibinfo{title}{The atr multilingual speech-to-speech translation
  system},
\newblock \bibinfo{journal}{IEEE Transactions on Audio, Speech, and Language
  Processing} \bibinfo{volume}{14} (\bibinfo{year}{2006})
  \bibinfo{pages}{365--376}.
%Type = Inproceedings
\bibitem[{Mikheev et~al.(1999)Mikheev, Moens, and Grover}]{mikheev1999named}
\bibinfo{author}{A.~Mikheev}, \bibinfo{author}{M.~Moens},
  \bibinfo{author}{C.~Grover},
\newblock \bibinfo{title}{Named entity recognition without gazetteers},
\newblock in: \bibinfo{booktitle}{Ninth Conference of the European Chapter of
  the Association for Computational Linguistics}, \bibinfo{year}{1999}, pp.
  \bibinfo{pages}{1--8}.
%Type = Article
\bibitem[{Marrero et~al.(2013)Marrero, Urbano, S{\'a}nchez-Cuadrado, Morato,
  and G{\'o}mez-Berb{\'\i}s}]{marrero2013named}
\bibinfo{author}{M.~Marrero}, \bibinfo{author}{J.~Urbano},
  \bibinfo{author}{S.~S{\'a}nchez-Cuadrado}, \bibinfo{author}{J.~Morato},
  \bibinfo{author}{J.~M. G{\'o}mez-Berb{\'\i}s},
\newblock \bibinfo{title}{Named entity recognition: fallacies, challenges and
  opportunities},
\newblock \bibinfo{journal}{Computer Standards \& Interfaces}
  \bibinfo{volume}{35} (\bibinfo{year}{2013}) \bibinfo{pages}{482--489}.
%Type = Article
\bibitem[{Nadkarni et~al.(2011)Nadkarni, Ohno-Machado, and
  Chapman}]{nadkarni2011natural}
\bibinfo{author}{P.~M. Nadkarni}, \bibinfo{author}{L.~Ohno-Machado},
  \bibinfo{author}{W.~W. Chapman},
\newblock \bibinfo{title}{Natural language processing: an introduction},
\newblock \bibinfo{journal}{Journal of the American Medical Informatics
  Association} \bibinfo{volume}{18} (\bibinfo{year}{2011})
  \bibinfo{pages}{544--551}.
%Type = Incollection
\bibitem[{Socher et~al.(2012)Socher, Bengio, and Manning}]{socher2012deep}
\bibinfo{author}{R.~Socher}, \bibinfo{author}{Y.~Bengio},
  \bibinfo{author}{C.~D. Manning},
\newblock \bibinfo{title}{Deep learning for nlp (without magic)},
\newblock in: \bibinfo{booktitle}{Tutorial Abstracts of ACL 2012},
  \bibinfo{year}{2012}, pp. \bibinfo{pages}{5--5}.
%Type = Article
\bibitem[{Chandra and Saini(2021)}]{chandra2021biden}
\bibinfo{author}{R.~Chandra}, \bibinfo{author}{R.~Saini},
\newblock \bibinfo{title}{{Biden vs Trump: modeling us general elections using
  BERT language }model},
\newblock \bibinfo{journal}{IEEE Access} \bibinfo{volume}{9}
  (\bibinfo{year}{2021}) \bibinfo{pages}{128494--128505}.
%Type = Article
\bibitem[{Garg and Agarwal(2018)}]{garg2018machine}
\bibinfo{author}{A.~Garg}, \bibinfo{author}{M.~Agarwal},
\newblock \bibinfo{title}{Machine translation: a literature review},
\newblock \bibinfo{journal}{arXiv preprint arXiv:1901.01122}
  (\bibinfo{year}{2018}).
%Type = Article
\bibitem[{Mizera-Pietraszko(2010)}]{mizera2010multilingual}
\bibinfo{author}{J.~Mizera-Pietraszko},
\newblock \bibinfo{title}{Multilingual document mining for unstructured
  information},
\newblock \bibinfo{journal}{Pahikkala, V{\"a}yrynen, Kortela and Airola (eds.)}
   (\bibinfo{year}{2010}) \bibinfo{pages}{16}.
%Type = Article
\bibitem[{Oard and Diekema(1998)}]{oard1998cross}
\bibinfo{author}{D.~W. Oard}, \bibinfo{author}{A.~R. Diekema},
\newblock \bibinfo{title}{Cross-language information retrieval.},
\newblock \bibinfo{journal}{Annual Review of Information Science and Technology
  (ARIST)} \bibinfo{volume}{33} (\bibinfo{year}{1998})
  \bibinfo{pages}{223--56}.
%Type = Book
\bibitem[{Beatty(2013)}]{beatty2013teaching}
\bibinfo{author}{K.~Beatty}, \bibinfo{title}{Teaching \& researching:
  Computer-assisted language learning}, \bibinfo{publisher}{Routledge},
  \bibinfo{year}{2013}.
%Type = Inproceedings
\bibitem[{Johnson et~al.(2007)Johnson, Martin, Foster, and
  Kuhn}]{johnson2007improving}
\bibinfo{author}{H.~Johnson}, \bibinfo{author}{J.~Martin},
  \bibinfo{author}{G.~Foster}, \bibinfo{author}{R.~Kuhn},
\newblock \bibinfo{title}{Improving translation quality by discarding most of
  the phrasetable},
\newblock in: \bibinfo{booktitle}{Proceedings of the 2007 Joint Conference on
  Empirical Methods in Natural Language Processing and Computational Natural
  Language Learning (EMNLP-CoNLL)}, \bibinfo{year}{2007}, pp.
  \bibinfo{pages}{967--975}.
%Type = Article
\bibitem[{Bisang et~al.(2022)Bisang, Br{\"u}nnh{\"a}u{\ss}er, L{\"u}nnemann,
  Kirsch, and Lindow}]{bisang2022evaluate}
\bibinfo{author}{U.~Bisang}, \bibinfo{author}{J.~Br{\"u}nnh{\"a}u{\ss}er},
  \bibinfo{author}{P.~L{\"u}nnemann}, \bibinfo{author}{L.~Kirsch},
  \bibinfo{author}{K.~Lindow},
\newblock \bibinfo{title}{Evaluate similarity of requirements with multilingual
  natural language processing},
\newblock \bibinfo{journal}{Proceedings of the Design Society}
  \bibinfo{volume}{2} (\bibinfo{year}{2022}) \bibinfo{pages}{1511--1520}.
%Type = Inproceedings
\bibitem[{Kalchbrenner and Blunsom(2013)}]{kalchbrenner2013recurrent}
\bibinfo{author}{N.~Kalchbrenner}, \bibinfo{author}{P.~Blunsom},
\newblock \bibinfo{title}{Recurrent continuous translation models},
\newblock in: \bibinfo{booktitle}{Proceedings of the 2013 conference on
  empirical methods in natural language processing}, \bibinfo{year}{2013}, pp.
  \bibinfo{pages}{1700--1709}.
%Type = Article
\bibitem[{Zhang and Zong(2020)}]{zhang2020neural}
\bibinfo{author}{J.~Zhang}, \bibinfo{author}{C.~Zong},
\newblock \bibinfo{title}{Neural machine translation: Challenges, progress and
  future},
\newblock \bibinfo{journal}{Science China Technological Sciences}
  \bibinfo{volume}{63} (\bibinfo{year}{2020}) \bibinfo{pages}{2028--2050}.
%Type = Article
\bibitem[{Sutskever et~al.(2014)Sutskever, Vinyals, and
  Le}]{sutskever2014sequence}
\bibinfo{author}{I.~Sutskever}, \bibinfo{author}{O.~Vinyals},
  \bibinfo{author}{Q.~V. Le},
\newblock \bibinfo{title}{Sequence to sequence learning with neural networks},
\newblock \bibinfo{journal}{Advances in neural information processing systems}
  \bibinfo{volume}{27} (\bibinfo{year}{2014}).
%Type = Article
\bibitem[{Sennrich et~al.(2015)Sennrich, Haddow, and
  Birch}]{sennrich2015improving}
\bibinfo{author}{R.~Sennrich}, \bibinfo{author}{B.~Haddow},
  \bibinfo{author}{A.~Birch},
\newblock \bibinfo{title}{Improving neural machine translation models with
  monolingual data},
\newblock \bibinfo{journal}{arXiv preprint arXiv:1511.06709}
  (\bibinfo{year}{2015}).
%Type = Article
\bibitem[{Vaswani et~al.(2017)Vaswani, Shazeer, Parmar, Uszkoreit, Jones,
  Gomez, Kaiser, and Polosukhin}]{vaswani2017attention}
\bibinfo{author}{A.~Vaswani}, \bibinfo{author}{N.~Shazeer},
  \bibinfo{author}{N.~Parmar}, \bibinfo{author}{J.~Uszkoreit},
  \bibinfo{author}{L.~Jones}, \bibinfo{author}{A.~N. Gomez},
  \bibinfo{author}{{\L}.~Kaiser}, \bibinfo{author}{I.~Polosukhin},
\newblock \bibinfo{title}{Attention is all you need},
\newblock \bibinfo{journal}{Advances in neural information processing systems}
  \bibinfo{volume}{30} (\bibinfo{year}{2017}).
%Type = Inproceedings
\bibitem[{Barrault et~al.(2019)Barrault, Bojar, Costa-Jussa, Federmann, Fishel,
  and Graham}]{barrault2019findings}
\bibinfo{author}{L.~Barrault}, \bibinfo{author}{O.~Bojar},
  \bibinfo{author}{M.~R. Costa-Jussa}, \bibinfo{author}{C.~Federmann},
  \bibinfo{author}{M.~Fishel}, \bibinfo{author}{Y.~Graham},
\newblock \bibinfo{title}{Findings of the 2019 conference on machine
  translation (wmt19)},
\newblock \bibinfo{organization}{Association for Computational Linguistics
  (ACL)}, \bibinfo{year}{2019}.
%Type = Article
\bibitem[{Wdowiak(2021)}]{wdowiak2021sicilian}
\bibinfo{author}{E.~Wdowiak},
\newblock \bibinfo{title}{Sicilian translator: A recipe for low-resource
  {NMT}},
\newblock \bibinfo{journal}{arXiv preprint arXiv:2110.01938}
  (\bibinfo{year}{2021}).
%Type = Article
\bibitem[{Mathur et~al.(2020)Mathur, Baldwin, and Cohn}]{mathur2020tangled}
\bibinfo{author}{N.~Mathur}, \bibinfo{author}{T.~Baldwin},
  \bibinfo{author}{T.~Cohn},
\newblock \bibinfo{title}{Tangled up in bleu: Reevaluating the evaluation of
  automatic machine translation evaluation metrics},
\newblock \bibinfo{journal}{arXiv preprint arXiv:2006.06264}
  (\bibinfo{year}{2020}).
%Type = Article
\bibitem[{Devlin et~al.(2018)Devlin, Chang, Lee, and
  Toutanova}]{devlin2018bert}
\bibinfo{author}{J.~Devlin}, \bibinfo{author}{M.-W. Chang},
  \bibinfo{author}{K.~Lee}, \bibinfo{author}{K.~Toutanova},
\newblock \bibinfo{title}{Bert: Pre-training of deep bidirectional transformers
  for language understanding},
\newblock \bibinfo{journal}{arXiv preprint arXiv:1810.04805}
  (\bibinfo{year}{2018}).
%Type = Article
\bibitem[{Tenney et~al.(2019)Tenney, Das, and Pavlick}]{tenney2019bert}
\bibinfo{author}{I.~Tenney}, \bibinfo{author}{D.~Das},
  \bibinfo{author}{E.~Pavlick},
\newblock \bibinfo{title}{Bert rediscovers the classical nlp pipeline},
\newblock \bibinfo{journal}{arXiv preprint arXiv:1905.05950}
  (\bibinfo{year}{2019}).
%Type = Article
\bibitem[{Kitaev et~al.(2020)Kitaev, Kaiser, and Levskaya}]{kitaev2020reformer}
\bibinfo{author}{N.~Kitaev}, \bibinfo{author}{{\L}.~Kaiser},
  \bibinfo{author}{A.~Levskaya},
\newblock \bibinfo{title}{Reformer: The efficient transformer},
\newblock \bibinfo{journal}{arXiv preprint arXiv:2001.04451}
  (\bibinfo{year}{2020}).
%Type = Book
\bibitem[{Gandhi(2010)}]{gandhi2010bhagavad}
\bibinfo{author}{M.~Gandhi}, \bibinfo{title}{The {Bhagavad Gita according to
  Gandhi}}, \bibinfo{publisher}{North Atlantic Books}, \bibinfo{year}{2010}.
%Type = Book
\bibitem[{Hiltebeitel(1976)}]{hiltebeitel1976ritual}
\bibinfo{author}{A.~Hiltebeitel}, \bibinfo{title}{Ritual of Battle, the:
  Krishna in the Mahabharata}, \bibinfo{publisher}{SUNY Press},
  \bibinfo{year}{1976}.
%Type = Book
\bibitem[{Dasgupta(1975)}]{dasgupta1975history}
\bibinfo{author}{S.~Dasgupta}, \bibinfo{title}{A history of Indian philosophy},
  volume~\bibinfo{volume}{2}, \bibinfo{publisher}{Motilal Banarsidass Publ.},
  \bibinfo{year}{1975}.
%Type = Book
\bibitem[{Rajagopalachari(1970)}]{rajagopalachari1970mahabharata}
\bibinfo{author}{C.~Rajagopalachari}, \bibinfo{title}{Mahabharata},
  volume~\bibinfo{volume}{1}, \bibinfo{publisher}{Diamond Pocket Books (P)
  Ltd.}, \bibinfo{year}{1970}.
%Type = Article
\bibitem[{Rao(2002)}]{rao2002mind}
\bibinfo{author}{A.~V. Rao},
\newblock \bibinfo{title}{‘mind’in {Indian} philosophy},
\newblock \bibinfo{journal}{Indian Journal of Psychiatry} \bibinfo{volume}{44}
  (\bibinfo{year}{2002}) \bibinfo{pages}{315}.
%Type = Book
\bibitem[{Gough(2013)}]{gough2013philosophy}
\bibinfo{author}{A.~E. Gough}, \bibinfo{title}{The philosophy of the
  {Upanishads and ancient Indian} metaphysics}, \bibinfo{publisher}{Routledge},
  \bibinfo{year}{2013}.
%Type = Article
\bibitem[{Lomperis(1984)}]{lomperis1984hindu}
\bibinfo{author}{T.~J. Lomperis},
\newblock \bibinfo{title}{{Hindu Influence on Greek Philosophy: The Odyssey of
  the Soul from the Upanishads to Plato}}  (\bibinfo{year}{1984}).
%Type = Book
\bibitem[{Scharfstein(1998)}]{scharfstein1998comparative}
\bibinfo{author}{B.-A. Scharfstein}, \bibinfo{title}{A comparative history of
  world philosophy: From the Upanishads to Kant}, \bibinfo{publisher}{State
  University of New York Press}, \bibinfo{year}{1998}.
%Type = Article
\bibitem[{Chandra and Ranjan(2022)}]{Chandra2022-xi}
\bibinfo{author}{R.~Chandra}, \bibinfo{author}{M.~Ranjan},
\newblock \bibinfo{title}{Artificial intelligence for topic modelling in {Hindu
  philosophy: Mapping themes between the Upanishads and the Bhagavad Gita}},
\newblock \bibinfo{journal}{PloS One} \bibinfo{volume}{17}
  (\bibinfo{year}{2022}) \bibinfo{pages}{e0273476}.
%Type = Misc
\bibitem[{Caswell and Bapn(2022)}]{sanskritgoogle}
\bibinfo{author}{I.~Caswell}, \bibinfo{author}{A.~Bapn},
  \bibinfo{title}{Unlocking zero-resource machine translation to support new
  languages in {Google Translate}, {Google AI Blog}, {(Retrieved August 2nd,
  2022)}}, \bibinfo{year}{2022}.
  \bibinfo{note}{\url{https://ai.googleblog.com/2022/05/24-new-languages-google-translate.html}}.
%Type = Article
\bibitem[{Siddhant et~al.(2020)Siddhant, Bapna, Cao, Firat, Chen, Kudugunta,
  Arivazhagan, and Wu}]{siddhant2020leveraging}
\bibinfo{author}{A.~Siddhant}, \bibinfo{author}{A.~Bapna},
  \bibinfo{author}{Y.~Cao}, \bibinfo{author}{O.~Firat},
  \bibinfo{author}{M.~Chen}, \bibinfo{author}{S.~Kudugunta},
  \bibinfo{author}{N.~Arivazhagan}, \bibinfo{author}{Y.~Wu},
\newblock \bibinfo{title}{Leveraging monolingual data with self-supervision for
  multilingual neural machine translation},
\newblock \bibinfo{journal}{arXiv preprint arXiv:2005.04816}
  (\bibinfo{year}{2020}).
%Type = Inproceedings
\bibitem[{Zhang and Zong(2016)}]{zhang2016exploiting}
\bibinfo{author}{J.~Zhang}, \bibinfo{author}{C.~Zong},
\newblock \bibinfo{title}{Exploiting source-side monolingual data in neural
  machine translation},
\newblock in: \bibinfo{booktitle}{Proceedings of the 2016 Conference on
  Empirical Methods in Natural Language Processing}, \bibinfo{year}{2016}, pp.
  \bibinfo{pages}{1535--1545}.
%Type = Inproceedings
\bibitem[{Zhao et~al.(2015)Zhao, Hassan, and Auli}]{zhao2015learning}
\bibinfo{author}{K.~Zhao}, \bibinfo{author}{H.~Hassan},
  \bibinfo{author}{M.~Auli},
\newblock \bibinfo{title}{Learning translation models from monolingual
  continuous representations},
\newblock in: \bibinfo{booktitle}{Proceedings of the 2015 Conference of the
  North American Chapter of the Association for Computational Linguistics:
  Human Language Technologies}, \bibinfo{year}{2015}, pp.
  \bibinfo{pages}{1527--1536}.
%Type = Article
\bibitem[{Song et~al.(2019)Song, Tan, Qin, Lu, and Liu}]{song2019mass}
\bibinfo{author}{K.~Song}, \bibinfo{author}{X.~Tan}, \bibinfo{author}{T.~Qin},
  \bibinfo{author}{J.~Lu}, \bibinfo{author}{T.-Y. Liu},
\newblock \bibinfo{title}{Mass: Masked sequence to sequence pre-training for
  language generation},
\newblock \bibinfo{journal}{arXiv preprint arXiv:1905.02450}
  (\bibinfo{year}{2019}).
%Type = Inproceedings
\bibitem[{Xiaoning et~al.(2008)Xiaoning, Peidong, Haoliang, Muyun, Guohua, and
  Yong}]{xiaoning2008using}
\bibinfo{author}{H.~Xiaoning}, \bibinfo{author}{W.~Peidong},
  \bibinfo{author}{Q.~Haoliang}, \bibinfo{author}{Y.~Muyun},
  \bibinfo{author}{L.~Guohua}, \bibinfo{author}{X.~Yong},
\newblock \bibinfo{title}{Using google translation in cross-lingual information
  retrieval},
\newblock in: \bibinfo{booktitle}{Proceedings of NTCIR-7 workshop meeting},
  \bibinfo{year}{2008}, pp. \bibinfo{pages}{16--19}.
%Type = Inproceedings
\bibitem[{Li et~al.(2014)Li, Graesser, and Cai}]{li2014comparison}
\bibinfo{author}{H.~Li}, \bibinfo{author}{A.~C. Graesser},
  \bibinfo{author}{Z.~Cai},
\newblock \bibinfo{title}{Comparison of google translation with human
  translation},
\newblock in: \bibinfo{booktitle}{The Twenty-Seventh International Flairs
  Conference}, \bibinfo{year}{2014}.
%Type = Article
\bibitem[{Zand~Rahimi et~al.(2017)Zand~Rahimi, Madayenzadeh, and
  Alizadeh}]{zand2017comparative}
\bibinfo{author}{M.~Zand~Rahimi}, \bibinfo{author}{M.~Madayenzadeh},
  \bibinfo{author}{M.~Alizadeh},
\newblock \bibinfo{title}{A comparative study of english-persian translation of
  neural google translation},
\newblock \bibinfo{journal}{Iranian Journal of Applied Language Studies}
  \bibinfo{volume}{9} (\bibinfo{year}{2017}) \bibinfo{pages}{279--286}.
%Type = Article
\bibitem[{Md~Abdur et~al.(2019)Md~Abdur, Islamb, Hossainc, and
  Jiang}]{md2019exploring}
\bibinfo{author}{R.~Md~Abdur}, \bibinfo{author}{M.~S. Islamb},
  \bibinfo{author}{S.~Hossainc}, \bibinfo{author}{J.~Jiang},
\newblock \bibinfo{title}{Exploring and learning english: An analysis of baidu
  and google translation},
\newblock \bibinfo{journal}{International Journal of Linguistics, Literature
  and Translation (IJLLT)}  (\bibinfo{year}{2019}).
%Type = Article
\bibitem[{Patil and Davies(2014)}]{patil2014use}
\bibinfo{author}{S.~Patil}, \bibinfo{author}{P.~Davies},
\newblock \bibinfo{title}{Use of {Google Translate} in medical communication:
  evaluation of accuracy},
\newblock \bibinfo{journal}{Bmj} \bibinfo{volume}{349} (\bibinfo{year}{2014}).
%Type = Article
\bibitem[{Ranathunga et~al.(2021)Ranathunga, Lee, Skenduli, Shekhar, Alam, and
  Kaur}]{ranathunga2021neural}
\bibinfo{author}{S.~Ranathunga}, \bibinfo{author}{E.-S.~A. Lee},
  \bibinfo{author}{M.~P. Skenduli}, \bibinfo{author}{R.~Shekhar},
  \bibinfo{author}{M.~Alam}, \bibinfo{author}{R.~Kaur},
\newblock \bibinfo{title}{Neural machine translation for low-resource
  languages: A survey},
\newblock \bibinfo{journal}{ACM Computing Surveys}  (\bibinfo{year}{2021}).
%Type = Book
\bibitem[{Gandhi and Desai(1946)}]{desai1946gospel}
\bibinfo{author}{M.~Gandhi}, \bibinfo{author}{M.~Desai}, \bibinfo{title}{The
  Gospel of Selfless Action: Or, The {Gita According to Gandhi}},
  \bibinfo{publisher}{Navajivan Publishing House Ahmedabad},
  \bibinfo{year}{1946}.
%Type = Misc
\bibitem[{Easwaran(1985)}]{easwaran1985trans}
\bibinfo{author}{E.~Easwaran}, \bibinfo{title}{The {Bhagavad Gita}},
  \bibinfo{year}{1985}.
%Type = Book
\bibitem[{Swami(1937)}]{swami1937bhagavad}
\bibinfo{author}{S.~P. Swami}, \bibinfo{title}{Bhagavad {Gita}},
  \bibinfo{publisher}{UK}, \bibinfo{year}{1937}.
%Type = Misc
\bibitem[{Och(2006)}]{googletranslate}
\bibinfo{author}{F.~Och}, \bibinfo{title}{Statistical machine translation live,
  {Google AI Blog} (retrieved august 2nd, 2022)}, \bibinfo{year}{2006}.
  \bibinfo{note}{\url{https://ai.googleblog.com/2006/04/statistical-machine-translation-live.html}}.
%Type = Misc
\bibitem[{Sommerlad(2021)}]{googletranslate3}
\bibinfo{author}{J.~Sommerlad}, \bibinfo{title}{Google translate: How does the
  multilingual interpreter actually work? (retrieved december 12th, 2022)},
  \bibinfo{year}{2021}.
  \bibinfo{note}{\url{https://www.independent.co.uk/tech/how-does-google-translate-work-b1821775.html}}.
%Type = Misc
\bibitem[{Turovsky(2016)}]{googletranslate2}
\bibinfo{author}{B.~Turovsky}, \bibinfo{title}{Found in translation: More
  accurate, fluent sentences in {Google Translate}, {Google Blog, (Retrieved
  August 2nd, 2022)}}, \bibinfo{year}{2016}.
  \bibinfo{note}{\url{https://blog.google/products/translate/found-translation-more-accurate-fluent-sentences-google-translate/}}.
%Type = Misc
\bibitem[{Team(2022)}]{googletranslate4}
\bibinfo{author}{G.~T. Team}, \bibinfo{title}{Translate (retrieved december
  12th, 2022)}, \bibinfo{year}{2022}.
  \bibinfo{note}{\url{https://translate.google.com/intl/en/about/languages/}}.
%Type = Misc
\bibitem[{saac Caswell and Liang(2020)}]{googletranslate5}
\bibinfo{author}{saac Caswell}, \bibinfo{author}{B.~Liang},
  \bibinfo{title}{Recent advances in google translate (retrieved december 13th,
  2022)}, \bibinfo{year}{2020}.
  \bibinfo{note}{\url{https://ai.googleblog.com/2020/06/recent-advances-in-google-translate.html}}.
%Type = Article
\bibitem[{McCartney(????)}]{mccartney2022sanskrit}
\bibinfo{author}{P.~McCartney},
\newblock \bibinfo{title}{{‘Sanskrit-Speaking’} villages, faith-based
  development and the {Indian Census}},
\newblock \bibinfo{journal}{Bhas{\textperiodcentered}}  (????)
  \bibinfo{pages}{77--110}.
%Type = Article
\bibitem[{Edunov et~al.(2018)Edunov, Ott, Auli, and
  Grangier}]{edunov2018understanding}
\bibinfo{author}{S.~Edunov}, \bibinfo{author}{M.~Ott},
  \bibinfo{author}{M.~Auli}, \bibinfo{author}{D.~Grangier},
\newblock \bibinfo{title}{Understanding back-translation at scale},
\newblock \bibinfo{journal}{arXiv preprint arXiv:1808.09381}
  (\bibinfo{year}{2018}).
%Type = Book
\bibitem[{Prabhupada(1972)}]{prabhupada1972bhagavad}
\bibinfo{author}{A.~B.~S. Prabhupada}, \bibinfo{title}{{Bhagavad Gita} as it
  is}, \bibinfo{publisher}{Bhaktivedanta Book Trust Los Angeles},
  \bibinfo{year}{1972}.
%Type = Article
\bibitem[{Bright(1996)}]{bright1996devanagari}
\bibinfo{author}{W.~Bright},
\newblock \bibinfo{title}{The devanagari script},
\newblock \bibinfo{journal}{The world’s writing systems}
  (\bibinfo{year}{1996}) \bibinfo{pages}{384--390}.
%Type = Incollection
\bibitem[{Li and Yang(2018)}]{li2018word}
\bibinfo{author}{Y.~Li}, \bibinfo{author}{T.~Yang},
\newblock \bibinfo{title}{Word embedding for understanding natural language: a
  survey},
\newblock in: \bibinfo{booktitle}{Guide to big data applications},
  \bibinfo{publisher}{Springer}, \bibinfo{year}{2018}, pp.
  \bibinfo{pages}{83--104}.
%Type = Inproceedings
\bibitem[{Ghannay et~al.(2016)Ghannay, Favre, Esteve, and
  Camelin}]{ghannay2016word}
\bibinfo{author}{S.~Ghannay}, \bibinfo{author}{B.~Favre},
  \bibinfo{author}{Y.~Esteve}, \bibinfo{author}{N.~Camelin},
\newblock \bibinfo{title}{Word embedding evaluation and combination},
\newblock in: \bibinfo{booktitle}{Proceedings of the Tenth International
  Conference on Language Resources and Evaluation (LREC'16)},
  \bibinfo{year}{2016}, pp. \bibinfo{pages}{300--305}.
%Type = Article
\bibitem[{Wang et~al.(2019)Wang, Wang, Chen, Wang, and
  Kuo}]{wang2019evaluating}
\bibinfo{author}{B.~Wang}, \bibinfo{author}{A.~Wang},
  \bibinfo{author}{F.~Chen}, \bibinfo{author}{Y.~Wang},
  \bibinfo{author}{C.-C.~J. Kuo},
\newblock \bibinfo{title}{Evaluating word embedding models: methods and
  experimental results},
\newblock \bibinfo{journal}{APSIPA transactions on signal and information
  processing} \bibinfo{volume}{8} (\bibinfo{year}{2019}).
%Type = Misc
\bibitem[{Mikolov et~al.(2013)Mikolov, Chen, Corrado, and Dean}]{mikolov}
\bibinfo{author}{T.~Mikolov}, \bibinfo{author}{K.~Chen},
  \bibinfo{author}{G.~Corrado}, \bibinfo{author}{J.~Dean},
  \bibinfo{title}{Efficient estimation of word representations in vector
  space}, \bibinfo{year}{2013}. \URLprefix
  \url{https://arxiv.org/abs/1301.3781}.
  \DOIprefix\doi{10.48550/ARXIV.1301.3781}.
%Type = Misc
\bibitem[{Devlin et~al.(2018)Devlin, Chang, Lee, and Toutanova}]{bert}
\bibinfo{author}{J.~Devlin}, \bibinfo{author}{M.-W. Chang},
  \bibinfo{author}{K.~Lee}, \bibinfo{author}{K.~Toutanova},
  \bibinfo{title}{Bert: Pre-training of deep bidirectional transformers for
  language understanding}, \bibinfo{year}{2018}. \URLprefix
  \url{https://arxiv.org/abs/1810.04805}.
  \DOIprefix\doi{10.48550/ARXIV.1810.04805}.
%Type = Inproceedings
\bibitem[{Shen and Liu(2021)}]{9647258}
\bibinfo{author}{Y.~Shen}, \bibinfo{author}{J.~Liu},
\newblock \bibinfo{title}{Comparison of text sentiment analysis based on bert
  and word2vec},
\newblock in: \bibinfo{booktitle}{2021 IEEE 3rd International Conference on
  Frontiers Technology of Information and Computer (ICFTIC)},
  \bibinfo{year}{2021}, pp. \bibinfo{pages}{144--147}.
  \DOIprefix\doi{10.1109/ICFTIC54370.2021.9647258}.
%Type = Article
\bibitem[{Medhat et~al.(2014)Medhat, Hassan, and Korashy}]{medhat2014sentiment}
\bibinfo{author}{W.~Medhat}, \bibinfo{author}{A.~Hassan},
  \bibinfo{author}{H.~Korashy},
\newblock \bibinfo{title}{Sentiment analysis algorithms and applications: A
  survey},
\newblock \bibinfo{journal}{Ain Shams engineering journal} \bibinfo{volume}{5}
  (\bibinfo{year}{2014}) \bibinfo{pages}{1093--1113}.
%Type = Article
\bibitem[{Zhang et~al.(2018)Zhang, Wang, and Liu}]{zhang2018deep}
\bibinfo{author}{L.~Zhang}, \bibinfo{author}{S.~Wang},
  \bibinfo{author}{B.~Liu},
\newblock \bibinfo{title}{Deep learning for sentiment analysis: A survey},
\newblock \bibinfo{journal}{Wiley Interdisciplinary Reviews: Data Mining and
  Knowledge Discovery} \bibinfo{volume}{8} (\bibinfo{year}{2018})
  \bibinfo{pages}{e1253}.
%Type = Article
\bibitem[{Asghar et~al.(2017)Asghar, Khan, Ahmad, Qasim, and
  Khan}]{asghar2017lexicon}
\bibinfo{author}{M.~Z. Asghar}, \bibinfo{author}{A.~Khan},
  \bibinfo{author}{S.~Ahmad}, \bibinfo{author}{M.~Qasim},
  \bibinfo{author}{I.~A. Khan},
\newblock \bibinfo{title}{Lexicon-enhanced sentiment analysis framework using
  rule-based classification scheme},
\newblock \bibinfo{journal}{PloS one} \bibinfo{volume}{12}
  (\bibinfo{year}{2017}) \bibinfo{pages}{e0171649}.
%Type = Incollection
\bibitem[{Mohammad(2016)}]{mohammad2016sentiment}
\bibinfo{author}{S.~M. Mohammad},
\newblock \bibinfo{title}{Sentiment analysis: Detecting valence, emotions, and
  other affectual states from text},
\newblock in: \bibinfo{booktitle}{Emotion measurement},
  \bibinfo{publisher}{Elsevier}, \bibinfo{year}{2016}, pp.
  \bibinfo{pages}{201--237}.
%Type = Article
\bibitem[{Appel et~al.(2016)Appel, Chiclana, Carter, and
  Fujita}]{appel2016hybrid}
\bibinfo{author}{O.~Appel}, \bibinfo{author}{F.~Chiclana},
  \bibinfo{author}{J.~Carter}, \bibinfo{author}{H.~Fujita},
\newblock \bibinfo{title}{A hybrid approach to the sentiment analysis problem
  at the sentence level},
\newblock \bibinfo{journal}{Knowledge-Based Systems} \bibinfo{volume}{108}
  (\bibinfo{year}{2016}) \bibinfo{pages}{110--124}.
%Type = Article
\bibitem[{Feldman(2013)}]{feldman2013techniques}
\bibinfo{author}{R.~Feldman},
\newblock \bibinfo{title}{Techniques and applications for sentiment analysis},
\newblock \bibinfo{journal}{Communications of the ACM} \bibinfo{volume}{56}
  (\bibinfo{year}{2013}) \bibinfo{pages}{82--89}.
%Type = Book
\bibitem[{Goddard(2011)}]{goddard2011semantic}
\bibinfo{author}{C.~Goddard}, \bibinfo{title}{Semantic analysis: A practical
  introduction}, \bibinfo{publisher}{Oxford University Press},
  \bibinfo{year}{2011}.
%Type = Inproceedings
\bibitem[{Nasukawa and Yi(2003)}]{nasukawa2003sentiment}
\bibinfo{author}{T.~Nasukawa}, \bibinfo{author}{J.~Yi},
\newblock \bibinfo{title}{Sentiment analysis: Capturing favorability using
  natural language processing},
\newblock in: \bibinfo{booktitle}{Proceedings of the 2nd international
  conference on Knowledge capture}, \bibinfo{year}{2003}, pp.
  \bibinfo{pages}{70--77}.
%Type = Article
\bibitem[{Maulud et~al.(2021)Maulud, Zeebaree, Jacksi, Sadeeq, and
  Sharif}]{maulud2021state}
\bibinfo{author}{D.~H. Maulud}, \bibinfo{author}{S.~R. Zeebaree},
  \bibinfo{author}{K.~Jacksi}, \bibinfo{author}{M.~A.~M. Sadeeq},
  \bibinfo{author}{K.~H. Sharif},
\newblock \bibinfo{title}{State of art for semantic analysis of natural
  language processing},
\newblock \bibinfo{journal}{Qubahan Academic Journal} \bibinfo{volume}{1}
  (\bibinfo{year}{2021}) \bibinfo{pages}{21--28}.
%Type = Article
\bibitem[{Yang et~al.(2020)Yang, Alamro, Albaradei, Salhi, Lv, Ma, Alshehri,
  Jaber, Tifratene, Wang et~al.}]{yang2020senwave}
\bibinfo{author}{Q.~Yang}, \bibinfo{author}{H.~Alamro},
  \bibinfo{author}{S.~Albaradei}, \bibinfo{author}{A.~Salhi},
  \bibinfo{author}{X.~Lv}, \bibinfo{author}{C.~Ma},
  \bibinfo{author}{M.~Alshehri}, \bibinfo{author}{I.~Jaber},
  \bibinfo{author}{F.~Tifratene}, \bibinfo{author}{W.~Wang}, et~al.,
\newblock \bibinfo{title}{Senwave: Monitoring the global sentiments under the
  covid-19 pandemic},
\newblock \bibinfo{journal}{arXiv preprint arXiv:2006.10842}
  (\bibinfo{year}{2020}).
%Type = Article
\bibitem[{Chandra and Krishna(2021)}]{chandra2021covid}
\bibinfo{author}{R.~Chandra}, \bibinfo{author}{A.~Krishna},
\newblock \bibinfo{title}{{COVID-19} sentiment analysis via deep learning
  during the rise of novel cases},
\newblock \bibinfo{journal}{PloS one} \bibinfo{volume}{16}
  (\bibinfo{year}{2021}) \bibinfo{pages}{e0255615}.
%Type = Article
\bibitem[{Song et~al.(2020)Song, Tan, Qin, Lu, and Liu}]{mpnet}
\bibinfo{author}{K.~Song}, \bibinfo{author}{X.~Tan}, \bibinfo{author}{T.~Qin},
  \bibinfo{author}{J.~Lu}, \bibinfo{author}{T.-Y. Liu},
\newblock \bibinfo{title}{Mpnet: Masked and permuted pre-training for language
  understanding},
\newblock \bibinfo{journal}{Advances in Neural Information Processing Systems}
  \bibinfo{volume}{33} (\bibinfo{year}{2020}) \bibinfo{pages}{16857--16867}.
%Type = Misc
\bibitem[{McInnes et~al.(2018)McInnes, Healy, and Melville}]{UMAP}
\bibinfo{author}{L.~McInnes}, \bibinfo{author}{J.~Healy},
  \bibinfo{author}{J.~Melville}, \bibinfo{title}{Umap: Uniform manifold
  approximation and projection for dimension reduction}, \bibinfo{year}{2018}.
  \bibinfo{note}{\url{https://arxiv.org/abs/1802.03426}}.
%Type = Inbook
\bibitem[{Rose et~al.(2010)Rose, Engel, Cramer, and Cowley}]{RAKE}
\bibinfo{author}{S.~Rose}, \bibinfo{author}{D.~Engel},
  \bibinfo{author}{N.~Cramer}, \bibinfo{author}{W.~Cowley},
  \bibinfo{title}{Automatic Keyword Extraction from Individual Documents},
  \bibinfo{publisher}{John Wiley and Sons, Ltd}, \bibinfo{year}{2010}, pp.
  \bibinfo{pages}{1--20}. \URLprefix
  \url{https://onlinelibrary.wiley.com/doi/abs/10.1002/9780470689646.ch1}.
  \DOIprefix\doi{https://doi.org/10.1002/9780470689646.ch1}.
%Type = Article
\bibitem[{Campos et~al.(2020)Campos, Mangaravite, Pasquali, Jorge, Nunes, and
  Jatowt}]{Yake}
\bibinfo{author}{R.~Campos}, \bibinfo{author}{V.~Mangaravite},
  \bibinfo{author}{A.~Pasquali}, \bibinfo{author}{A.~Jorge},
  \bibinfo{author}{C.~Nunes}, \bibinfo{author}{A.~Jatowt},
\newblock \bibinfo{title}{Yake! keyword extraction from single documents using
  multiple local features},
\newblock \bibinfo{journal}{Information Sciences} \bibinfo{volume}{509}
  (\bibinfo{year}{2020}) \bibinfo{pages}{257--289}.
%Type = Article
\bibitem[{Salton(1984)}]{TF-IDF}
\bibinfo{author}{G.~Salton},
\newblock \bibinfo{title}{The use of extended boolean logic in information
  retrieval},
\newblock \bibinfo{journal}{SIGMOD Rec.} \bibinfo{volume}{14}
  (\bibinfo{year}{1984}) \bibinfo{pages}{277–285}.
%Type = Article
\bibitem[{Robertson and Willett(1998)}]{robertson1998applications}
\bibinfo{author}{A.~M. Robertson}, \bibinfo{author}{P.~Willett},
\newblock \bibinfo{title}{Applications of n-grams in textual information
  systems},
\newblock \bibinfo{journal}{Journal of Documentation}  (\bibinfo{year}{1998}).
%Type = Book
\bibitem[{Badhe(2015)}]{badhe2015}
\bibinfo{author}{S.~Badhe}, \bibinfo{title}{Bhagavad Gita: Rhythm of Krishna},
  \bibinfo{publisher}{Sri Aurobindo's Action}, \bibinfo{year}{2015}.
%Type = Article
\bibitem[{Bhatia et~al.(2013)Bhatia, Madabushi, Kolli, Bhatia, and
  Madaan}]{bhatia2013bhagavad}
\bibinfo{author}{S.~C. Bhatia}, \bibinfo{author}{J.~Madabushi},
  \bibinfo{author}{V.~Kolli}, \bibinfo{author}{S.~K. Bhatia},
  \bibinfo{author}{V.~Madaan},
\newblock \bibinfo{title}{The {Bhagavad Gita} and contemporary
  psychotherapies},
\newblock \bibinfo{journal}{Indian journal of psychiatry} \bibinfo{volume}{55}
  (\bibinfo{year}{2013}) \bibinfo{pages}{S315}.
%Type = Book
\bibitem[{Sharma(1960)}]{sharma1960history}
\bibinfo{author}{B.~K. Sharma}, \bibinfo{title}{A history of the {Dvaita school
  of Vedanta} and its literature}, volume~\bibinfo{volume}{2},
  \bibinfo{publisher}{Motilal Banarsidass Publishe}, \bibinfo{year}{1960}.
%Type = Article
\bibitem[{Anderson(2012)}]{anderson2012investigation}
\bibinfo{author}{J.~Anderson},
\newblock \bibinfo{title}{An investigation of {Moksha in the Advaita Vedanta of
  Shankara and Gaudapada}},
\newblock \bibinfo{journal}{Asian Philosophy} \bibinfo{volume}{22}
  (\bibinfo{year}{2012}) \bibinfo{pages}{275--287}.
%Type = Book
\bibitem[{Capra(2010)}]{capra2010tao}
\bibinfo{author}{F.~Capra}, \bibinfo{title}{The {Tao} of physics: An
  exploration of the parallels between modern physics and eastern mysticism},
  \bibinfo{publisher}{Shambhala publications}, \bibinfo{year}{2010}.
%Type = Book
\bibitem[{Phillips(1986)}]{phillips1986aurobindinos}
\bibinfo{author}{S.~H. Phillips}, \bibinfo{title}{Aurobindinos Philosophy of
  {Brahman}}, \bibinfo{publisher}{Brill Archive}, \bibinfo{year}{1986}.
%Type = Article
\bibitem[{Chaudhuri(1954)}]{chaudhuri1954concept}
\bibinfo{author}{H.~Chaudhuri},
\newblock \bibinfo{title}{The concept of {Brahman in Hindu} philosophy},
\newblock \bibinfo{journal}{Philosophy East and West} \bibinfo{volume}{4}
  (\bibinfo{year}{1954}) \bibinfo{pages}{47--66}.
%Type = Article
\bibitem[{Krishna(????)}]{krishnaancient}
\bibinfo{author}{K.~Krishna},
\newblock \bibinfo{title}{The ancient {Indian Poornam Mantra} and the paradox
  of infinity},
\newblock \bibinfo{journal}{Pi in the Sky}  (????) \bibinfo{pages}{16}.
%Type = Book
\bibitem[{Easwaran(2007)}]{easwaran2007upanishads}
\bibinfo{author}{E.~Easwaran}, \bibinfo{title}{The {Upanishads}},
  volume~\bibinfo{volume}{2}, \bibinfo{publisher}{Nilgiri Press},
  \bibinfo{year}{2007}.
%Type = Article
\bibitem[{Greeff(1998)}]{greeff1998mysticism}
\bibinfo{author}{T.~d. Greeff},
\newblock \bibinfo{title}{The mysticism of {Isha Upanishad}},
\newblock \bibinfo{journal}{Indian theological studies} \bibinfo{volume}{35}
  (\bibinfo{year}{1998}) \bibinfo{pages}{265--290}.
%Type = Article
\bibitem[{Whitney(1890)}]{whitney1890bohtlingk}
\bibinfo{author}{W.~D. Whitney},
\newblock \bibinfo{title}{B{\"o}htlingk's upanishads},
\newblock \bibinfo{journal}{The American Journal of Philology}
  \bibinfo{volume}{11} (\bibinfo{year}{1890}) \bibinfo{pages}{407--439}.
%Type = Book
\bibitem[{Mehta(1970)}]{mehta1970call}
\bibinfo{author}{R.~Mehta}, \bibinfo{title}{The call of the Upanishads},
  \bibinfo{publisher}{Motilal Banarsidass Publ.}, \bibinfo{year}{1970}.
%Type = Article
\bibitem[{Nelson(1998)}]{nelson1998dualism}
\bibinfo{author}{L.~E. Nelson},
\newblock \bibinfo{title}{The dualism of nondualism: {Advaita Vedanta} and the
  irrelevance of nature},
\newblock \bibinfo{journal}{Purifying the earthly body of God: Religion and
  ecology in Hindu India}  (\bibinfo{year}{1998}) \bibinfo{pages}{61--88}.
%Type = Article
\bibitem[{Widgery(1942)}]{widgery1942dvaita}
\bibinfo{author}{A.~G. Widgery},
\newblock \bibinfo{title}{The {Dvaita} philosophy and its place in the
  {Vedanta}},
\newblock \bibinfo{journal}{The Philosophical Review} \bibinfo{volume}{51}
  (\bibinfo{year}{1942}) \bibinfo{pages}{618--621}.
%Type = Article
\bibitem[{Varma(2018)}]{varma2018adi}
\bibinfo{author}{P.~K. Varma},
\newblock \bibinfo{title}{{Adi Shankaracharya}: Hinduism's greatest thinker}
  (\bibinfo{year}{2018}).
%Type = Article
\bibitem[{Namboodiripad(1989)}]{namboodiripad1989adi}
\bibinfo{author}{E.~Namboodiripad},
\newblock \bibinfo{title}{Adi sankara and his philosophy: A marxist view},
\newblock \bibinfo{journal}{Social scientist}  (\bibinfo{year}{1989})
  \bibinfo{pages}{3--12}.
%Type = Book
\bibitem[{Gambhirananda(1984)}]{gambhirananda1984bhagavad}
\bibinfo{author}{S.~Gambhirananda}, \bibinfo{title}{Bhagavad {Gita: With the
  commentary of Shankaracharya}}, \bibinfo{publisher}{Advaita Ashrama (A
  publication branch of Ramakrishna Math, Belur Math)}, \bibinfo{year}{1984}.

\end{thebibliography}

\end{document}